%% file: 360monodepth.tex
\documentclass[10pt,twocolumn,letterpaper]{article}

\usepackage[utf8]{inputenc}
\usepackage[T1]{fontenc}
\usepackage{cvpr}              %

\usepackage[accsupp]{axessibility}
\usepackage{textcomp}
\usepackage{epsfig}
\usepackage{graphicx}
\usepackage{amsmath}
\usepackage{amssymb}
\usepackage[numbers,sort]{natbib}
\usepackage{booktabs}
\usepackage{multirow}
\usepackage{xcolor}
\usepackage{comment}
\usepackage{microtype}
\makeatletter\@namedef{ver@everyshi.sty}{}\makeatother %
\usepackage{tikz}
\usetikzlibrary{calc,spy,arrows,arrows.meta} %
\usepackage{capt-of}
\graphicspath{{figures/}}

\makeatletter
\renewcommand\paragraph{\@startsection{paragraph}{4}{\z@}%
	{0.75ex \@plus.5ex \@minus.2ex}%
	{-1em}%
	{\normalfont\normalsize\bfseries\maybe@addperiod}}
\newcommand{\maybe@addperiod}[1]{#1\@addpunct{.}}
\makeatother

\usepackage[breaklinks,colorlinks,bookmarks=false,linkcolor=blue,citecolor=blue,urlcolor=blue]{hyperref} %
\usepackage[nameinlink,noabbrev,capitalise]{cleveref}
\crefformat{equation}{#2Equation~#1#3}
\def\cvprPaperID{41}
\def\confName{CVPR}
\def\confYear{2022}

\newcommand{\abs}[1]{\left\lvert#1\right\rvert}
\newcommand{\norm}[1]{\left\lVert#1\right\rVert}

\DeclareMathOperator*{\argmin}{argmin}
\DeclareMathOperator*{\median}{median}

\newcommand{\MData}{\textsuperscript{M}} %
\newcommand{\SData}{\textsuperscript{S}} %
\newcommand{\Mvii}{\textsuperscript{M2}} %
\newcommand{\Mviii}{\textsuperscript{M3}} %

\newcommand{\triup}{\textcolor{blue!70!black}{$\blacktriangle$}}
\newcommand{\tridown}{\textcolor{blue!70!black}{$\blacktriangledown$}}

\AtBeginDocument{%
\setlength\abovedisplayskip{4pt plus 1pt minus 3pt} %
\setlength\belowdisplayskip{4pt plus 1pt minus 3pt} %
\setlength\abovedisplayshortskip{0pt plus 3pt} %
\setlength\belowdisplayshortskip{4pt plus 1.5pt minus 1.5pt} %
}

\begin{document}
\title{360MonoDepth: High-Resolution 360° Monocular Depth Estimation}

\author{%
	\begin{tabular}{c@{\qquad}c@{\qquad}c}
		Manuel Rey-Area\footnotemark &
		Mingze Yuan\footnotemark[1] &
		Christian Richardt
	\end{tabular}
	\\[0.25em]%
	University of Bath%
}

\twocolumn[{
	\renewcommand\twocolumn[1][]{#1}
	\maketitle
	{\centering
		\includegraphics[width=\linewidth]{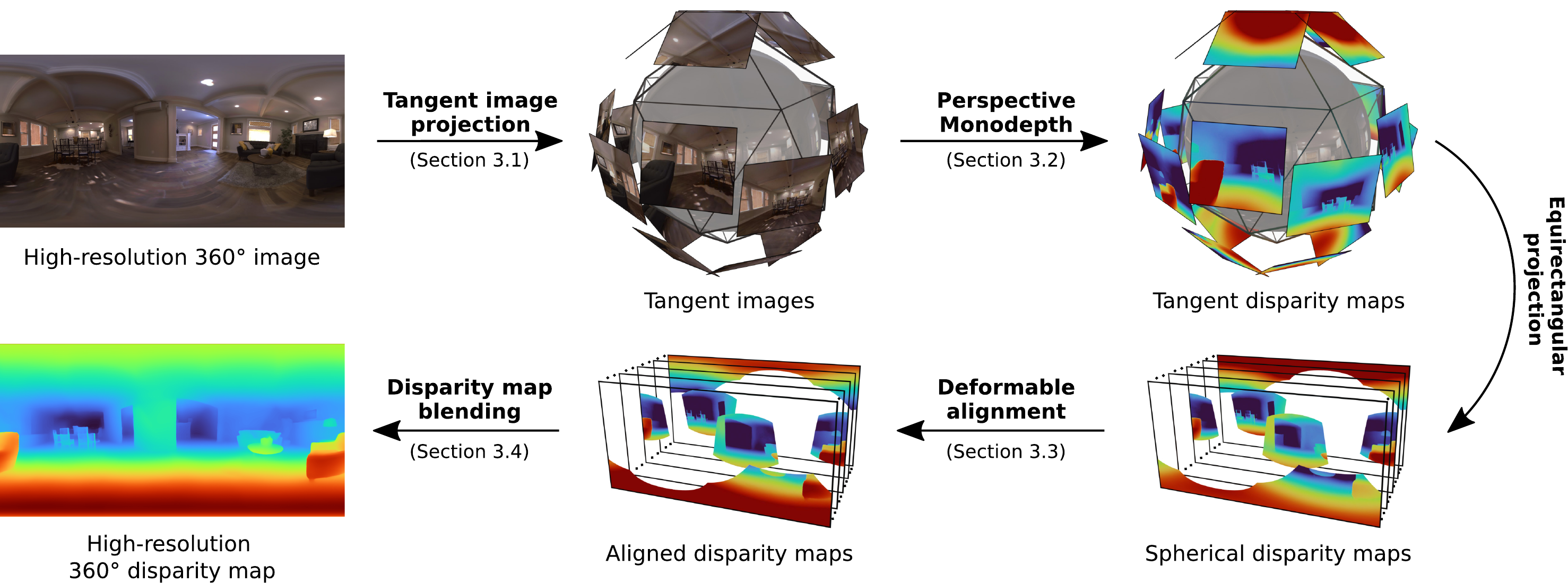}
		\captionof{figure}{\label{fig:teaser}%
			We present a flexible framework for estimating high-resolution disparity maps from a single 360° input image by decomposing it into perspective tangent images, which are used for monocular depth estimation.
			We then globally align all disparity maps using multi-scale alignment fields, and blend them in the gradient domain to produce a detailed, consistent and high-resolution 360° spherical disparity map.
		}
	}
	\vspace{0.5cm}
}]

\begin{abstract}
360° cameras can capture complete environments in a single shot, which makes 360° imagery alluring in many computer vision tasks.
However, monocular depth estimation remains a challenge for 360° data, particularly for high resolutions like 2K (2048$\times$1024) and beyond that are important for novel-view synthesis and virtual reality applications.
Current CNN-based methods do not support such high resolutions due to limited GPU memory.
In this work, we propose a flexible framework for monocular depth estimation from high-resolution 360° images using tangent images.
We project the 360° input image onto a set of tangent planes that produce perspective views, which are suitable for the latest, most accurate state-of-the-art perspective monocular depth estimators.
To achieve globally consistent disparity estimates, we recombine the individual depth estimates using deformable multi-scale alignment followed by gradient-domain blending. %
The result is a dense, high-resolution 360° depth map with a high level of detail, also for outdoor scenes which are not supported by existing methods.
Our source code and data are available at \href{https://manurare.github.io/360monodepth/}{https://manurare.github.io/360monodepth/}.
\end{abstract}

\def\thefootnote{*}\footnotetext{These authors contributed equally to this work.} %
\def\thefootnote{\arabic{footnote}}

\input{1-introduction}

\input{2-related-work}

\input{3-method}

\input{4-results}
\input{5-discussion}

\paragraph{Acknowledgements}
This work was supported by
the EPSRC CDT in Digital Entertainment (EP/L016540/1),
an EPSRC-UKRI Innovation Fellowship (EP/S001050/1)
and EPSRC grant CAMERA (EP/M023281/1, EP/T022523/1).

\input{s-supplement}

{\small
\bibliographystyle{ieeenat_fullname}
\bibliography{360monodepth}
}

\end{document}

%% file: 1-introduction.tex
\section{Introduction}

Monocular depth estimation has recently seen a significant boost thanks to convolutional neural networks.
CNNs have demonstrated an unprecedented expressive power to learn intricate geometric relationships from data, resembling the capability of humans to exploit visual cues to perceive depth.
Monocular depth estimates have enabled impressive new approaches for 3D photography \cite{ShihSKH2020, KopfMAQGCPFWYZHVSC2020} and novel-view synthesis of dynamic scenes \cite{LiNSW2021, GaoSKH2021}.
However, most approaches for monocular depth estimation are limited to low-resolution\footnote{For example, 384×384$\approx$0.15\,megapixels for MiDaS \cite{RanftLHSK2021,RanftBK2021}.} perspective images, with a limited field-of-view.

Nevertheless, 360° cameras are becoming increasingly popular and widespread in the computer vision community.
The omnidirectional 360° field-of-view captured by these devices is appealing for tasks such as
robust, omnidirectional SLAM \cite{WonSCPL2020, SumikSS2019},
scene understanding and layout estimation \cite{JinXZZTXYG2020, ZengKG2020, SunSC2021, WangYSCT2021},
or VR photography and video \cite{BerteYLR2020, SerraKCDGHM2019}.
State-of-the-art monocular depth estimation approaches for 360° images \cite{JiangSZDH2021, WangYSCT2020, PintoAASG2021, SunSC2021, LiGYHDR2022} are currently limited to resolutions of 1024×512$\approx$0.5\,megapixels.
While this is sufficient for tasks like layout estimation, it is insufficient for VR applications as they require resolutions of at least 2\,megapixels to match the resolution of VR headsets \cite{KouliASMMR2019} and achieve full immersion \cite{LouisTRB2019, CummiB2016}.
Our work aims to fill this gap.

Existing monocular 360° depth estimation approaches build on CNNs whose spatial resolution is fundamentally limited by the GPU memory available during training.
These methods are therefore restricted to small batch sizes of 4 to 8 for 0.5\,megapixel images on an NVIDIA 2080 Ti with 11\,GB memory \cite{JiangSZDH2021, PintoAASG2021, SunSC2021}.
For this reason, single-CNN approaches become impractical for predicting high-resolution depth maps with multiple megapixels.

In this work, we introduce a general and flexible framework for monocular depth estimation from high-resolution 360° images inspired by Eder et al.'s tangent images \cite{EderSLF2020}.
Our approach projects the input 360° image to a collection of perspective tangent images, e.g. using the faces of an icosahedron.
We then use state-of-the-art perspective monocular depth estimators endowed with powerful generalisation capability for obtaining dense, detailed depth maps for each tangent image.
Subsequently, we optimally align individual depth maps using multi-scale spatially-varying deformation fields to bring them into global agreement.
Finally, we merge the aligned depth maps using gradient-based blending for a seamless high-resolution 360° depth map.
Our technical contributions are as follow:
\begin{enumerate}\itemsep0em
\item
A simple, yet powerful and practical framework for high-quality multi-megapixel 360° monocular depth estimation based on aligning and blending depth maps predicted from perspective tangent images.

\item
Support for increased resolutions
using tangent images, and improved quality by forward compatibility for future monocular depth estimation approaches.

\item
We provide 2048×1024 ground-truth depth maps for Matterport3D's stitched skyboxes to advance future high-resolution depth estimation approaches.
\end{enumerate}

%% file: 2-related-work.tex
\section{Related Work}

\paragraph{Monocular depth estimation}

Predicting a dense depth map from a single input image is a challenging, ill-posed task due to the high level of ambiguity between possible reconstructions.
Early approaches relied on simple geometric assumptions \cite{HoiemEH2005},
geometric reasoning using Markov random fields \cite{SaxenSN2009},
or non-parametric depth transfer \cite{KarscLK2014}.
The rise of deep learning has made it possible to train convolutional neural networks that are supervised by ground-truth depth maps \cite{EigenPF2014,LiuSH2014,LainaRBTN2016}, e.g. from synthetic renderings or depth sensors, or by exploiting defocus blur \cite{ShiTXJ2015,SriniGWNB2018}.
However, suitable training data is scarce, particularly for outdoor scenes.
\looseness-1

Subsequent work therefore explored alternative training regimes, in particular
from stereo views that provide self-supervision via view synthesis \cite{GargKCR2016,GodarAB2017,LuoRLPSLL2018,XianSCLXLL2018,GodarAFB2019,WangLPW2019,RamamFWLT2021},
from camera ego-motion in videos \cite{ZhouBSL2017,GordoLJA2019,ZhanGWLAR2018,WangBZL2018,RanjaJBKSWB2018,LuoYWWXNY2020,MahjoWA2018},
and from multiview stereo reconstructions \cite{LiS2018,LiDCTSLF2021}.
Ranftl and Lasinger et al.'s MiDaS \cite{RanftLHSK2021} demonstrated substantial improvements and generalisation performance by learning from five varied datasets using multi-objective learning.
The fidelity of depth predictions can also be improved by merging estimates at multiple scales \cite{MiangDMPA2021}.
Recently, \citet{RanftBK2021} introduced transformers \cite{VaswaSPUJGKP2017,DosovBKWZUDMHGUH2021} into monocular depth estimation, to produce finer-grained and more globally consistent results than CNN-based methods.
We base our new monocular 360° depth estimation method on their state-of-the-art performance, but our method would transparently benefit from future advances in monocular depth estimation.

\paragraph{Spherical CNNs}

Most CNNs are applied to flat 2D images with little image distortion.
However, 360° images need a different approach to correctly handle the inevitable distortions of projecting a spherical image onto a plane, e.g. in the commonly used equirectangular projection.
Su and Grauman proposed a pragmatic solution using wider kernels near the poles \cite{SuG2017a}.
However, these kernels do not share any information, which leads to suboptimal performance.
Another pragmatic approach is to project the spherical image into a padded cubemap, process all sides as perspective images, and to recombine the results \cite{ChengCDWLS2018}.
This approach struggles for the top and bottom faces, as kernel orientations become ambiguous due to 90-degree rotational symmetry.
\citet{EderSLF2020} generalise this approach to more than six tangent images, which achieves higher and more uniform angular pixel resolutions.
However, predictions on tangent images are recombined per pixel without any alignment or blending, which works poorly for monocular depth estimation (see our experiments in \cref{sec:ablations}).

Cubemaps have since been generalised to the 20 triangular faces of an icosahedron, which can be unwrapped into 5 rectangles with shared convolution kernels \cite{LeeJYCY2022,ZhangLSC2019,CohenWKW2019}. %
Distortion-aware convolutions \cite{TatenNT2018,CoorsCG2018,SuG2019,ZhaoZDMJZ2018,FernaFPDCG2020} can directly model the distortions of equirectangular projection.
Interestingly, this also enables the transfer of models trained on perspective images to equirectangular images without any additional training, but it requires matching angular pixel resolutions.
Full rotation-equi\-variance can be achieved using spherical convolutions \cite{CohenGKW2018,EstevAMD2018}, but this may not always be desirable as the down direction is usually consistent with gravity.
These approaches have high memory requirements that make them unsuitable for multi-megapixel resolutions.

\paragraph{360° depth estimation}

Deep learning has also boosted monocular depth estimation for 360° images.
Most methods are supervised using synthetic datasets due to the difficulty of acquiring ground-truth spherical depth maps \cite{ZioulKZD2018,EderMG2019}.
Similar to the perspective case, several methods perform self-supervised training via view synthesis \cite{LaGarAB2018,WangHCLYSCS2018,ZioulKZAD2019,LiYDR2021}.
\citet{TatenNT2018} adapt pre-trained monocular depth estimation for perspective images \cite{LainaRBTN2016} to spherical images using distortion-aware convolutional filters.
Depth accuracy can be improved by fusing predictions for equirectangular and cubemap projections \cite{WangYSCT2020, BaiLQGG2022},
while deformable \cite{ChenLFLCG2021} or dilated \cite{ZhuanLWXW2022} convolutions can make methods more distortion-aware.
\citet{PintoAASG2021} and \citet{SunSC2021} exploit gravity-aligned features in man-made interior environments using vertical slicing.
However, the performance of these learning-based approaches highly depends on their training data.
Most datasets are synthetic, low-resolution (1024$\times$512) and only consider indoor scenes.
These methods therefore tend to perform poorly on real high-resolution or outdoor scenes.

Learning-based spherical stereo methods again mostly rely on synthetic training data, making them unsuitable for real outdoor scenes.
They assume a known, fixed camera baseline \cite{LaiXLL2019,WangSTCS2020,LiLM2021}, or estimate the relative pose between cameras \cite{WangHCLYSCS2018}.
Under the assumption of a moving camera in a static environment, structure-from-motion and multi-view stereo can be used \cite{ImHRJCK2016,HuangCCJ2017,SilveJ2019}.
However, these assumptions are violated by most usage scenarios, in which the camera might be stationary or environments are dynamic.
Crucially, these techniques do not work for a single monocular input image as information from multiple viewpoints or points in time must be combined.

%% file: 3-method.tex
\section{The 360MonoDepth framework}

Our approach builds on a general framework for estimating high-resolution depth maps from just a single monocular 360° input image.
\Cref{fig:teaser} illustrates the four main steps of our approach.
We start by projecting the 360° input image to a set of overlapping perspective tangent images (\cref{sec:tangent-images}), for instance the 20 faces of an icosahedron for an equirectangular image of resolution 2048$\times$1024\,pixels.
For each tangent image, we independently predict a depth map (\cref{sec:tangent-monodepth}) using state-of-the-art perspective monocular depth estimation \cite{RanftLHSK2021,RanftBK2021}.
Such methods predict disparity maps that are ambiguous up to affine ambiguity with unknown scale and shift \cite{YinZWNMCS2021}.
We thus formulate a global optimisation to align all tangent disparity maps in the spherical domain (\cref{sec:dispmap-alignment}).
Finally, we merge the aligned tangent disparity maps using Poisson blending \cite{PerezGB2003} into a high-resolution spherical disparity map (\cref{sec:dispmap-blending}).

In this paper, we use equirectangular projection (ERP) as the default format for spherical 360° images due to its wide adoption in the computer vision community.
However, our approach can easily be adapted to any other spherical projection by adapting the projection to/from tangent images.

\subsection{Tangent image projection}
\label{sec:tangent-images}

\begin{figure}%
	\centering
	\includegraphics[width=\linewidth]{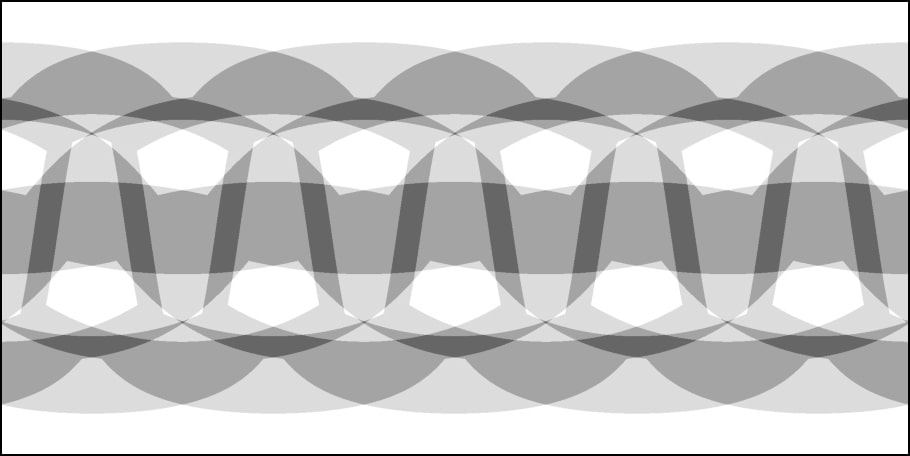}
	\caption{\label{fig:method:icosahedron_subimage}%
		Coverage of the sphere by the 20 tangent images of an icosahedron (with padding factor $p\!=\!0.3$).
		The darkest regions have an overlap of 2, the brightest of 5 images.
	}
\end{figure}

Carl Friedrich Gauss proved that any projection of a spherical image to a plane introduces some degree of distortion.
For example, equirectangular projection stretches the regions near the poles across the longitudinal dimension.
To minimise distortion, we project the spherical image to a set of perspective tangent images, each of which can be processed separately and then recombined.
We found it convenient to work with the 20 tangent images produced by the faces of an icosahedron that circumscribes a sphere, as this arrangement fairly uniformly covers the sphere's surface (see \cref{fig:method:icosahedron_subimage}), but our framework easily adapts to different numbers.
Each triangular face of the icosahedron is tangent to the sphere at its centroid, which we use to create the tangent images using gnomonic projection.

\begin{figure}[b]
	\centering
	\includegraphics[width=0.36\linewidth]{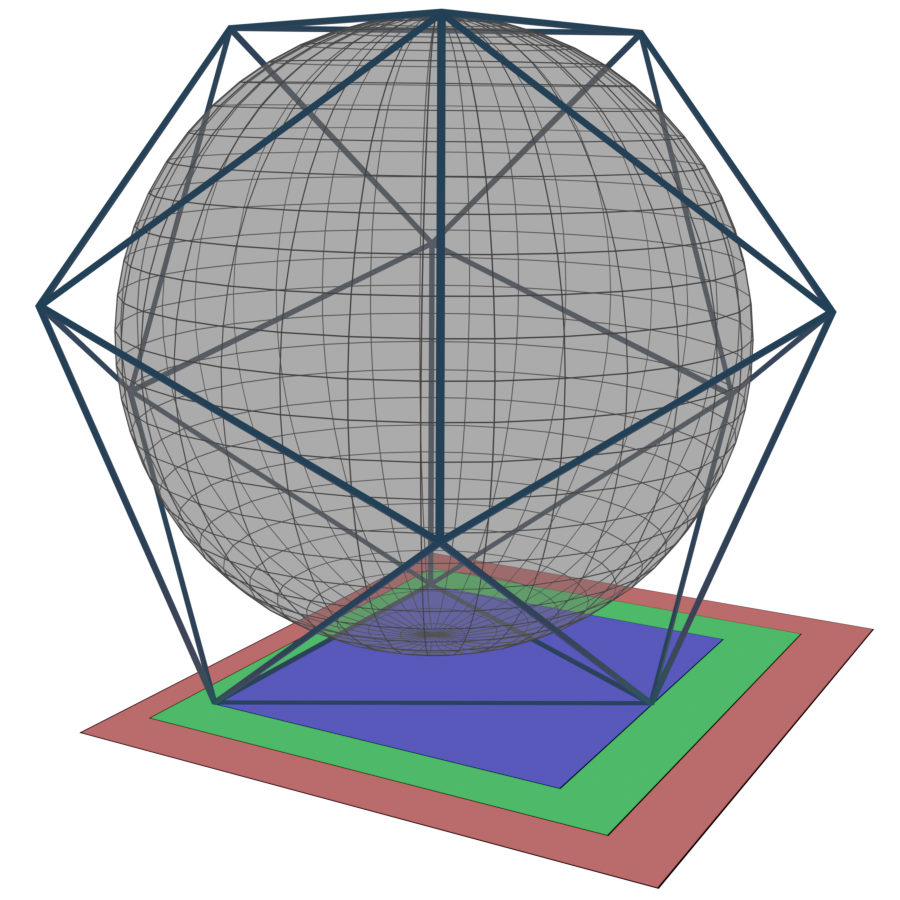}
	\includegraphics[width=0.63\linewidth]{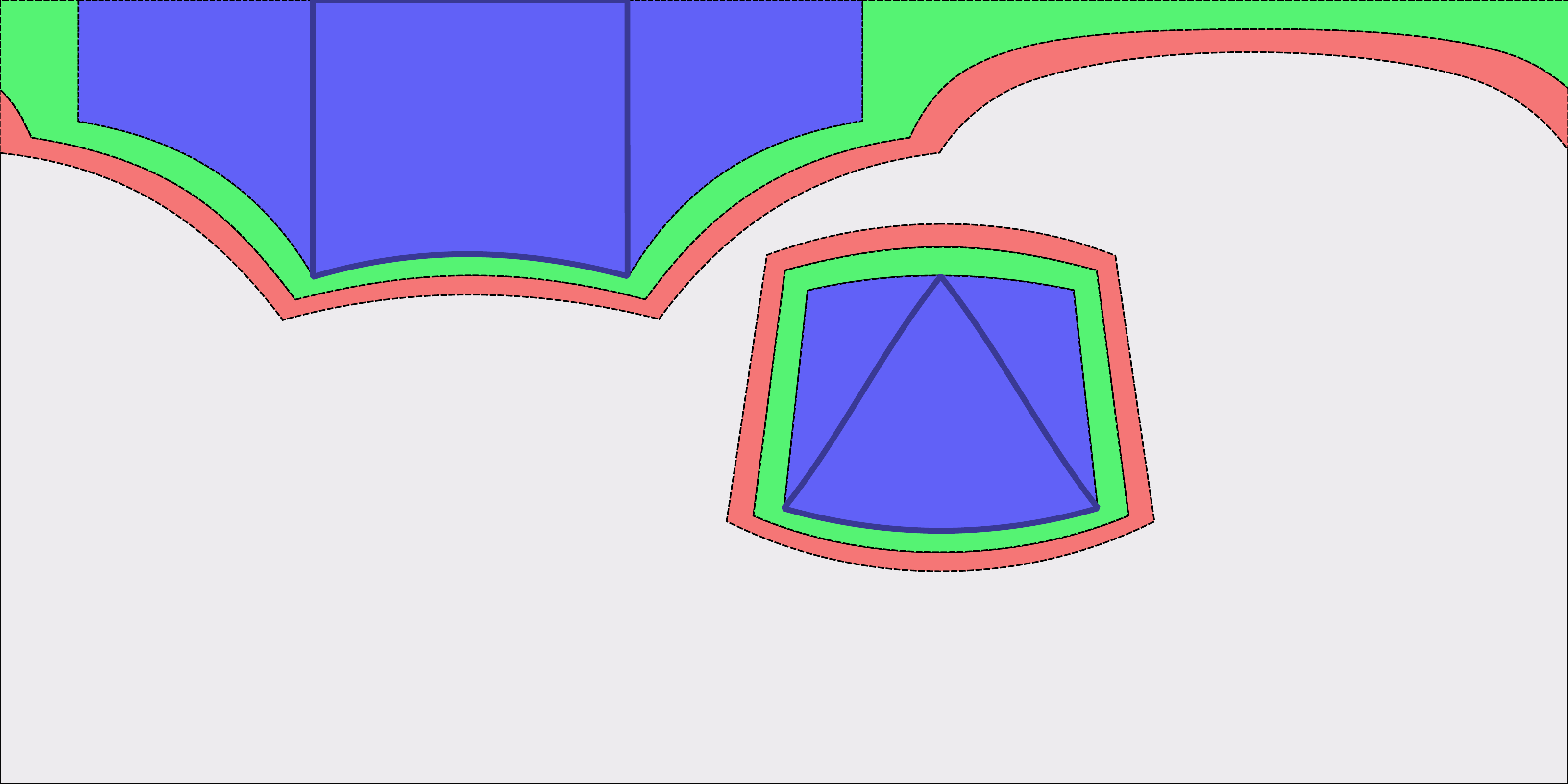}
	\caption{\label{fig:method:padding}%
		Each icosahedron face (thick triangle outline) is fit within a rectangular tangent image (blue) without padding, i.e. $p \!=\! 0$).
		The green region shows a padding of $p \!=\! 0.1$, and red shows $p \!=\! 0.2$.
		Right: Equirectangular projection for two padded tangent images.
	}
\end{figure}

\paragraph{Padding.}

By default, the size of each tangent image is constrained by the size of its icosahedron face, producing a field of view of 72°.
Tightly cropped tangent images include some overlap with adjacent icosahedron faces that share an edge, by nature of packing a triangular shape into a rectangular image (see the blue region in \cref{fig:method:padding}).
However, more overlap between tangent images, especially for icosahedron faces that only share a single vertex, is desirable for providing consistency constraints in our disparity map alignment step in \cref{sec:dispmap-alignment}, as this helps find a globally consistent alignment.
Therefore, we extend the boundaries of tangent images by a padding factor of $p \in [0, 1]$ relative to the base shape, as illustrated in \cref{fig:method:padding}.
We use a padding of $p \!=\! 0.3$, which extends the default tangent image by 30\% in all directions.

\subsection{Tangent disparity map estimation}
\label{sec:tangent-monodepth}

We use monocular depth estimation on each individual tangent image to predict dense disparity maps that will be aligned and merged in the next steps.
Specifically, we use MiDaS v2 \cite{RanftLHSK2021} and v3 \cite{RanftBK2021} for their state-of-the-art performance for both indoor and outdoor images.
Nevertheless, our framework is agnostic to the specific perspective monocular depth estimator and will benefit from future improvements.

MiDaS predicts disparity maps that correspond to inverse depth, but with an unknown scale factor and shift offset due to its scale- and shift-invariant training procedure.
Our method works consistently in disparity space, as this improves the numerical stability during the optimisation in \cref{sec:dispmap-alignment}, particularly for distant parts of the environment.

\paragraph{Perspective to spherical disparity.}

Perspective disparity maps, as predicted by MiDaS, describe disparity estimates with respect to the viewing direction of a tangent image, i.e. the $z$-component of a camera ray to a 3D point (in camera coordinates).
However, each tangent image has a different viewing direction, so the definitions of disparity are incompatible between tangent images.
In contrast, spherical disparity is the inverse (radial) Euclidean distance from the camera's centre of projection to a 3D point.
This definition is consistent for all tangent images as they all share the same centre of projection.
We convert the tangent disparity maps from perspective to spherical disparity,
and from tangent image space to the equirectangular projection of the input image in preparation for the disparity map alignment step.

\input{figures/blending-weights.tex}

\subsection{Global disparity map alignment}
\label{sec:dispmap-alignment}

The individual disparity maps $D(\cdot)$ estimated in the previous step may have inconsistent scales and offsets, as they are predicted independently from each other.
Nonetheless, each individual prediction should by design correspond to the ground-truth disparity (i.e. inverse depth) subject to a different unknown affine transform (i.e. scale and offset).
To ensure that disparity estimates are consistent with each other, we need to align them globally by finding suitable scale and offset values for each disparity map.

Our global disparity map alignment method is inspired by Hedman and Kopf's deformable depth alignment \cite{HedmaK2018}.
Instead of finding a constant scale and offset per disparity map, they use spatially varying affine adjustment fields.
These adjustment fields are modelled as 2D grids of size $m \!\times\! n$ in tangent image space.
Each grid-point $i$ stores a pair of scale and offset variables $(s^i, o^i)$ that are interpolated bilinearly across the tangent image domain.
The rescaled disparity $\tilde{D}$ of a pixel at position $\mathbf{x}$ is computed using
\begin{align}
\tilde{D}(\mathbf{x}) &= s(\mathbf{x}) D(\mathbf{x}) + o(\mathbf{x}) \text{,}
\end{align}
where $s(\mathbf{x}) = \sum_i w_i(\mathbf{x}) s^i$ and $o(\mathbf{x}) = \sum_i w_i(\mathbf{x}) o^i$ are the interpolated scale and offset values, and $w_i(\mathbf{x})$ the bilinear interpolation weights for pixel location $\mathbf{x}$.

To globally align all tangent disparity maps, we optimise for the affine adjustment fields that minimise the energy
\begin{equation}
	\argmin_{\left\{s_a^i, o_a^i\right\}}
	E_\text{alignment} + \lambda_\text{smoothness} E_\text{smoothness} + \lambda_\text{scale} E_\text{scale} \text{,}
	\label{eq:alignment_energy}
\end{equation}
which trades off alignment with the spatial smoothness of adjustment fields and a scale regularisation term.
We use $\lambda_\text{smoothness}=40$ and $\lambda_\text{scale}=0.007$ for all results.

\paragraph{Disparity alignment term.}

Once aligned, disparity maps should agree where they overlap as they represent the same region of a scene.
Given the set $\mathcal{T}$ of tangent image indices, we create the set
$\mathcal{Z} = \left\{ (a, b) \mid a, b \in \mathcal{T}, a < b\right\}$ of ordered pairs of tangent images and use $\Omega(a,b)$ to denote the set of overlapping pixels in images $a$ and $b$.
We quantify the alignment between rescaled disparity maps $\tilde{D}_a$ and $\tilde{D}_b$ using:
\vspace*{-\baselineskip}%
\begin{align}
E_\text{alignment} =
\frac{1}{z_\text{a}}
\sum_{ (a, b) \in \mathcal{Z}}
\sum_{\mathbf{x} \in \Omega(a,b)}\!\!
\left(\!\tilde{D}_a(\mathbf{x}) - \tilde{D}_b(\mathbf{x}) \!\right)^{\!2} \!\!\!\text{,}
\end{align}
where $z_\text{a} = \sum_{ (a, b) \in \mathcal{Z}} \abs{\Omega(a, b)}$ is used for normalising by the number of considered pixel pairs.
For efficiency, we only sample 1\% of pixels from the overlap regions $\Omega(a,b)$.

\paragraph{Smoothness term.}

We encourage the deformable adjustment fields to be spatially smooth between neighbouring grid-points $i$ and $j$ using
\begin{equation}
E_\text{smoothness} =
\frac{1}{z_\text{s}}
\sum_{a \in \mathcal{T}} \sum_{(i, j)} \norm{s_a^i - s_a^j}_2^2 + \norm{o_a^i - o_a^j}_2^2 \text{,}
\end{equation}
where $z_\text{s} = \abs{\mathcal{T}} \!\cdot\! m \!\cdot\! n$ normalises by the number of grid-points in all tangent images.

\paragraph{Scale term.}

The final term regularises the scale to avoid a collapse to the trivial solution of scale $s = 0$:
\begin{equation}
E_\text{scale} =
\sum_{a \in \mathcal{T}} \sum_i \left(s_a^i\right)^{-1}.
\end{equation}

\paragraph{Initialisation.}

We standardise the input spherical disparity maps to unit scale and zero offset \cite{RanftLHSK2021} using
\begin{align}
D'(\mathbf{x}) = \frac{D(\mathbf{x}) - \median(D)}{\abs{\mathcal{P}}^{-1} \sum_{\mathbf{x} \in \mathcal{P}} \abs{D(\mathbf{x}) - \median(D)}}
\end{align}
to pre-align their ranges, where $\mathcal{P}$ is the set of pixel coordinates.
Similarly, we initialise the deformation fields to unit scale $s_a^i \!=\! 1$ and zero offset $o_a^i \!=\! 0$ for all $a$ and $i$.

\subsubsection{Multi-scale deformable alignment}

Different from Hedman and Kopf, we perform deformable alignment at multiple scales, which we found to be beneficial for fine-tuning the global alignment.
We start by optimising for a coarse deformation grid of 4$\times$3 grid-points per tangent disparity map.
We then apply these deformation fields to the disparity maps, and perform a new optimisation for a 8$\times$7 grid without re-standardising the input disparity maps.
We again apply these deformation fields to the disparity maps, and perform a final refinement with a grid size of 16$\times$14.

\subsection{Disparity map blending}
\label{sec:dispmap-blending}

After the alignment, the individual disparity maps need to be merged into a single spherical disparity map, similar to how multiple photos are merged into a panorama during stitching. %
Na\"ively merging the tangent disparity maps using nearest-neighbour (`NN') or averaging per-pixel (`mean') leads to undesirable seams, as shown in \cref{fig:blending_weights}.
Using smoothly feathered blending weights \cite{YuanR2021} in the shape of a frustum reduces seams, but may produce blurrier results.

For the highest fidelity blending, we take inspiration from panorama stitching \cite{Szeli2006} and blend disparity maps in the gradient domain using Poisson blending \cite{PerezGB2003}.
Specifically, we look for the blended disparity map $B(\cdot)$ that minimises:
\begin{equation}
	\begin{aligned}
		\argmin_{B}
		\sum_{a \in \mathcal{T}}
		\sum_{\mathbf{x}}
		\omega_a(\mathbf{x}) \norm{\nabla B(\mathbf{x}) - \nabla \tilde D_a(\mathbf{x})}_2^2 \\
		+ \lambda_\text{fidelity} \cdot \sum_{\mathbf{x}} \left(B(\mathbf{x}) - D_\text{NN}(\mathbf{x})\right)^2
	\end{aligned}
	\label{eq:method:poisson}
\end{equation}
where $\omega_a(\mathbf{x})$ are the spatially varying `frustum' blending weights that modulate the influence of pixels (see \cref{fig:blending_weights}),
and $\lambda_\text{fidelity} \!=\! 0.1$ is a weight to encourage the solution to stay close to the nearest-neighbour disparity map stitch $D_\text{NN}$.

%% file: figures/blending-weights.tex
\begin{figure*}[t!]
\centering
\begin{tikzpicture}[
	image/.style={inner sep=0pt, outer sep=0pt},
	collabel/.style={above=9pt, anchor=north, inner ysep=0pt, scale=0.8, align=center},
	rowlabel/.style={left=9pt, rotate=90, anchor=north, inner ysep=0pt, scale=0.8, align=center},
	subcaption/.style={inner xsep=0.75mm, inner ysep=0.75mm, below right},
	arrow/.style={-{Latex[length=2.5mm,width=4mm]}, line width=2mm},
	spy using outlines={rectangle, size=1.85cm, magnification=4, connect spies, ultra thick, every spy on node/.append style={thick}},
	style1/.style={cyan!90!black,thick},
	style2/.style={orange!90!black},
	style3/.style={blue!90!black},
	style4/.style={red!90!black},
	]
	
	\def\padding{0.005\linewidth}
	\newcommand{\subfig}[2][0px 0px 0px 0px]{\includegraphics[width=4.2cm, trim=#1, clip]{figures/#2}}

	\node [image]                 (col1-row1) at (0,0)             {\subfig{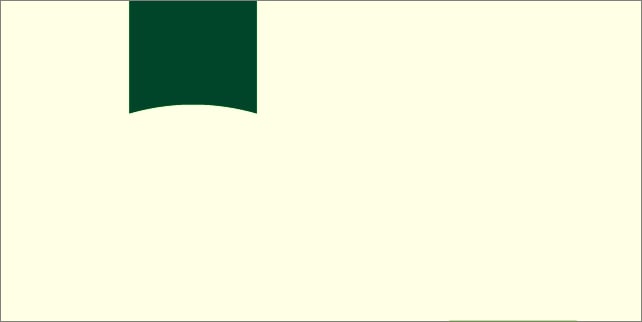}};
	\node [image,right=\padding]  (col2-row1) at (col1-row1.east)  {\subfig{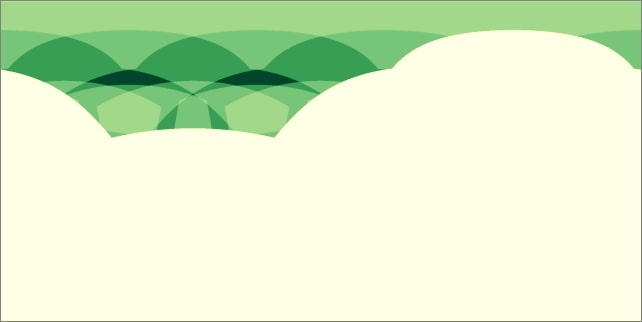}};
	\node [image,right=\padding]  (col3-row1) at (col2-row1.east)  {\subfig{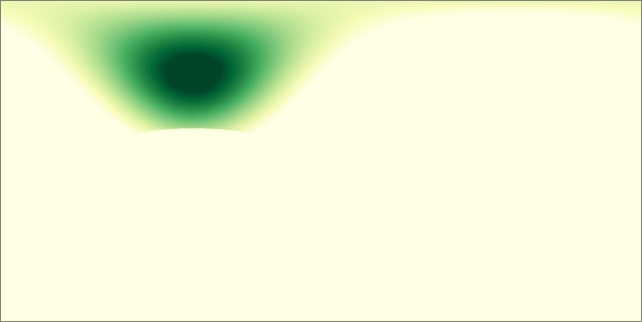}};
	\node [image,right=\padding]  (col4-row1) at (col3-row1.east)  {\subfig{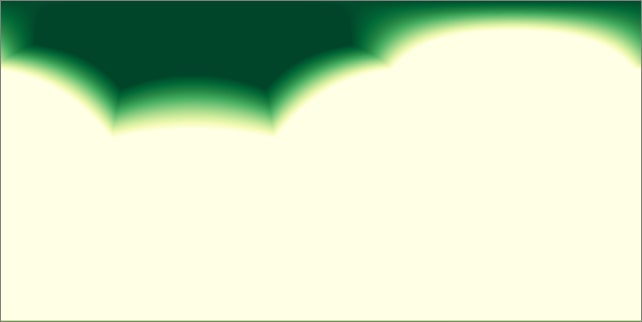}};

	\node [image,below=\padding]  (col1-row2) at (col1-row1.south) {\subfig{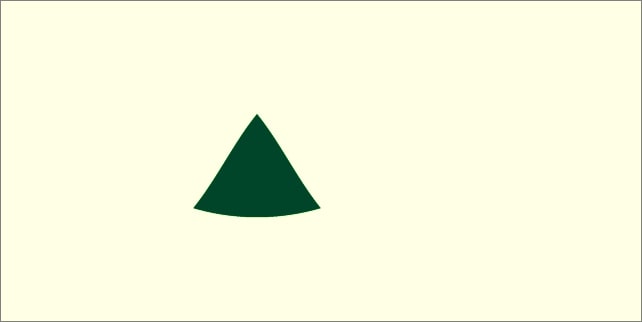}};
	\node [image,below=\padding]  (col2-row2) at (col2-row1.south) {\subfig{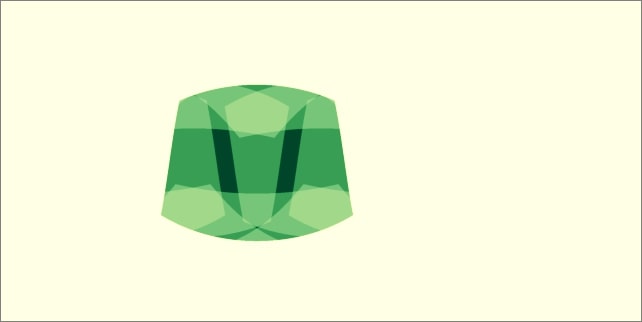}};
	\node [image,below=\padding]  (col3-row2) at (col3-row1.south) {\subfig{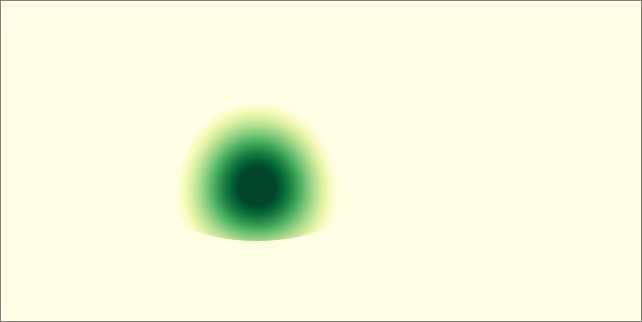}};
	\node [image,below=\padding]  (col4-row2) at (col4-row1.south) {\subfig{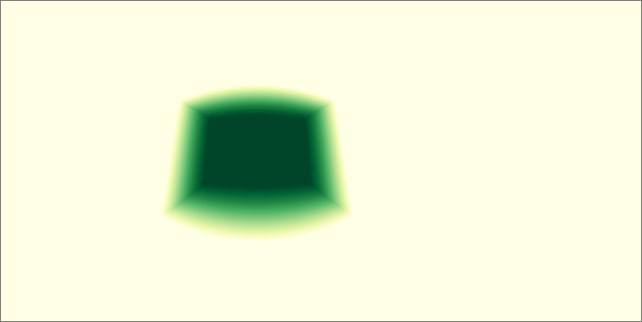}};
	
	\node [image,below=\padding]  (col1-row3) at (col1-row2.south) {\subfig[650 330 750 330]{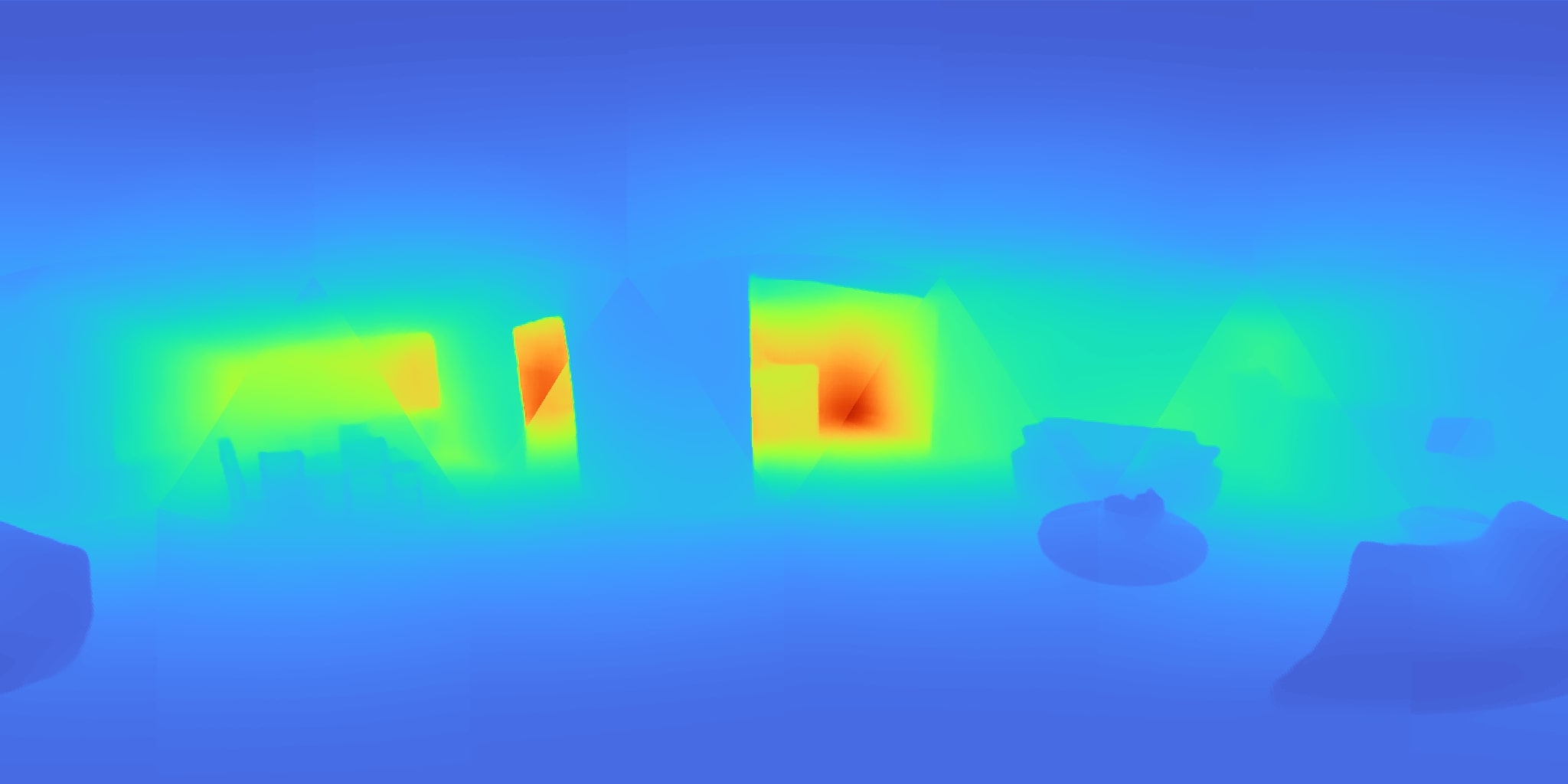}};
	\node [image,below=\padding]  (col2-row3) at (col2-row2.south) {\subfig[650 330 750 330]{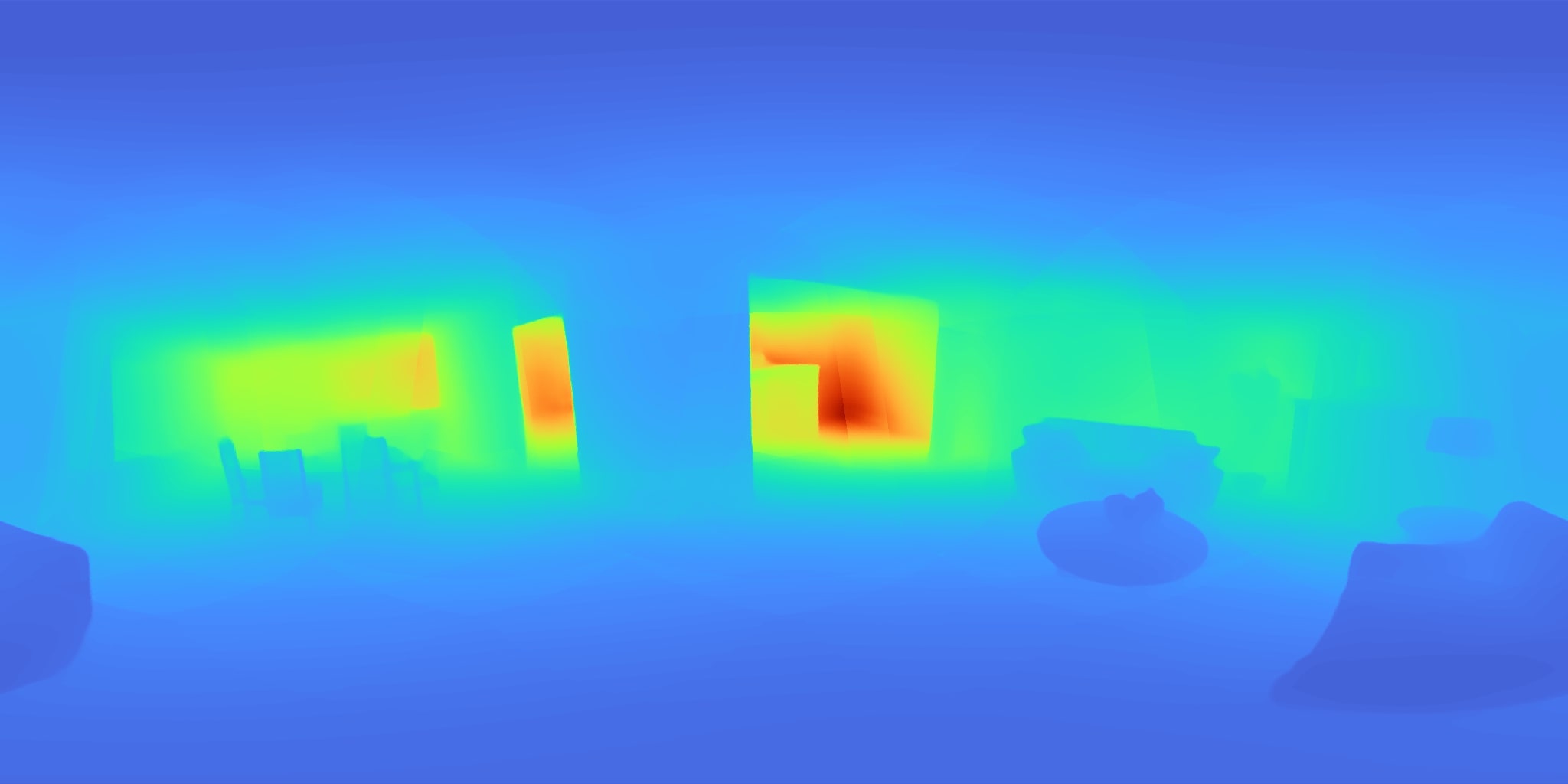}};
	\node [image,below=\padding]  (col3-row3) at (col3-row2.south) {\subfig[650 330 750 330]{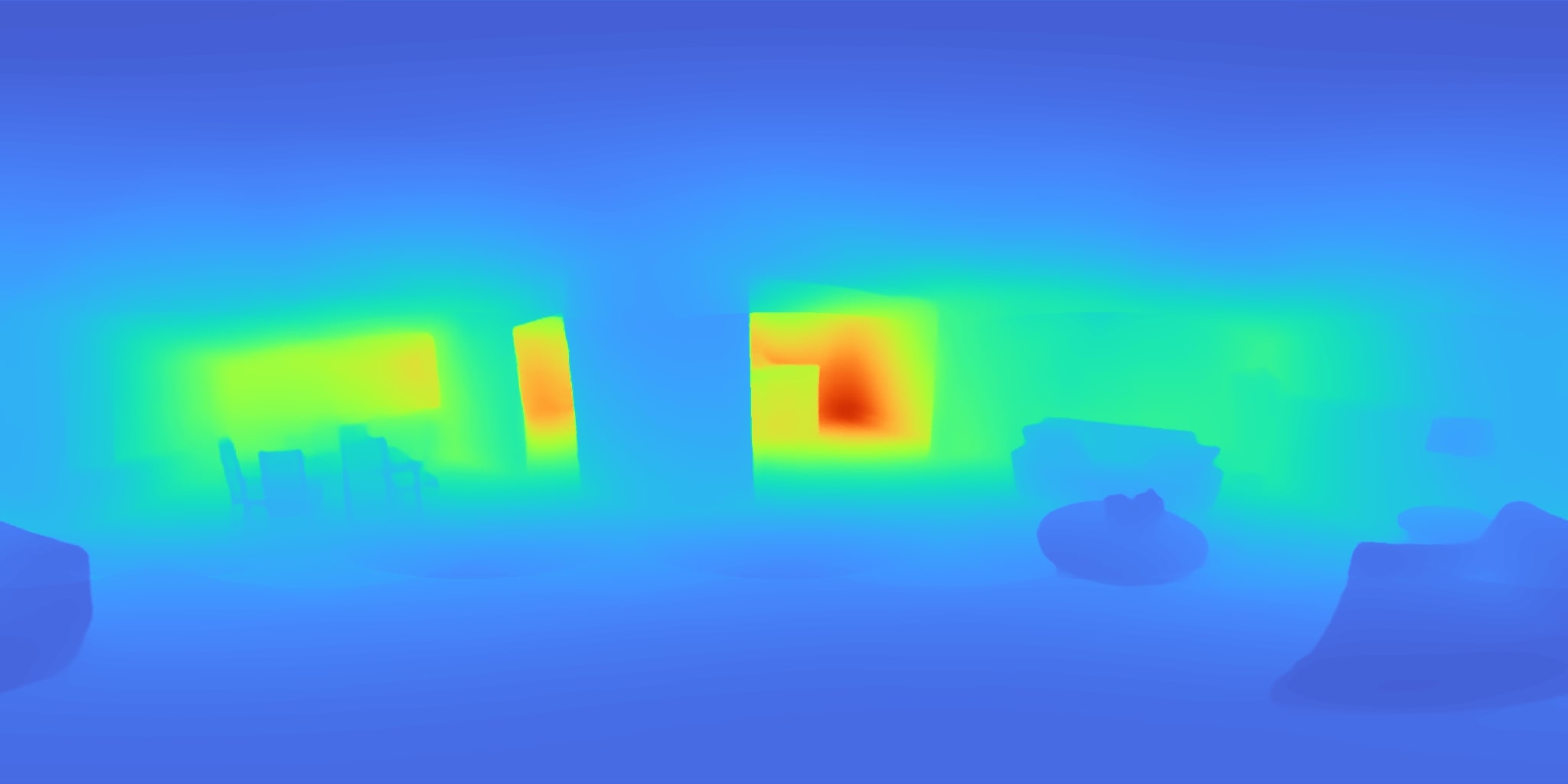}};
	\node [image,below=\padding]  (col4-row3) at (col4-row2.south) {\subfig[650 330 750 330]{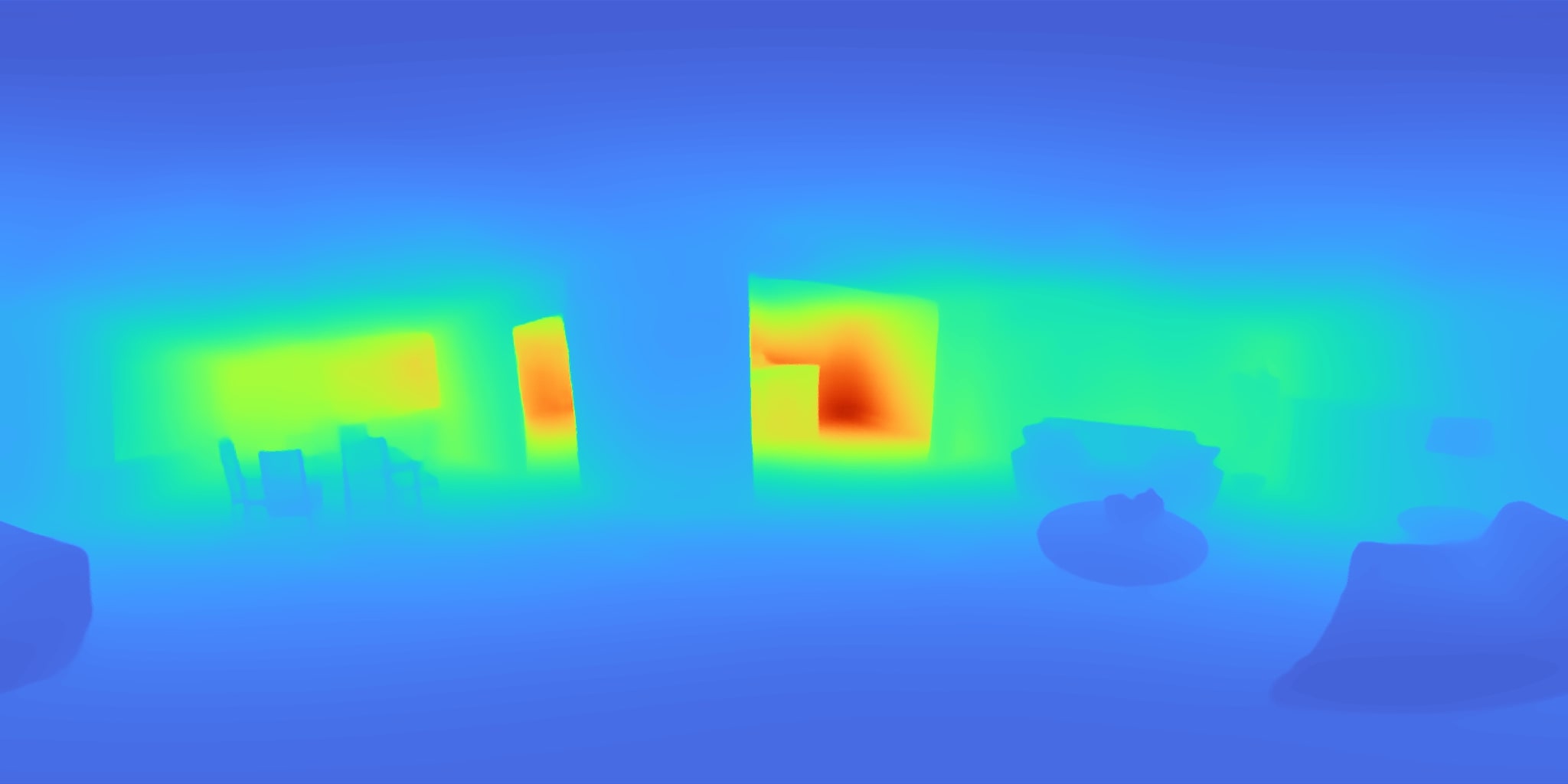}};

	\node [collabel] at (col1-row1.north) {Nearest-neighbour weights (`NN')};
	\node [collabel] at (col2-row1.north) {Mean blending weights};
	\node [collabel] at (col3-row1.north) {Radial blending weights};
	\node [collabel] at (col4-row1.north) {Frustum blending weights};
	\node [rowlabel] at (col1-row1.west) {Face 2};
	\node [rowlabel] at (col1-row2.west) {Face 13};
	\node [rowlabel] at (col1-row3.west) {Depth map (crop)};
	
	\draw[arrow,red!50!black] ($(col1-row3.center)+(-1.2,-0.5)$) -- ++(-0.4,0.4);
	\draw[arrow,red!50!black] ($(col2-row3.center)+( 0.3,-0.7)$) -- ++( 0.4,0.4);
	\draw[arrow,red!50!black] ($(col3-row3.center)+(-0.2, 0.1)$) -- ++( 0.4,0.4);

\end{tikzpicture}%
\caption{\label{fig:blending_weights}%
	Comparison of blending weights for icosahedron tangent images, in equirectangular projection.
	Vanilla tangent images \cite{EderSLF2020} select estimates only from the nearest tangent image (`NN').
	Mean weights average all overlapping tangent images per pixel.
	Radial weights start decaying at 15° from the centre of projection.
	Frustum blending weights start decaying 30\% diagonally towards the principal point off from each corner.
	Notice that disparity maps blended using `NN', `mean' and `radial' weights contain visible seams, which `frustum' minimises.
}
\end{figure*}

%% file: 4-results.tex
\section{Experiments and Results}

\paragraph{Implementation}

When processing equirectangular images at a resolution of 2048$\times$1024\,pixels, we use the 20 tangent images of a icosahedron.
We project each tangent image using a padding of $p \!=\! 0.3$ to a resolution of 400$\times$346\,pixels.
This closely matches the 384$\times$384 training resolution used by MiDaS v2/v3 \cite{RanftBK2021,RanftLHSK2021}, for which we use the authors' implementation.
We solve the global disparity map alignment problem in \cref{eq:alignment_energy} using the Ceres non-linear least-squares solver \cite{AgarwMO}.
Specifically, we perform L-BFGS line search for 50 iterations at each scale.
The gradient-based disparity map blending in \cref{eq:method:poisson} is a large sparse least-squares problem that we solve using Eigen's biconjugate gradient stabilized solver (BiCGSTAB) \cite{GuennJO2010}.
As the Matterport3D dataset \cite{ChangDFHNSSZZ2017} does not include the top and bottom regions of the scene, we exclude a circular region of radius 25° at the top and bottom from our alignment step.

\input{figures/quantitative-table-v2.tex}

\paragraph{Datasets}

For benchmarking, we use equirectangular input images and ground-truth depth maps created from the Matterport3D \cite{ChangDFHNSSZZ2017} and Replica \cite{StrauWMCWGEMRVCYBYPYZLCBGMPSBSNGLN2019} datasets.
These datasets contain indoor environments reconstructed as a textured mesh and thus provide ground-truth depth.
We also show qualitative results on varied outdoor images from OmniPhotos \cite{BerteYLR2020}, for which no ground-truth depth maps are available.

Matterport3D \cite{ChangDFHNSSZZ2017} is a real indoor dataset that comprises 10,800 panoramic images.
Unfortunately, the poses of these `skybox' images relative to the mesh reconstruction are not provided, which prevents rendering aligned ground-truth depth maps.
Previous work overcame this by rendering both images and depth maps from the textured mesh \cite{ZioulKZD2018}.
However, the image quality of these synthetic images is worse than the real skybox images, particularly at the 2048$\times$1024 resolution we are targeting.
We therefore estimate the poses for the real skybox images relative to the mesh using 360° structure-from-motion \cite{MouloMPM2016} applied to a mixture of real and rendered skybox images at known camera positions.
The estimated camera poses allow us to render ground-truth depth maps with pixel accuracy from the provided scene mesh.
From the original test split of Matterport3D with 2,014 samples, we managed to estimate accurate camera poses for 1,850 (92\%) skybox images, and rendered the aligned ground-truth depth maps at 2048$\times$1024 resolution.
We will make skybox poses and ground-truth depth maps available.

To assess the generalisation capability and scalability of our framework against baselines, we also evaluate on 360° RGBD data from the Replica dataset \cite{StrauWMCWGEMRVCYBYPYZLCBGMPSBSNGLN2019}, which features high-quality indoor room scans that have not been used for training any method.
For 13 rooms, we rendered 10 images and ground-truth depth maps at 2048$\times$1024 and 4096$\times$2048 resolution with random poses using the Replica360 renderer \cite{AttalLGRT2020}, for a total of 130 samples each.

\paragraph{Baselines}

We compare our results to 
OmniDepth \cite{ZioulKZD2018}, %
BiFuse \cite{WangYSCT2020}, %
HoHoNet \cite{SunSC2021} and %
UniFuse \cite{JiangSZDH2021} %
using the authors' public implementations and pretrained weights.
OmniDepth is trained for 512$\times$256 input, while the other methods are for 1024$\times$512.
For each method, we downscale the input images to match the expected resolution, and upsample the estimated depth map bilinearly to the input image resolution.

\input{figures/comparison-figure.tex}

\paragraph{Metrics}
We use the standard evaluation metrics adopted for monocular depth estimation evaluation \cite{EigenPF2014}.
Although our method operates in disparity space, we report metrics in depth space for fair comparisons with baselines.
Please see our supplemental document for details.

\input{figures/matterport-replica-comparison.tex}

\subsection{Quantitative evaluation}
\input{figures/omniphotos-comparison.tex}

\Cref{tab:results} shows the quantitative comparison of our method to the baselines on the Matterport3D-2K and Replica360-2K test sets.
Matterport3D is often used for training and evaluating 360° monodepth methods.
Indeed, methods trained on it (HoHoNet, UniFuse) tend to perform best.
Our method produces competitive results (in several metrics) without any training on Matterport3D, while producing depth maps at a higher resolution and level of detail (see \cref{fig:comparison-figure,fig:matterport-replica-comparison}).
Replica360 has not been used for training any method, so we can use it to measure generalisation to unseen data.
In most metrics, our approach clearly outperforms the baselines, which struggle to generalise to this new dataset.
The other two metrics, MAE and RMSE, are closely related to the L1 and BerHu (mixed L1/L2) losses used for training HoHoNet \cite{SunSC2021} and UniFuse \cite{JiangSZDH2021}, respectively, which explains these methods' better performance in these specific metrics.
We further show results at 4K resolution in \cref{tab:results-4k}.
Our results improved across all metrics compared to 2K resolution, and our approach ranks as top-2 in 6 out of 7 metrics, up from 5 out of 7 at 2K resolution (8\% improvement in MAE).
This shows that our method robustly scales to higher resolutions.

\input{figures/quantitative-table-4k.tex}

\subsection{Qualitative comparisons}\vspace{-0.3em}

We show qualitative comparisons in \cref{fig:comparison-figure}, \ref{fig:matterport-replica-comparison} and \ref{fig:omniphotos-comparison}, and our \href{https://manurare.github.io/360monodepth/demo.html}{supplemental results website}.
For datasets with available ground-truth depth maps, we show depth maps, otherwise disparity maps.
On Matterport3D, our results are mostly on par with UniFuse (best in \cref{tab:results}).
On Replica360, our results show fewer errors and cleaner surfaces.
Our approach clearly outperforms the baselines on the outdoor OmniPhotos, as no baseline is trained on outdoor data.
Our results show the highest level of detail and the sharpest depth edges.

\subsection{Ablation studies}\vspace{-0.3em}
\label{sec:ablations}

We perform two ablation studies to test our design choices in the disparity maps alignment and blending stages of our method, summarised in \cref{tab:ablations-table}.
Our multi-scale alignment and Poisson blending approaches outperform other alternatives.
In particular, our alignment step substantially outperforms the “No alignment” of Eder et al. \cite{EderSLF2020} across all metrics.
Both deformable multi-scale alignment and blending are necessary for the best results.

\input{figures/ablations-table.tex}

%% file: figures/quantitative-table-v2.tex
\begin{table*}
\newcommand{\ul}{\color{blue!70!black}\bf}
\caption{\label{tab:results}%
	Quantitative results for Matterport3D-2K and Replica360-2K, at 2048$\times$1024 with Poisson blending.
	Highlighting: {\ul best}, \textbf{second-best}.
}
\renewcommand*{\arraystretch}{1.15}%
\setlength{\tabcolsep}{2pt}%
\vspace{-0.2em}\centering%
\resizebox{\linewidth}{!}{%
	\begin{tabular}{l@{\hspace{10pt}}ccccccc@{\hspace{14pt}}ccccccc}
		& \multicolumn{7}{c@{\hspace{14pt}}}{Matterport3D-2K} & \multicolumn{7}{c}{Replica360-2K} \\
		\toprule
		Method &
		\scriptsize AbsRel\tridown &
		\scriptsize MAE\tridown &
		\scriptsize RMSE\tridown &
		\scriptsize RMSE-log\tridown &
		\scriptsize $\delta \!<$1.25\triup &
		\scriptsize $\delta \!<$1.25\textsuperscript{2}\triup &
		\scriptsize $\delta \!<$1.25\textsuperscript{3}\triup &
		\scriptsize AbsRel\tridown &
		\scriptsize MAE\tridown &
		\scriptsize RMSE\tridown &
		\scriptsize RMSE-log\tridown &
		\scriptsize $\delta \!<$1.25\triup &
		\scriptsize $\delta \!<$1.25\textsuperscript{2}\triup &
		\scriptsize $\delta \!<$1.25\textsuperscript{3}\triup
		\\ \midrule
		OmniDepth \cite{ZioulKZD2018}
		&     0.473 &     0.946 &     1.317 &     0.212 &     0.378 &     0.647 &     0.820  %
		&     0.352 &     0.589 &     0.787 &     0.168 &     0.479 &     0.776 &     0.906  %
		\\
		BiFuse \cite{WangYSCT2020}
		&     0.321 &     0.649 &     0.994 &     0.158 &     0.564 &     0.802 &     0.910  %
		&     0.318 &     0.468 &     0.663 &     0.152 &     0.591 &     0.840 &     0.927  %
		\\
		HoHoNet\MData \cite{SunSC2021}
		&     0.227 & \bf 0.430 & \bf 0.686 &     0.132 & \bf 0.723 &     0.887 &     0.946  %
		&     0.259 &     0.381 &     0.520 &     0.131 &     0.672 &     0.888 &     0.942  %
		\\
		HoHoNet\SData \cite{SunSC2021}
		&     0.234 &     0.487 &     0.736 &     0.120 &     0.654 &     0.886 & \bf 0.959  %
		&     0.221 & \bf 0.355 & \bf 0.480 &     0.112 &     0.701 &     0.905 &     0.960  %
		\\
		UniFuse\MData \cite{JiangSZDH2021}
		& \ul 0.200 & \ul 0.396 & \ul 0.652 & \ul 0.113 & \ul 0.769 & \ul 0.908 &     0.958  %
		&     0.233 & \ul 0.330 & \ul 0.474 &     0.120 &     0.728 &     0.905 &     0.954  %
		\\ \midrule
		Ours\Mvii (single-scale)  %
		&     0.223 &     0.491 &     0.828 &     0.129 &     0.619 &     0.867 &     0.953  %
		& \bf 0.182 &     0.412 &     0.732 & \bf 0.095 & \bf 0.750 & \bf 0.935 & \bf 0.971  %
		\\
		Ours\Mviii (single-scale)  %
		&     0.210 &     0.476 &     0.840 &     0.121 &     0.656 &     0.889 &     0.958  %
		&     0.192 &     0.447 &     0.805 &     0.100 &     0.737 &     0.925 &     0.969  %
		\\
		Ours\Mvii (multi-scale)  %
		&     0.224 &     0.494 &     0.831 &     0.130 &     0.616 &     0.866 &     0.953  %
		& \ul 0.167 &     0.364 &     0.619 & \ul 0.089 & \ul 0.769 & \ul 0.948 & \ul 0.981  %
		\\
		Ours\Mviii (multi-scale)  %
		& \bf 0.208 &     0.446 &     0.791 & \bf 0.119 &     0.656 & \bf 0.890 & \ul 0.961  %
		&     0.198 &     0.465 &     0.841 &     0.103 &     0.730 &     0.920 &     0.965  %
		\\
		\bottomrule
	\end{tabular}%
}\\[1mm]
\footnotesize%
\MData~Trained on Matterport3D \cite{ChangDFHNSSZZ2017}
\qquad\SData~Trained on Stanford 2D-3D-S \cite{ArmenSZS2017}
\qquad\Mvii~Using MiDaS v2 \cite{RanftLHSK2021}
\qquad\Mviii~Using MiDaS v3 \cite{RanftBK2021}
\end{table*}

%% file: figures/comparison-figure.tex
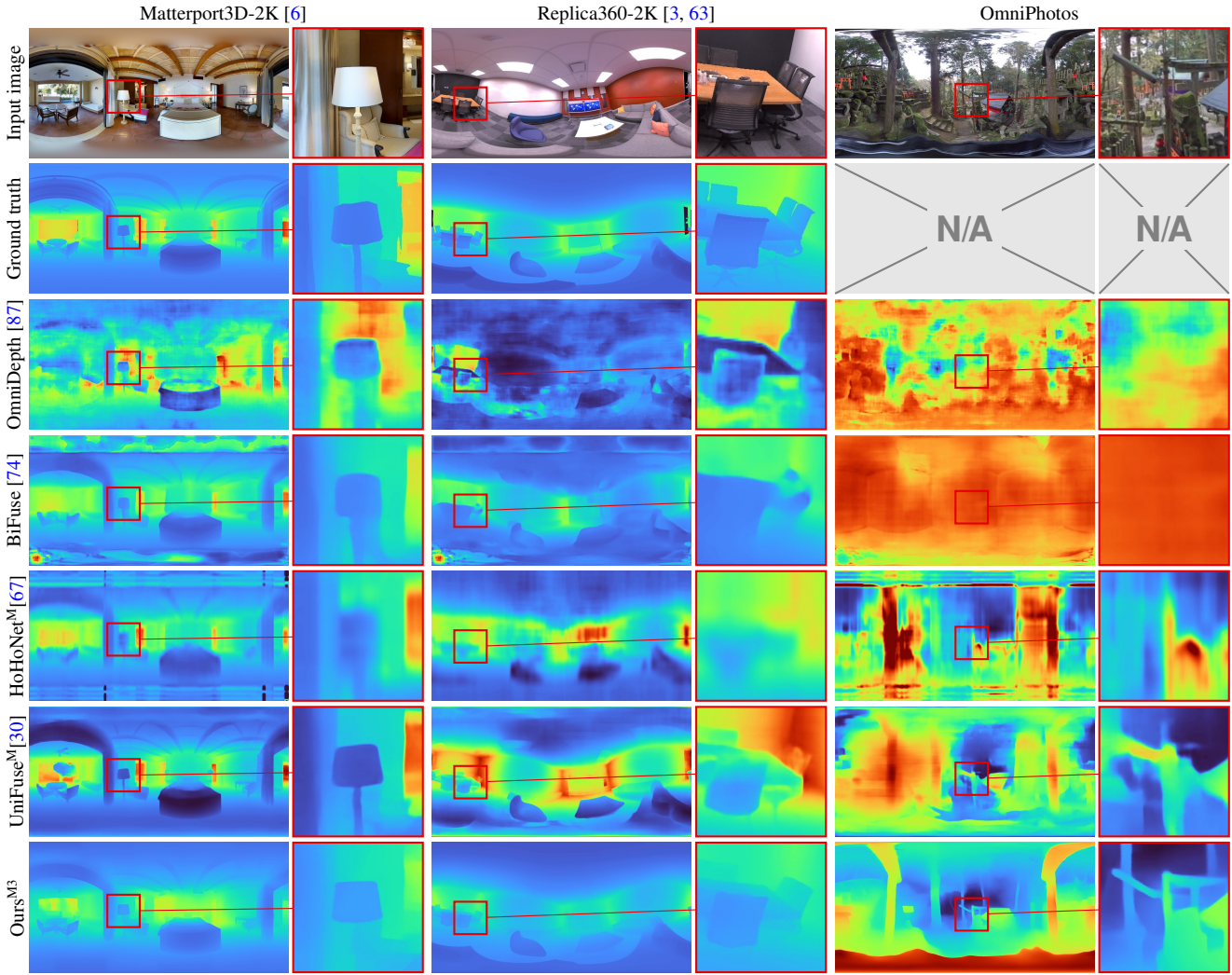
\begin{figure*}[tp]
\centering
\begin{tikzpicture}[
	image/.style={inner sep=0pt, outer sep=0pt},
	collabel/.style={above=9pt, anchor=north, inner ysep=0pt, scale=0.8, align=center},
	rowlabel/.style={left=9pt, rotate=90, anchor=north, inner ysep=0pt, scale=0.8, align=center},
	subcaption/.style={inner xsep=0.75mm, inner ysep=0.75mm, below right},
	arrow/.style={-{Latex[length=2.5mm,width=4mm]}, line width=2mm},
	spy using outlines={rectangle, size=1.85cm, magnification=4, connect spies, ultra thick, every spy on node/.append style={thick}},
	style1/.style={cyan!90!black,thick},
	style2/.style={orange!90!black},
	style3/.style={blue!90!black},
	style4/.style={red!90!black},
	]
	
	\def\padding{0.005\linewidth}
	\def\matterportimage{481}
	\def\replicaimage{001}
	\def\omniphotosimage{009}
	\newcommand{\subfig}[2][0px 0px 0px 0px]{\includegraphics[height=1.85cm, trim=#1, clip]{figures/#2}}

	\node [image]                 (col1-row1) at (0,0)             {\subfig{matterport/\matterportimage_rgb}};
	\node [image,right=2.05cm]    (col2-row1) at (col1-row1.east)  {\subfig{replica/\replicaimage_rgb}};
	\node [image,right=2.05cm]    (col3-row1) at (col2-row1.east)  {\subfig{omniphotos/\omniphotosimage_rgb}};
	
	\node [image,below=\padding]  (col1-row2) at (col1-row1.south) {\subfig{matterport/\matterportimage_GT}};
	\node [image,below=\padding]  (col2-row2) at (col2-row1.south) {\subfig{replica/\replicaimage_GT}};
	\node [image,below=\padding]  (col3-row2) at (col3-row1.south) {\textcolor{black!10}{\rule{3.7cm}{1.85cm}}};
	\draw[thick,black!50] (col3-row2.north west) -- (col3-row2.south east);
	\draw[thick,black!50] (col3-row2.north east) -- (col3-row2.south west);
	\node[fill=black!10] at (col3-row2.center) {\Large \textcolor{black!50}{\textsf{\textbf{N/\hspace{-1.5pt}A}}}};
	
	\node [image,below=\padding]  (col1-row3) at (col1-row2.south) {\subfig{matterport/\matterportimage_omnidepth}};
	\node [image,below=\padding]  (col2-row3) at (col2-row2.south) {\subfig{replica/\replicaimage_omnidepth}};
	\node [image,below=\padding]  (col3-row3) at (col3-row2.south) {\subfig{omniphotos/\omniphotosimage_omnidepth}};
	
	\node [image,below=\padding]  (col1-row4) at (col1-row3.south) {\subfig{matterport/\matterportimage_bifuse}};
	\node [image,below=\padding]  (col2-row4) at (col2-row3.south) {\subfig{replica/\replicaimage_bifuse}};
	\node [image,below=\padding]  (col3-row4) at (col3-row3.south) {\subfig{omniphotos/\omniphotosimage_bifuse}};
	
	\node [image,below=\padding]  (col1-row5) at (col1-row4.south) {\subfig{matterport/\matterportimage_hohonet}};
	\node [image,below=\padding]  (col2-row5) at (col2-row4.south) {\subfig{replica/\replicaimage_hohonet}};
	\node [image,below=\padding]  (col3-row5) at (col3-row4.south) {\subfig{omniphotos/\omniphotosimage_hohonet}};
	
	\node [image,below=\padding]  (col1-row6) at (col1-row5.south) {\subfig{matterport/\matterportimage_unifuse}};
	\node [image,below=\padding]  (col2-row6) at (col2-row5.south) {\subfig{replica/\replicaimage_unifuse}};
	\node [image,below=\padding]  (col3-row6) at (col3-row5.south) {\subfig{omniphotos/\omniphotosimage_unifuse}};
	
	\node [image,below=\padding]  (col1-row7) at (col1-row6.south) {\subfig{matterport/\matterportimage_360monodepth_poisson}};
	\node [image,below=\padding]  (col2-row7) at (col2-row6.south) {\subfig{replica/\replicaimage_360monodepth_poisson}};
	\node [image,below=\padding]  (col3-row7) at (col3-row6.south) {\subfig{omniphotos/\omniphotosimage_360monodepth_poisson_midas3}};

	\spy[style4] on ($(col1-row1.center)+(-0.5,-0.05)$) in node (crop-col1-row1) [anchor=west] at ($(col1-row1.east)+(2pt,0)$);
	\spy[style4] on ($(col1-row2.center)+(-0.5,-0.05)$) in node (crop-col1-row2) [anchor=west] at ($(col1-row2.east)+(2pt,0)$);
	\spy[style4] on ($(col1-row3.center)+(-0.5,-0.05)$) in node (crop-col1-row3) [anchor=west] at ($(col1-row3.east)+(2pt,0)$);
	\spy[style4] on ($(col1-row4.center)+(-0.5,-0.05)$) in node (crop-col1-row4) [anchor=west] at ($(col1-row4.east)+(2pt,0)$);
	\spy[style4] on ($(col1-row5.center)+(-0.5,-0.05)$) in node (crop-col1-row5) [anchor=west] at ($(col1-row5.east)+(2pt,0)$);
	\spy[style4] on ($(col1-row6.center)+(-0.5,-0.05)$) in node (crop-col1-row6) [anchor=west] at ($(col1-row6.east)+(2pt,0)$);
	\spy[style4] on ($(col1-row7.center)+(-0.5,-0.05)$) in node (crop-col1-row7) [anchor=west] at ($(col1-row7.east)+(2pt,0)$);
	
	\spy[style4] on ($(col2-row1.center)+(-1.3,-0.15)$) in node (crop-col2-row1) [anchor=west] at ($(col2-row1.east)+(2pt,0)$);
	\spy[style4] on ($(col2-row2.center)+(-1.3,-0.15)$) in node (crop-col2-row2) [anchor=west] at ($(col2-row2.east)+(2pt,0)$);
	\spy[style4] on ($(col2-row3.center)+(-1.3,-0.15)$) in node (crop-col2-row3) [anchor=west] at ($(col2-row3.east)+(2pt,0)$);
	\spy[style4] on ($(col2-row4.center)+(-1.3,-0.15)$) in node (crop-col2-row4) [anchor=west] at ($(col2-row4.east)+(2pt,0)$);
	\spy[style4] on ($(col2-row5.center)+(-1.3,-0.15)$) in node (crop-col2-row5) [anchor=west] at ($(col2-row5.east)+(2pt,0)$);
	\spy[style4] on ($(col2-row6.center)+(-1.3,-0.15)$) in node (crop-col2-row6) [anchor=west] at ($(col2-row6.east)+(2pt,0)$);
	\spy[style4] on ($(col2-row7.center)+(-1.3,-0.15)$) in node (crop-col2-row7) [anchor=west] at ($(col2-row7.east)+(2pt,0)$);
	
	\spy[style4] on ($(col3-row1.center)+(0.1,-0.1)$) in node (crop-col3-row1) [anchor=west] at ($(col3-row1.east)+(2pt,0)$);
	\node[image,right] (crop-col3-row2) at ($(col3-row2.east)+(2pt,0)$) {\textcolor{black!10}{\rule{1.85cm}{1.85cm}}};
	\draw[thick,black!50] (crop-col3-row2.north west) -- (crop-col3-row2.south east);
	\draw[thick,black!50] (crop-col3-row2.north east) -- (crop-col3-row2.south west);
	\node[fill=black!10] at (crop-col3-row2.center) {\Large \textcolor{black!50}{\textsf{\textbf{N/\hspace{-1.5pt}A}}}};
	\spy[style4] on ($(col3-row3.center)+(0.1,-0.1)$) in node (crop-col3-row3) [anchor=west] at ($(col3-row3.east)+(2pt,0)$);
	\spy[style4] on ($(col3-row4.center)+(0.1,-0.1)$) in node (crop-col3-row4) [anchor=west] at ($(col3-row4.east)+(2pt,0)$);
	\spy[style4] on ($(col3-row5.center)+(0.1,-0.1)$) in node (crop-col3-row5) [anchor=west] at ($(col3-row5.east)+(2pt,0)$);
	\spy[style4] on ($(col3-row6.center)+(0.1,-0.1)$) in node (crop-col3-row6) [anchor=west] at ($(col3-row6.east)+(2pt,0)$);
	\spy[style4] on ($(col3-row7.center)+(0.1,-0.1)$) in node (crop-col3-row7) [anchor=west] at ($(col3-row7.east)+(2pt,0)$);
	
	\node [collabel] at ($(col1-row1.north west)!0.75!(col1-row1.north east)$) {Matterport3D-2K \cite{ChangDFHNSSZZ2017}};
	\node [collabel] at ($(col2-row1.north west)!0.75!(col2-row1.north east)$) {Replica360-2K \cite{StrauWMCWGEMRVCYBYPYZLCBGMPSBSNGLN2019,AttalLGRT2020}};
	\node [collabel] at ($(col3-row1.north west)!0.75!(col3-row1.north east)$) {OmniPhotos};
	\node [rowlabel] at (col1-row1.west) {Input image};
	\node [rowlabel] at (col1-row2.west) {Ground truth};
	\node [rowlabel] at (col1-row3.west) {OmniDepth \cite{ZioulKZD2018}};
	\node [rowlabel] at (col1-row4.west) {BiFuse \cite{WangYSCT2020}};
	\node [rowlabel] at (col1-row5.west) {HoHoNet\MData \cite{SunSC2021}};
	\node [rowlabel] at (col1-row6.west) {UniFuse\MData \cite{JiangSZDH2021}};
	\node [rowlabel] at (col1-row7.west) {Ours\Mviii};

\end{tikzpicture}%
	\caption{\label{fig:comparison-figure}%
		Qualitative comparison to different methods on different datasets.
		Our results show the highest level of detail of all predictions.
	}
\end{figure*}

%% file: figures/matterport-replica-comparison.tex
\begin{figure*}%
\centering
\begin{tikzpicture}[
	image/.style={inner sep=0pt, outer sep=0pt},
	collabel/.style={above=9pt, anchor=north, inner ysep=0pt, scale=0.8, align=center},
	rowlabel/.style={left=9pt, rotate=90, anchor=north, inner ysep=0pt, scale=0.8, align=center},
	subcaption/.style={inner xsep=0.75mm, inner ysep=0.75mm, below right},
	arrow/.style={-{Latex[length=2.5mm,width=4mm]}, line width=2mm},
	spy using outlines={rectangle, size=1.85cm, magnification=4, connect spies, ultra thick, every spy on node/.append style={thick}},
	style1/.style={cyan!90!black,thick},
	style2/.style={orange!90!black},
	style3/.style={blue!90!black},
	style4/.style={red!90!black},
	]
	
	\def\padding{0.005\linewidth}
	\newcommand{\subfig}[2][0px 0px 0px 0px]{\includegraphics[height=2.1cm, trim=#1, clip]{figures/#2}}

	\node [image]                 (col1-row1) at (0,0)             {\subfig{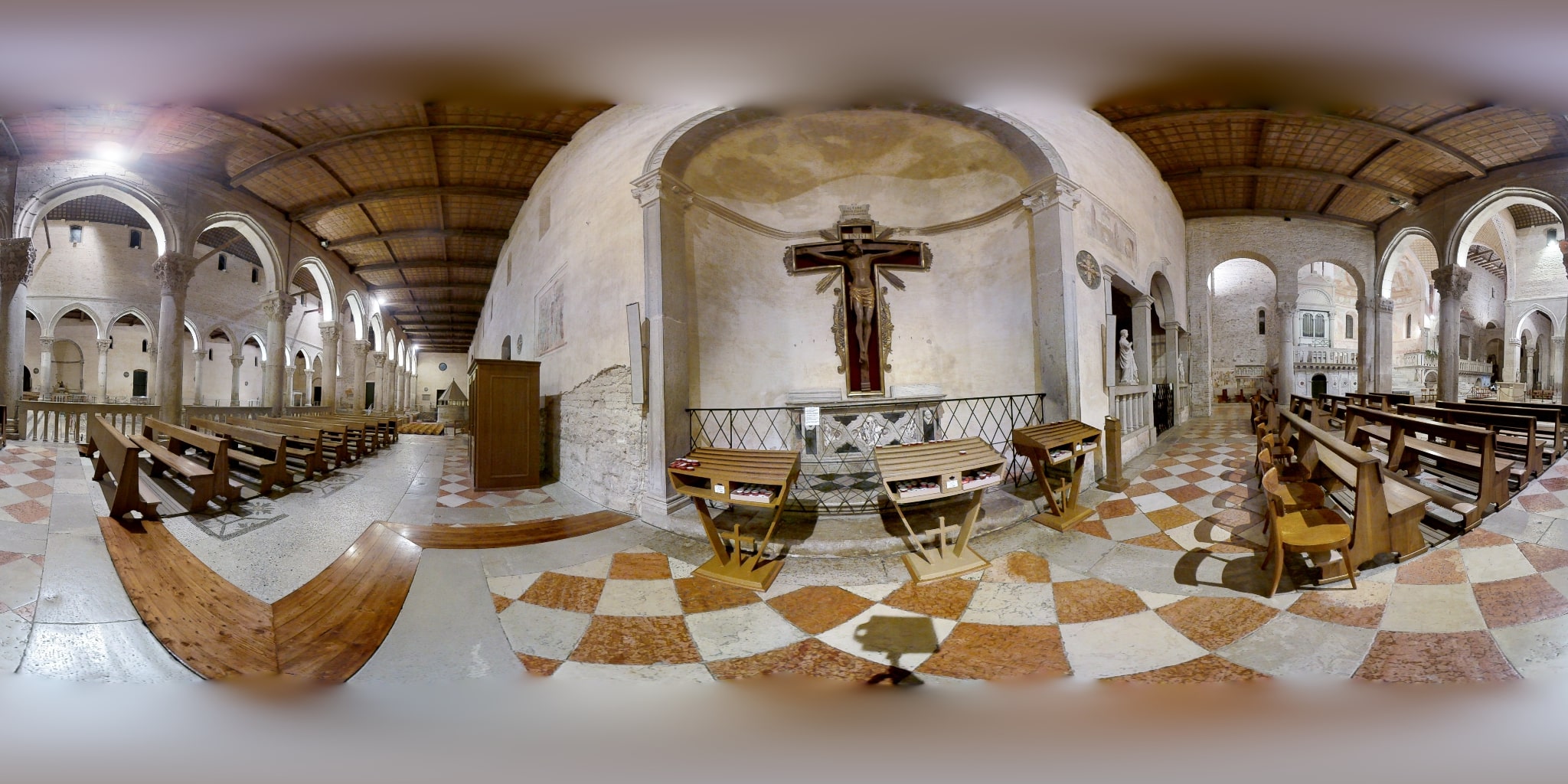}};
	\node [image,right=\padding]  (col2-row1) at (col1-row1.east)  {\subfig{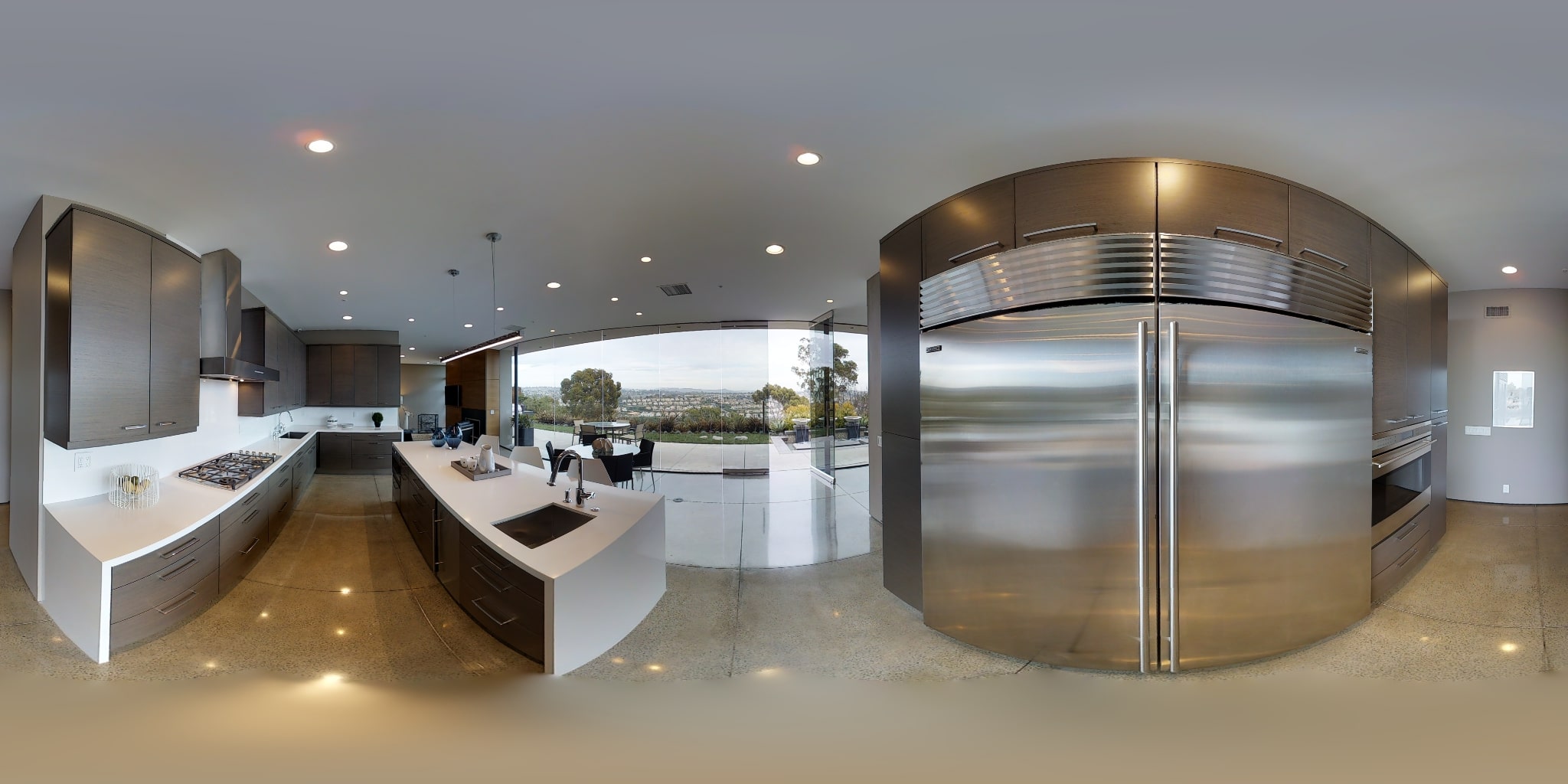}};
	\node [image,right=\padding]  (col3-row1) at (col2-row1.east)  {\subfig{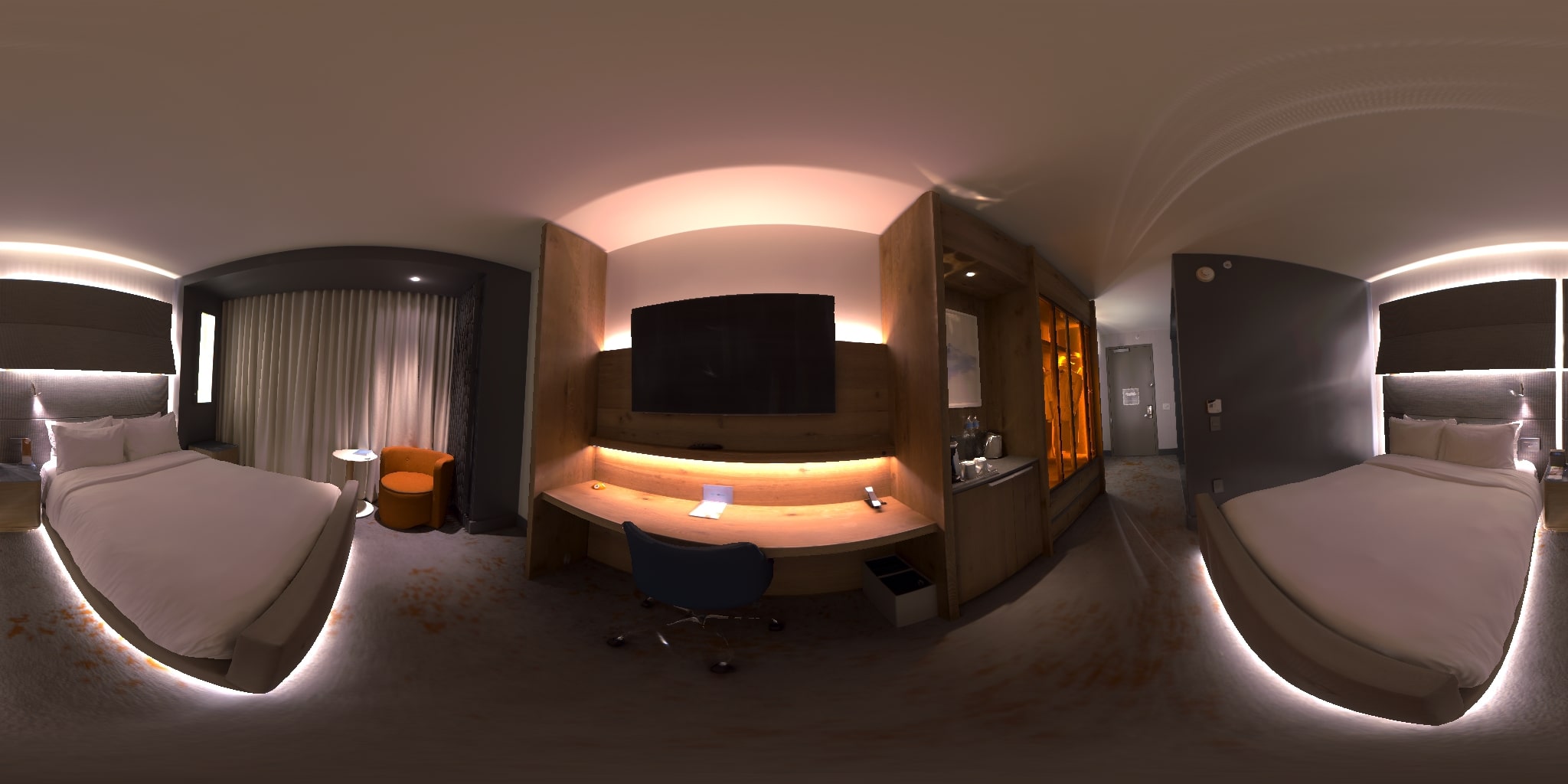}};
	\node [image,right=\padding]  (col4-row1) at (col3-row1.east)  {\subfig{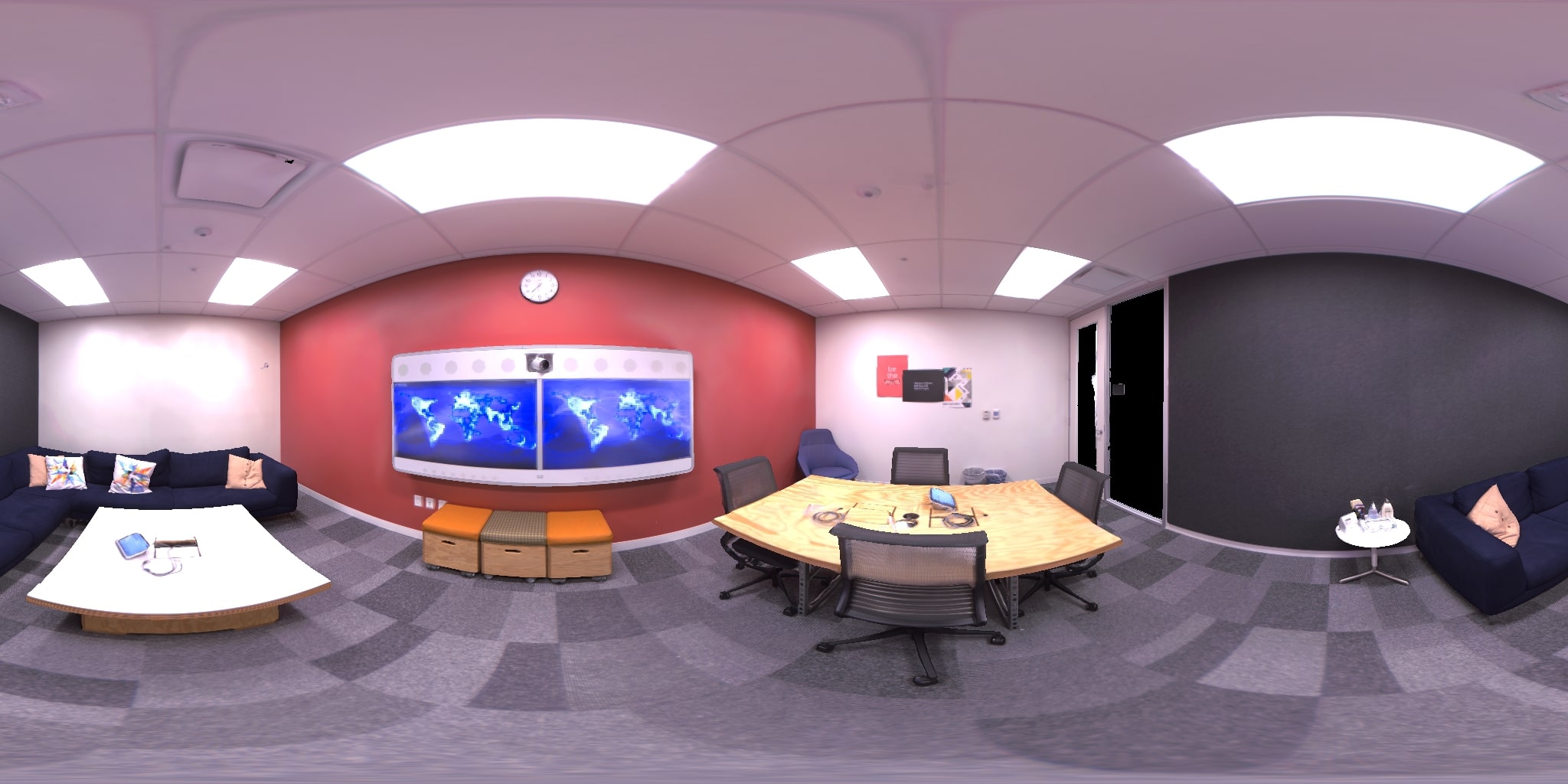}};

	\node [image,below=\padding]  (col1-row2) at (col1-row1.south) {\subfig{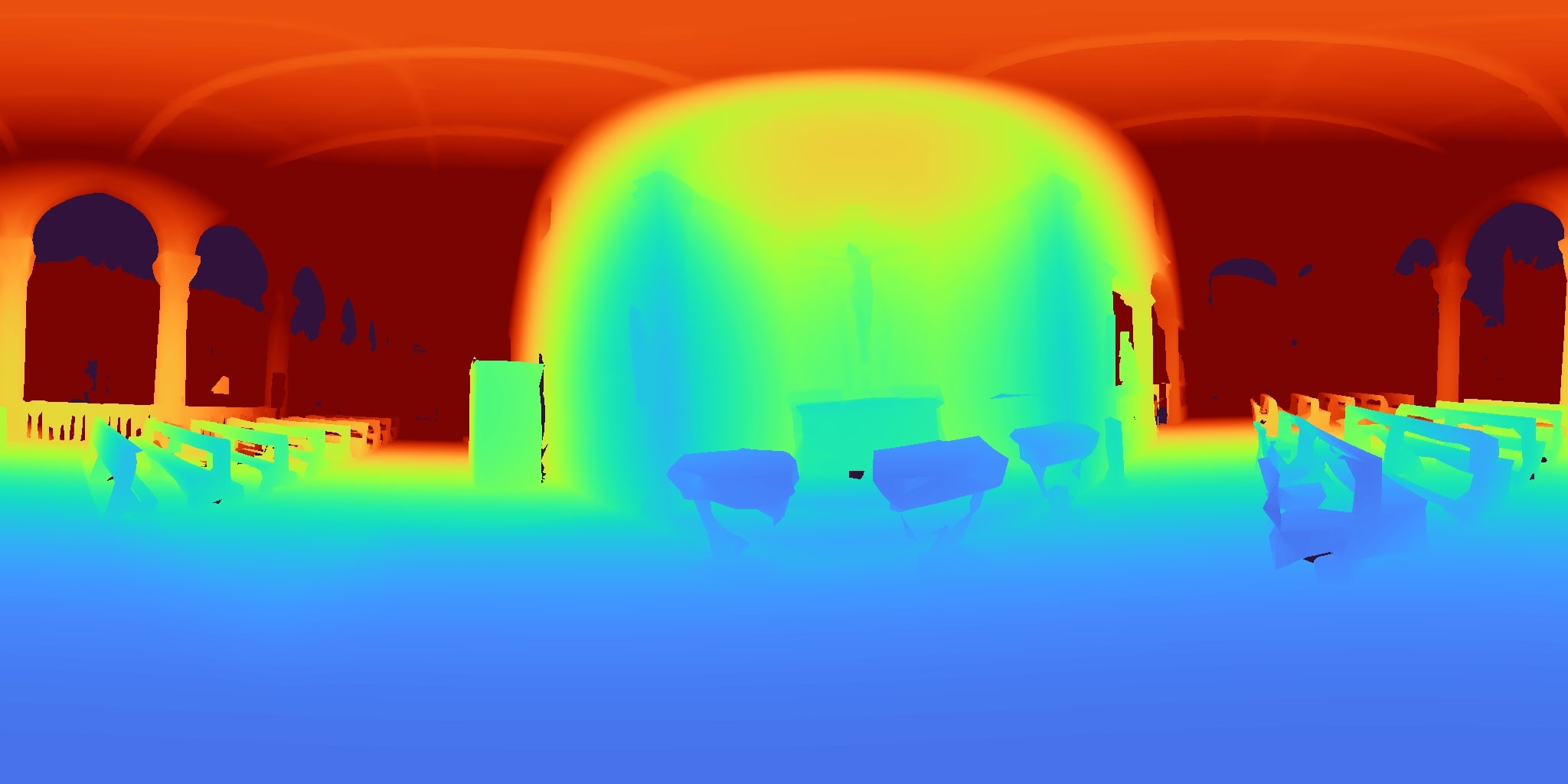}};
	\node [image,below=\padding]  (col2-row2) at (col2-row1.south) {\subfig{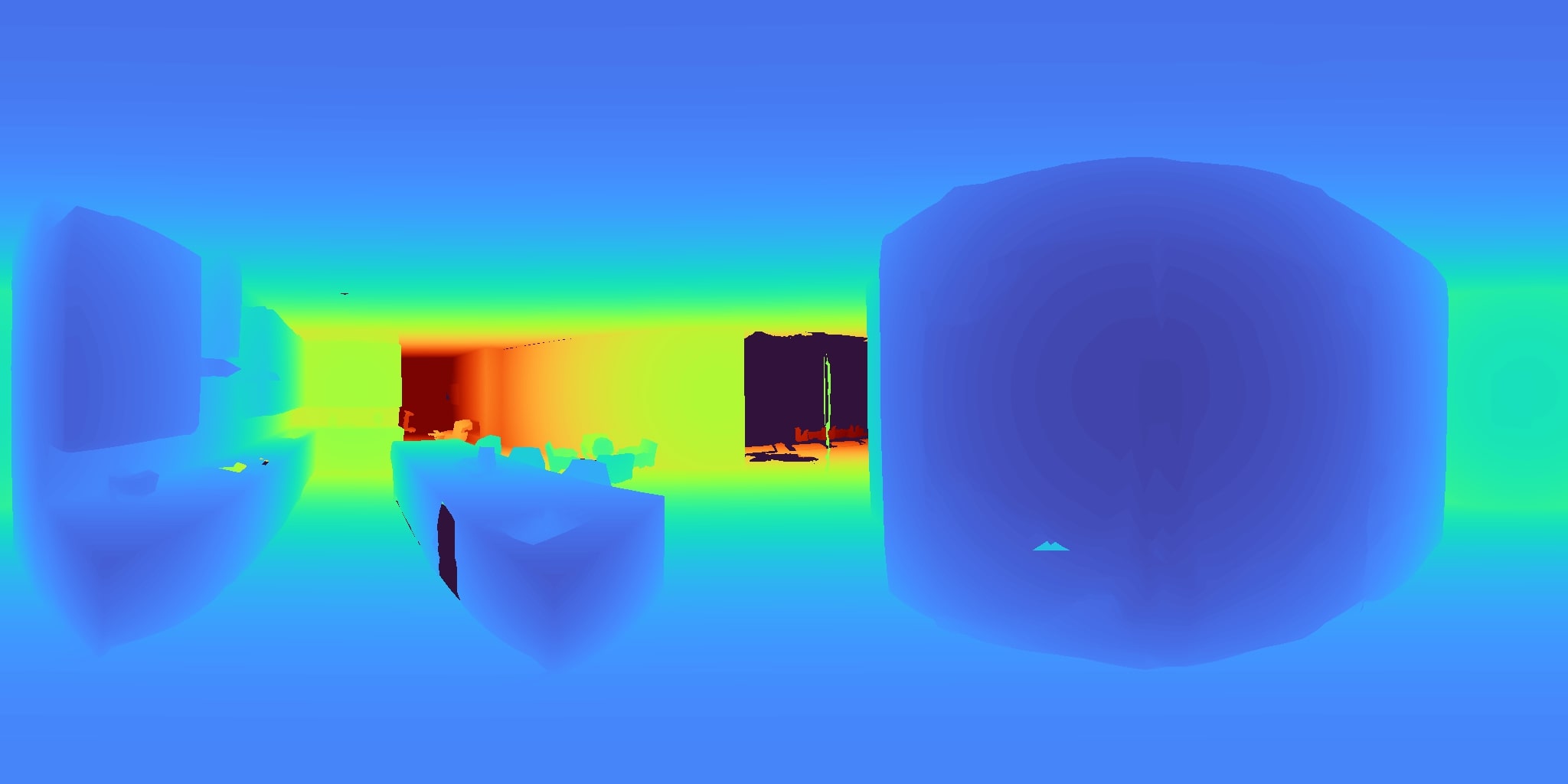}};
	\node [image,below=\padding]  (col3-row2) at (col3-row1.south) {\subfig{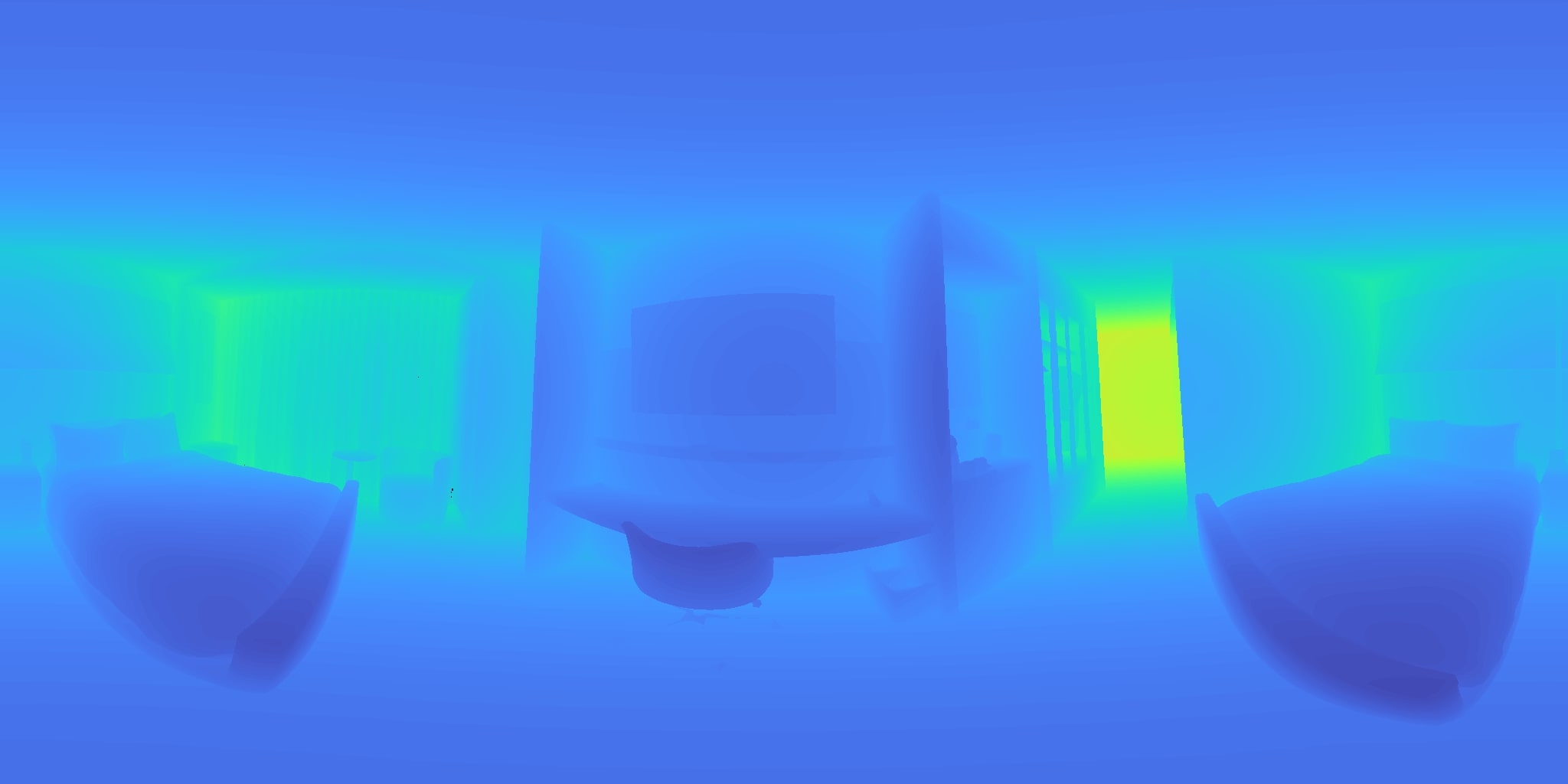}};
	\node [image,below=\padding]  (col4-row2) at (col4-row1.south) {\subfig{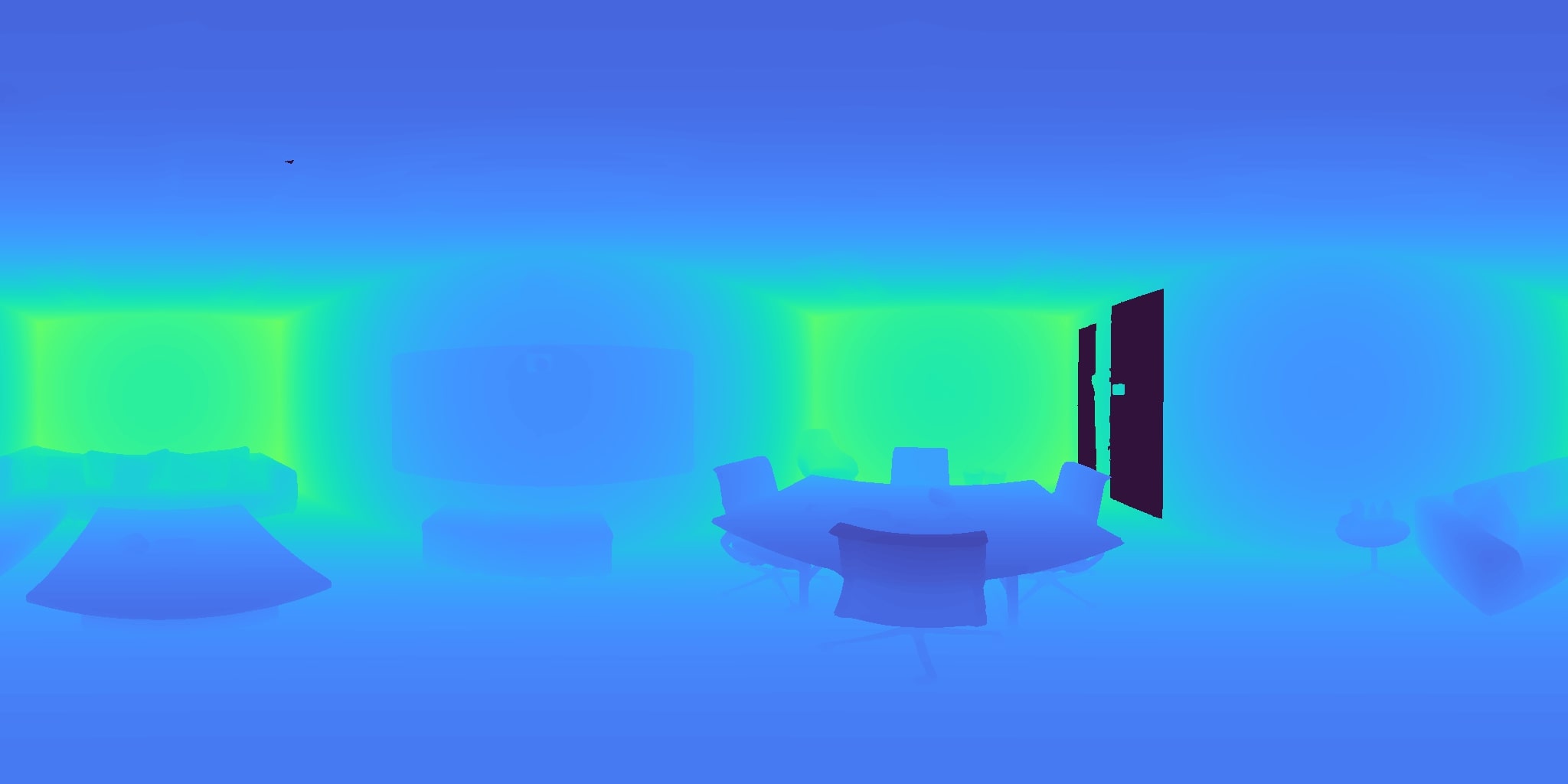}};
	
	\node [image,below=\padding]  (col1-row3) at (col1-row2.south) {\subfig{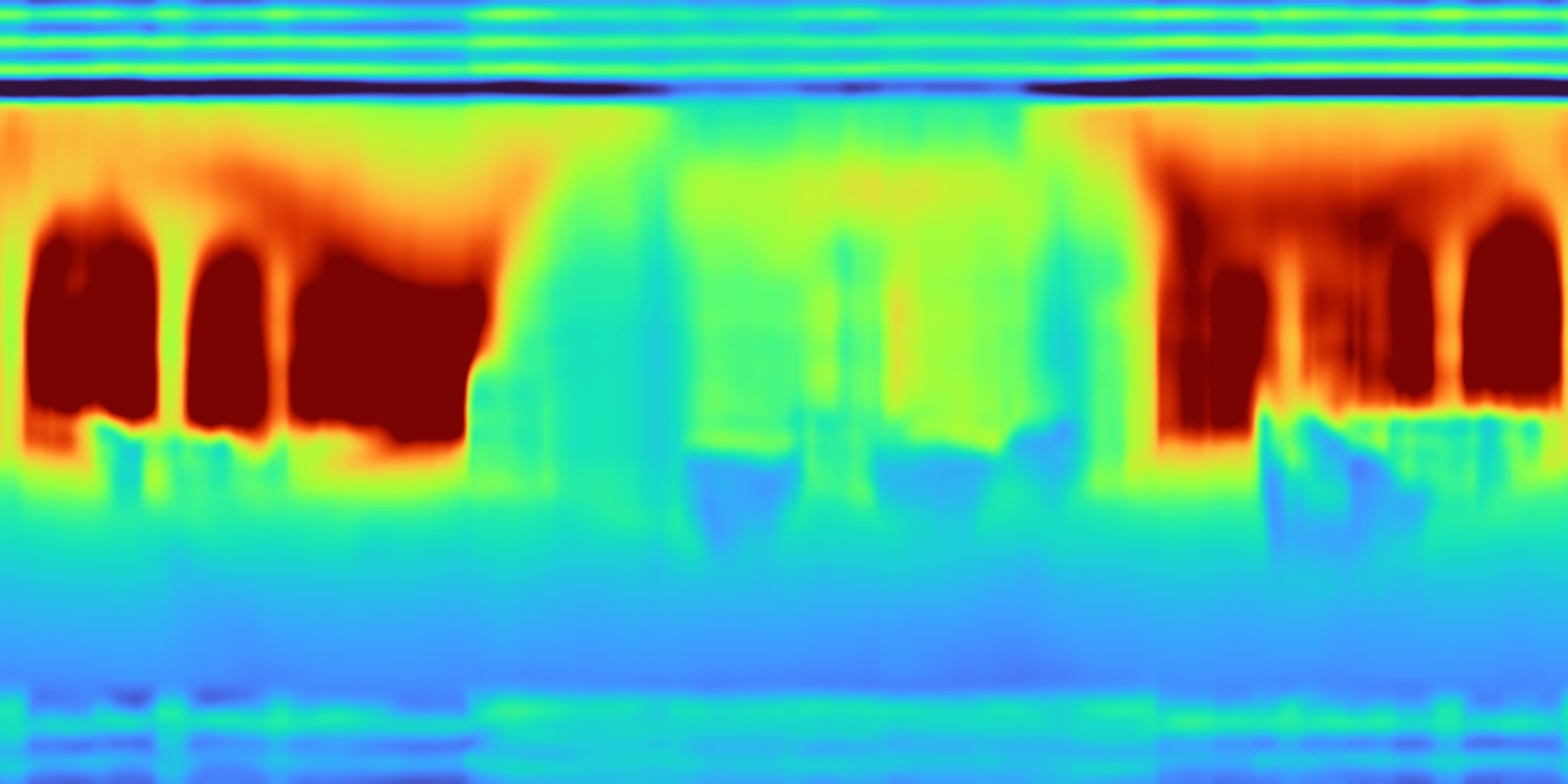}};
	\node [image,below=\padding]  (col2-row3) at (col2-row2.south) {\subfig{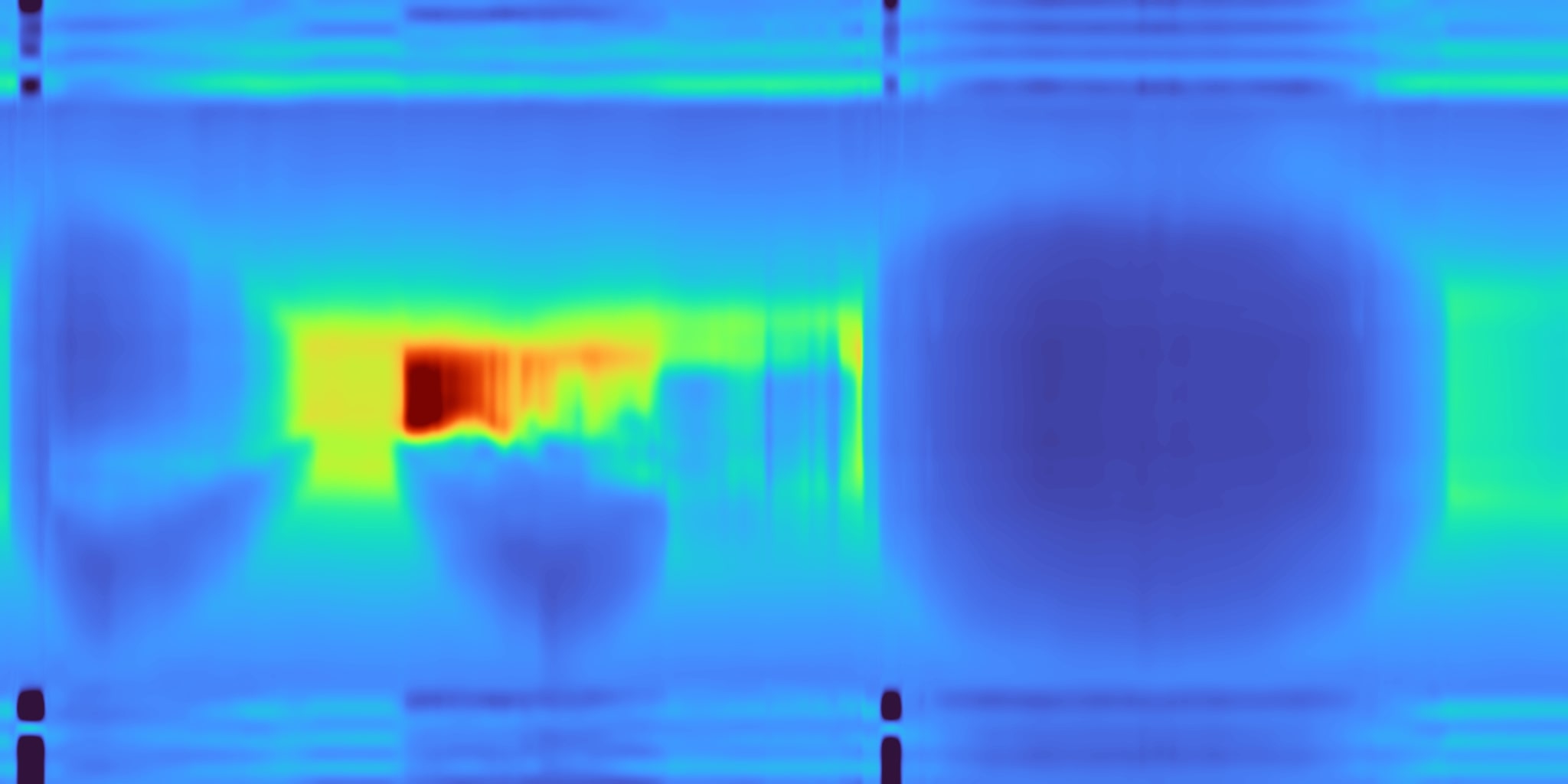}};
	\node [image,below=\padding]  (col3-row3) at (col3-row2.south) {\subfig{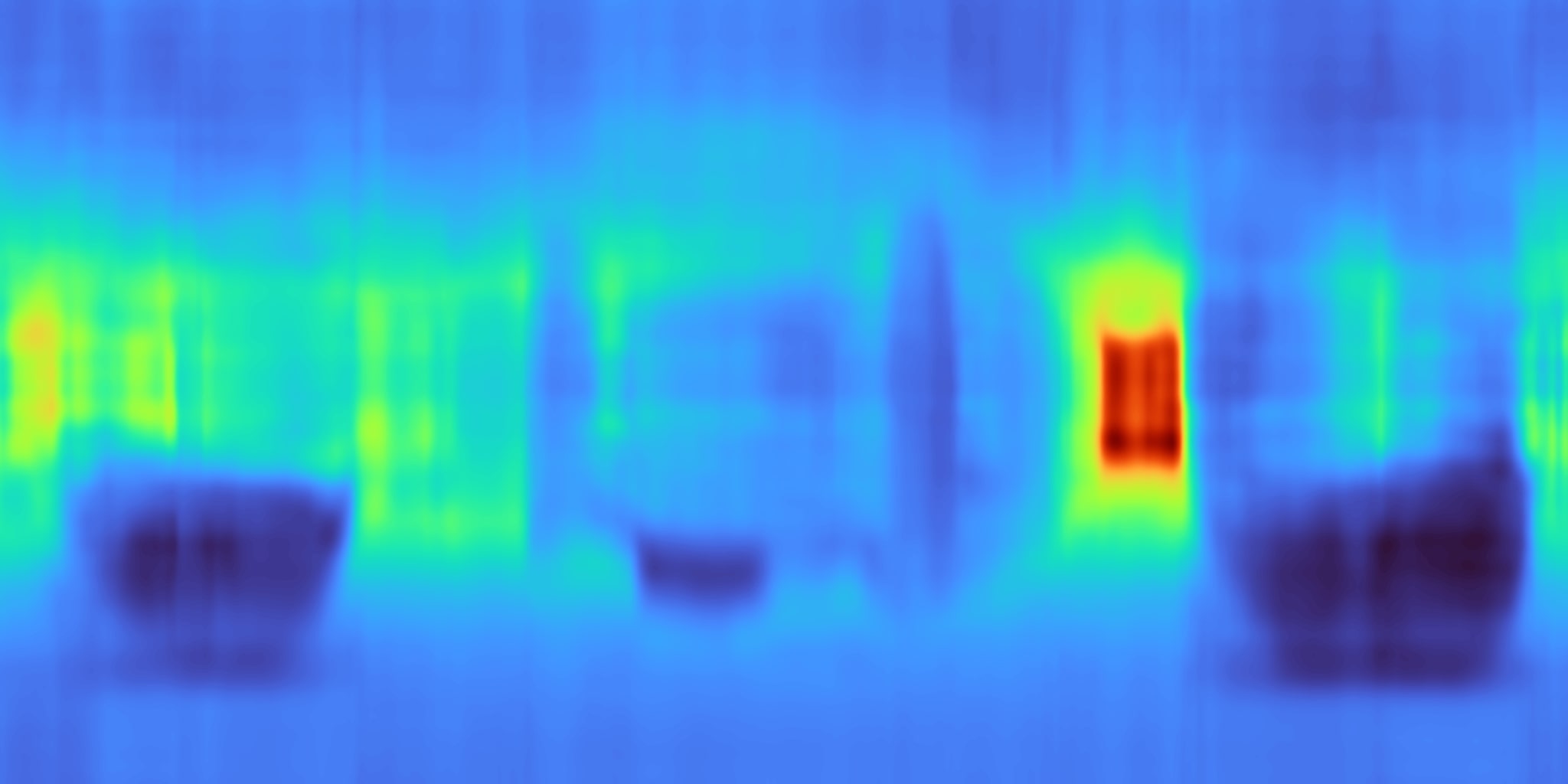}};
	\node [image,below=\padding]  (col4-row3) at (col4-row2.south) {\subfig{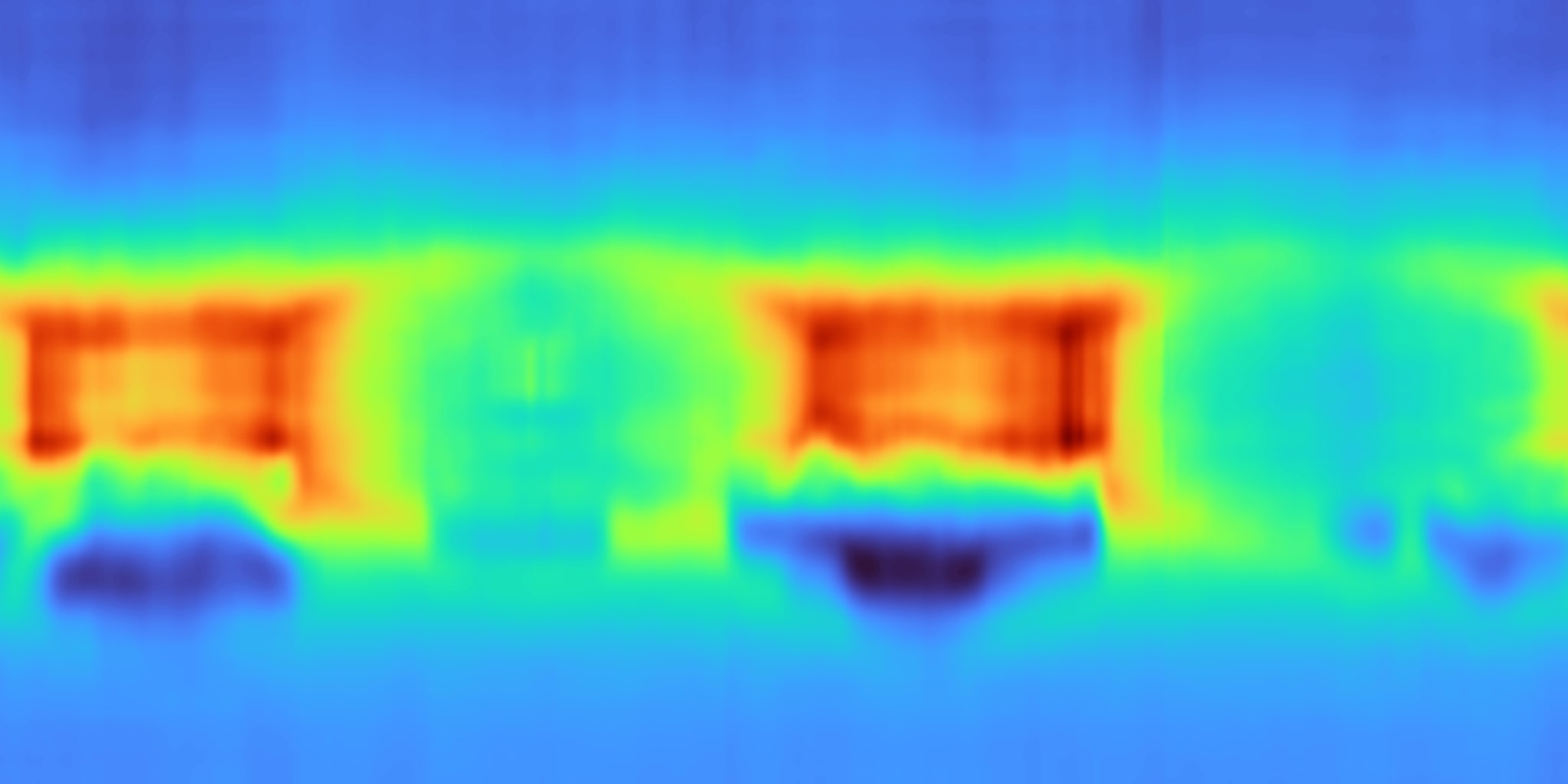}};
	
	\node [image,below=\padding]  (col1-row4) at (col1-row3.south) {\subfig{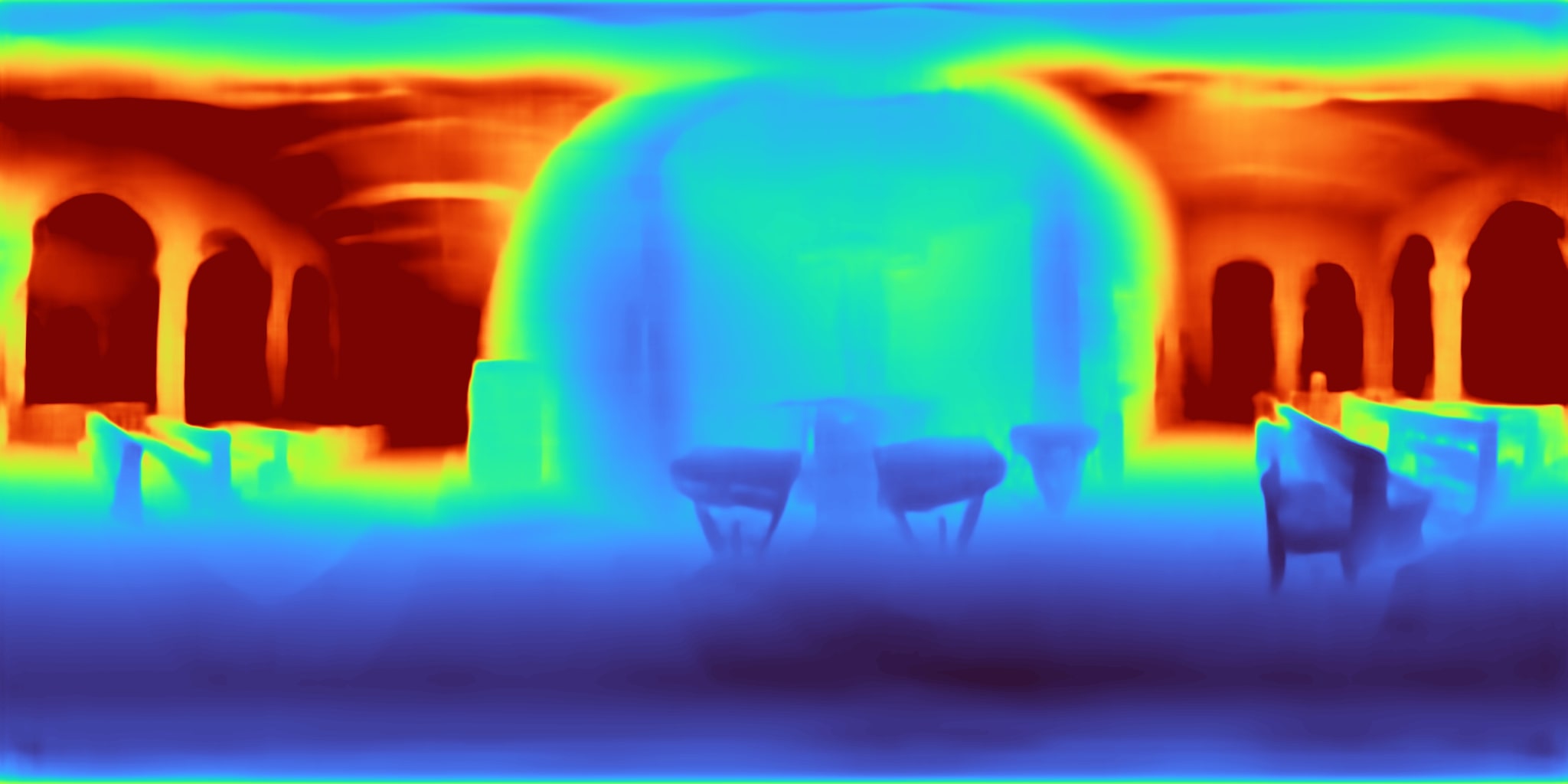}};
	\node [image,below=\padding]  (col2-row4) at (col2-row3.south) {\subfig{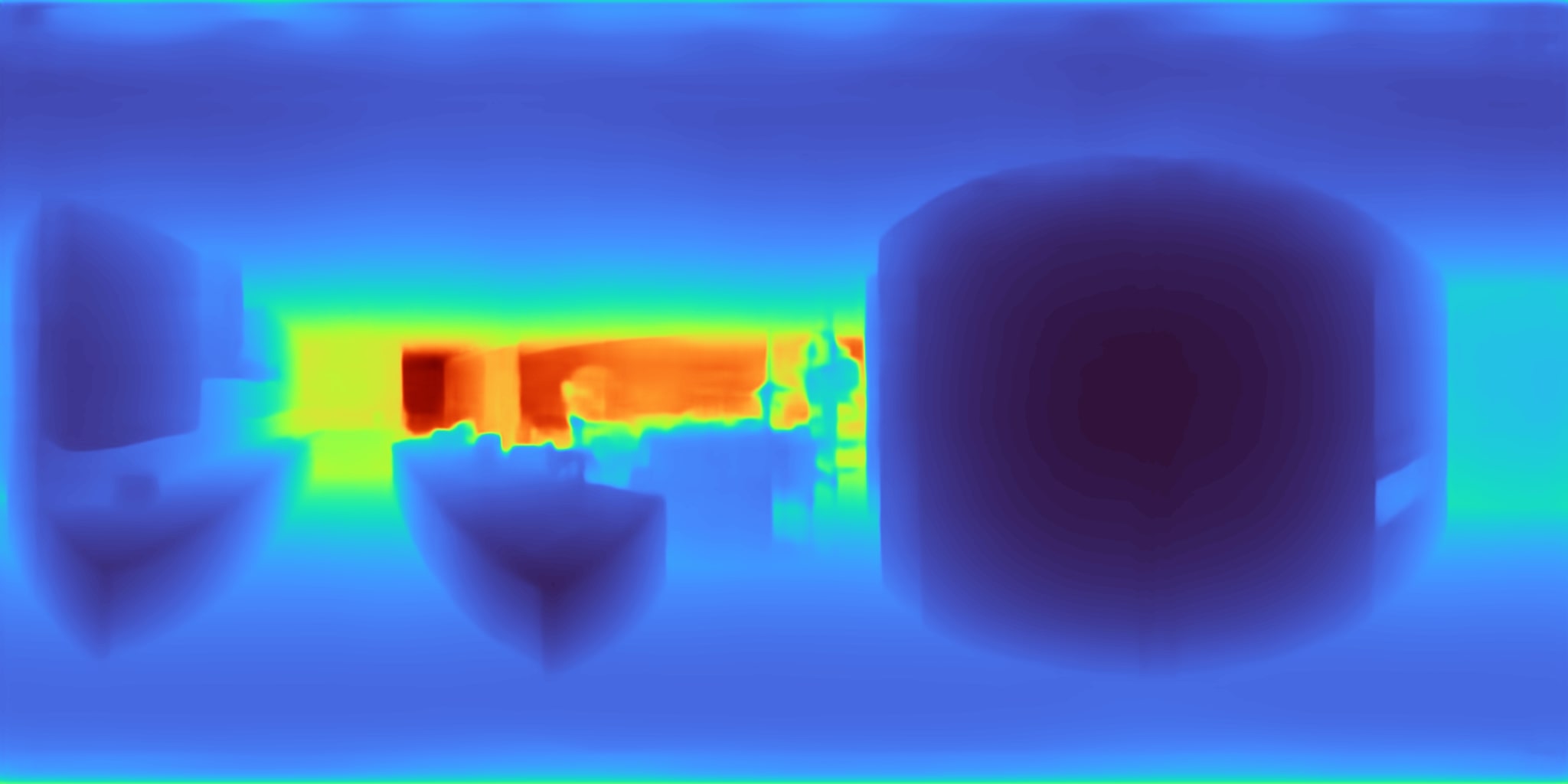}};
	\node [image,below=\padding]  (col3-row4) at (col3-row3.south) {\subfig{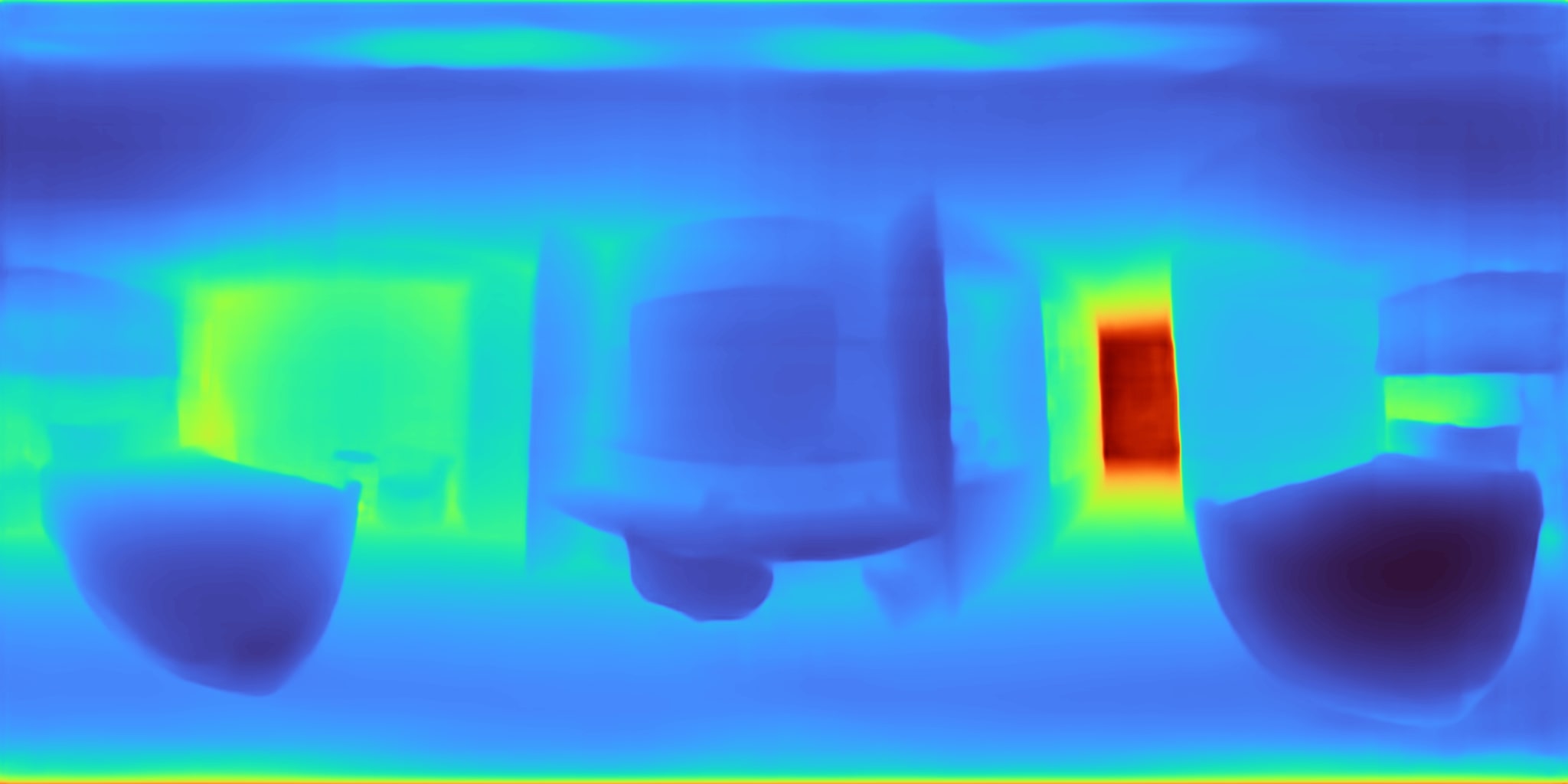}};
	\node [image,below=\padding]  (col4-row4) at (col4-row3.south) {\subfig{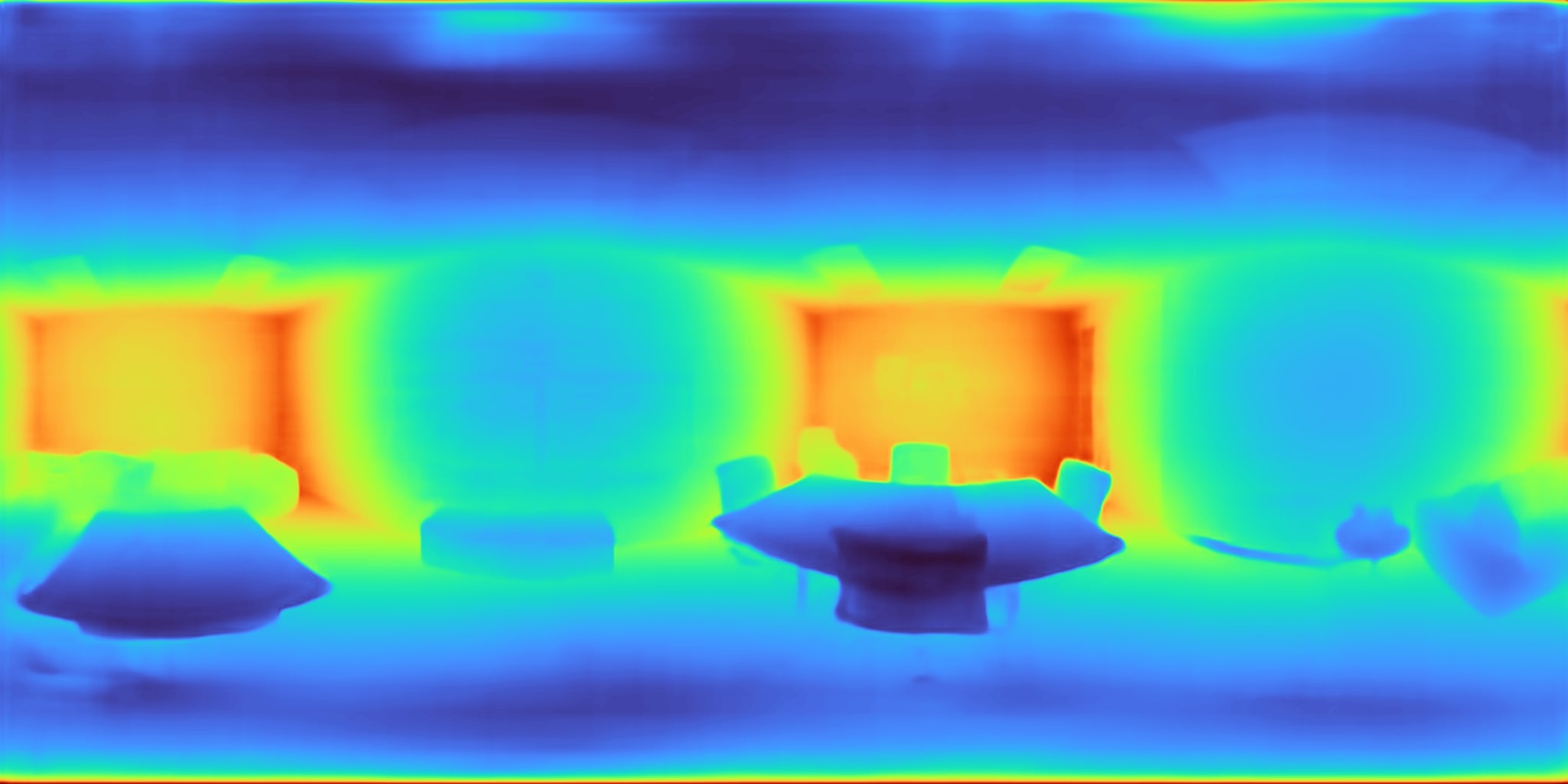}};
	
	\node [image,below=\padding]  (col1-row5) at (col1-row4.south) {\subfig{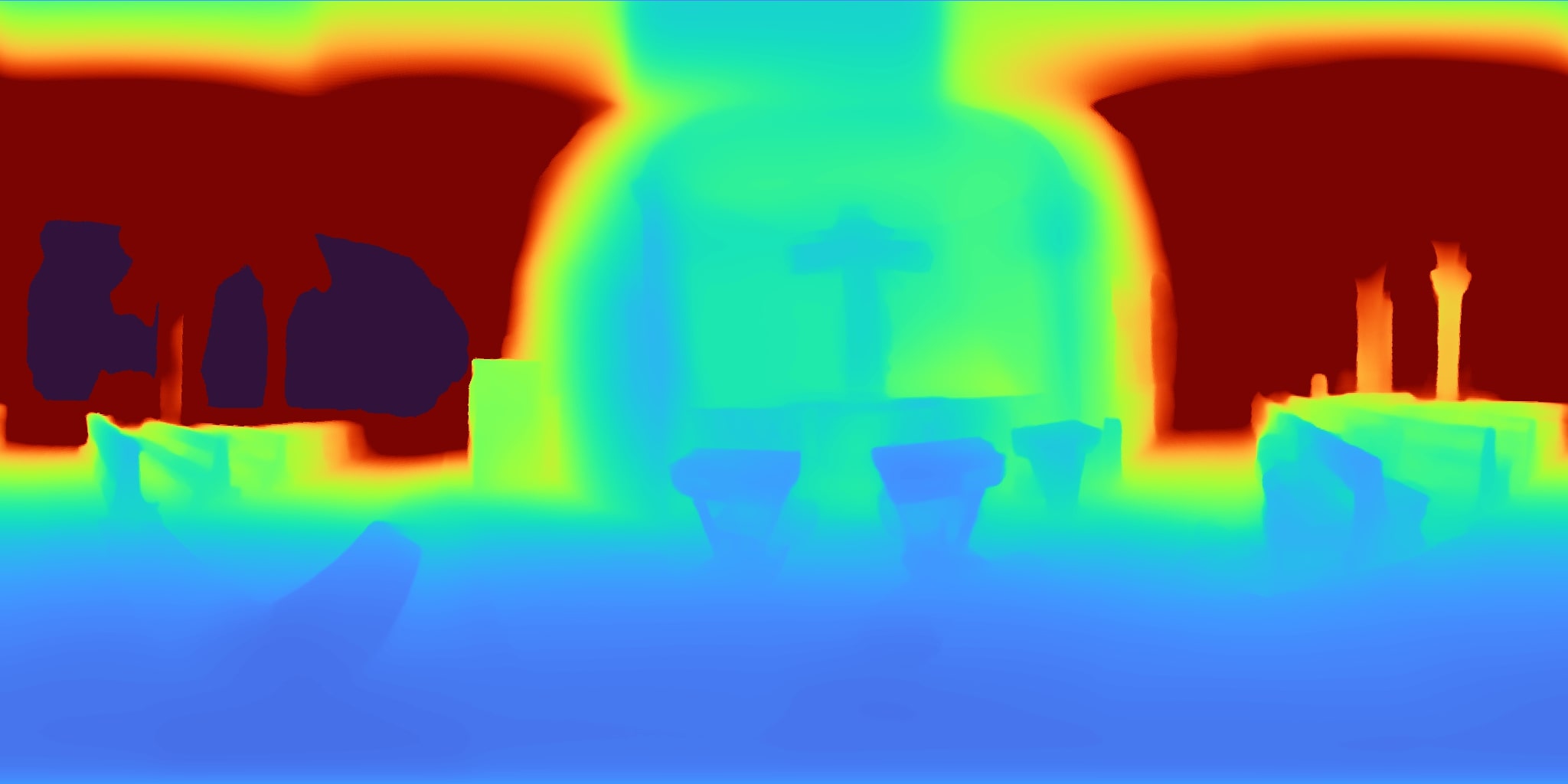}};
	\node [image,below=\padding]  (col2-row5) at (col2-row4.south) {\subfig{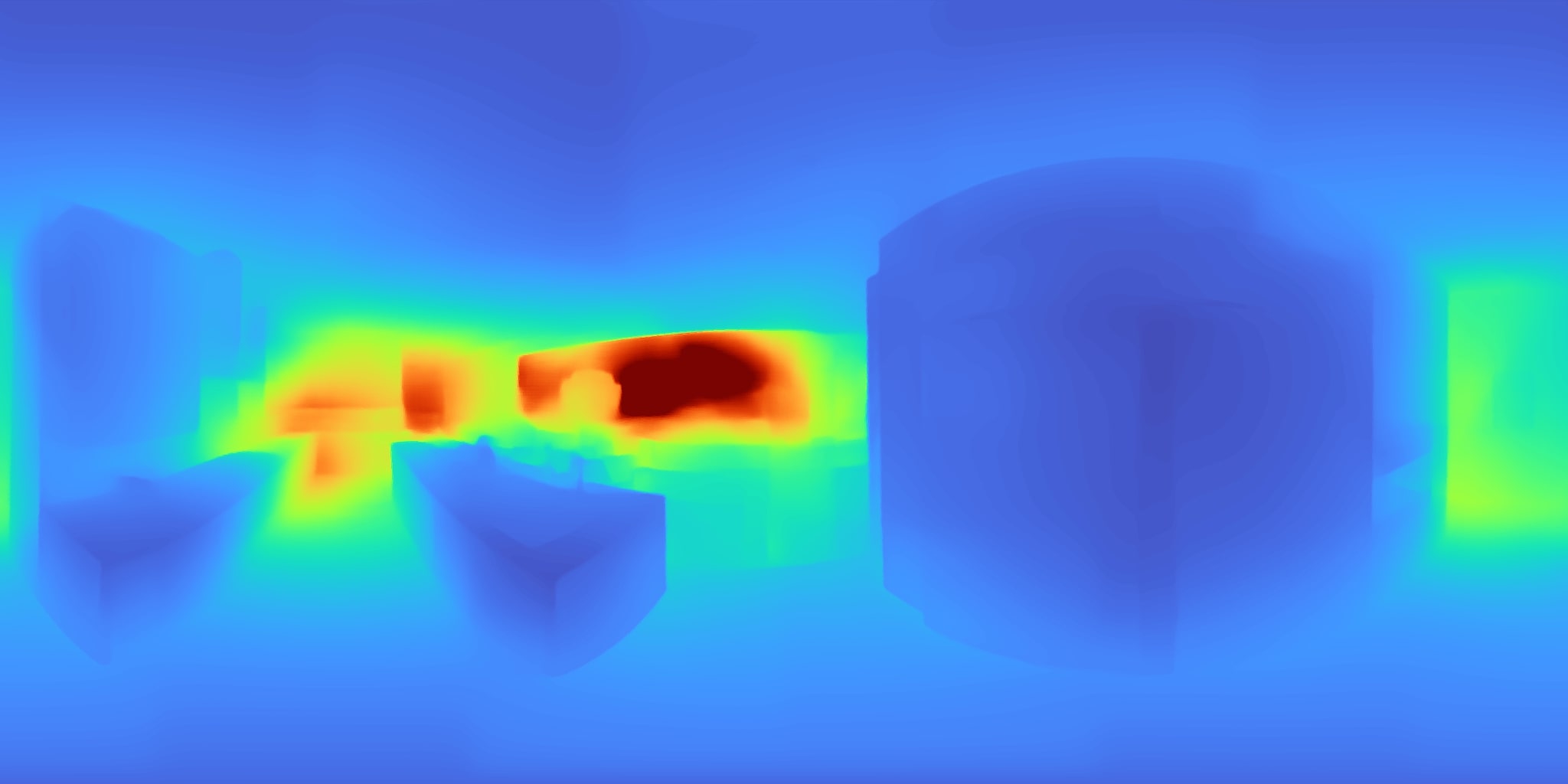}};
	\node [image,below=\padding]  (col3-row5) at (col3-row4.south) {\subfig{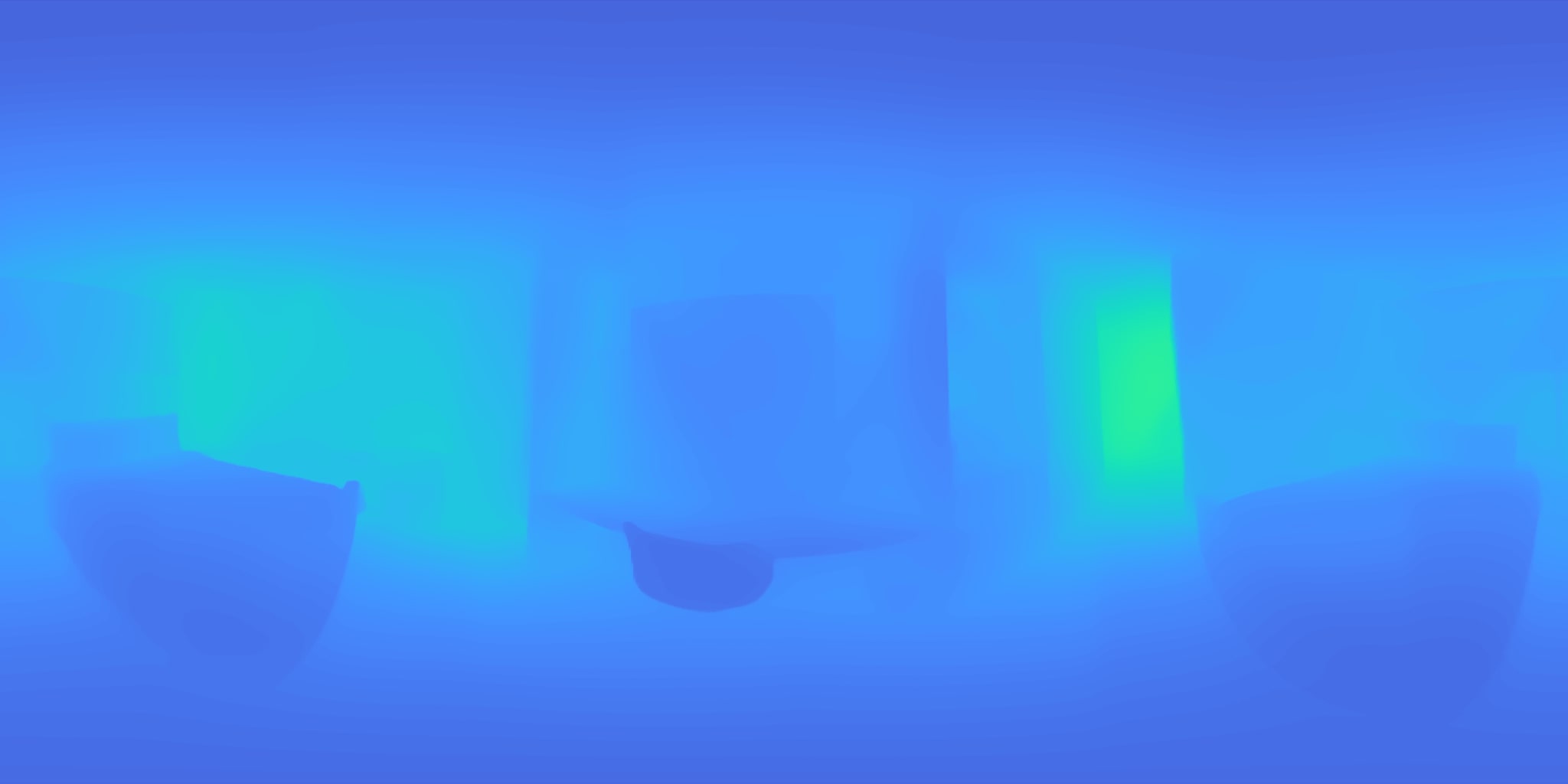}};
	\node [image,below=\padding]  (col4-row5) at (col4-row4.south) {\subfig{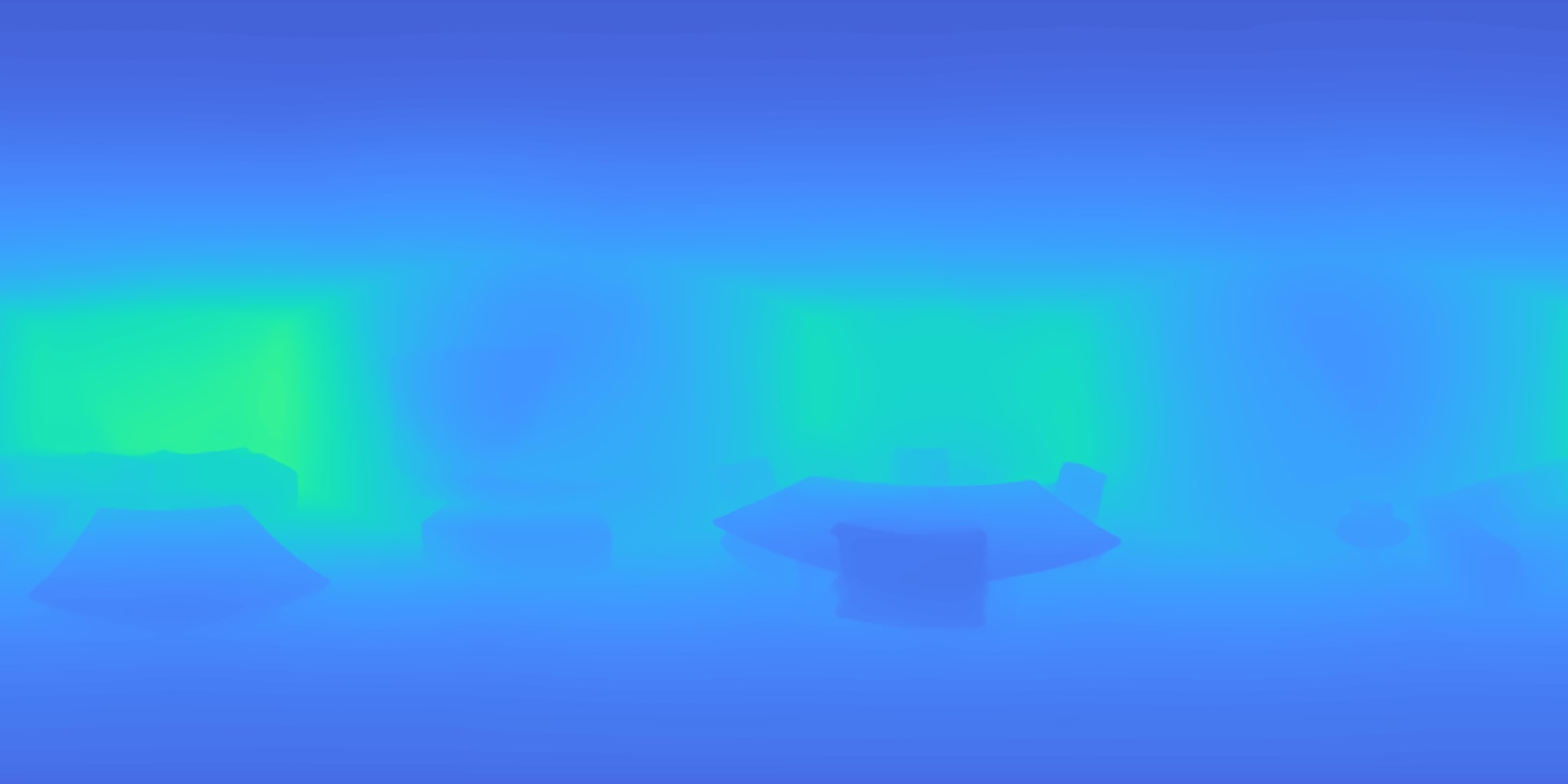}};

	\node [collabel] at ($(col1-row1.north west)!0.5!(col2-row1.north east)$) {Matterport3D-2K \cite{ChangDFHNSSZZ2017}};
	\node [collabel] at ($(col3-row1.north west)!0.5!(col4-row1.north east)$) {Replica360-2K \cite{StrauWMCWGEMRVCYBYPYZLCBGMPSBSNGLN2019,AttalLGRT2020}};
	\node [rowlabel] at (col1-row1.west) {Input image};
	\node [rowlabel] at (col1-row2.west) {Ground truth};
	\node [rowlabel] at (col1-row3.west) {HoHoNet\MData \cite{SunSC2021}};
	\node [rowlabel] at (col1-row4.west) {UniFuse\MData \cite{JiangSZDH2021}};
	\node [rowlabel] at (col1-row5.west) {Ours\Mviii};

\end{tikzpicture}%
	\caption{\label{fig:matterport-replica-comparison}%
		Estimated 360° depth maps at 2K resolution for indoor environments.
		Our results are closer to the ground-truth depth maps.
	}
\end{figure*}

%% file: figures/omniphotos-comparison.tex
\begin{figure}%
\centering
\begin{tikzpicture}[
	image/.style={inner sep=0pt, outer sep=0pt},
	collabel/.style={above=9pt, anchor=north, inner ysep=0pt, scale=0.8, align=center},
	rowlabel/.style={left=9pt, rotate=90, anchor=north, inner ysep=0pt, scale=0.8, align=center},
	subcaption/.style={inner xsep=0.75mm, inner ysep=0.75mm, below right},
	arrow/.style={-{Latex[length=2.5mm,width=4mm]}, line width=2mm},
	spy using outlines={rectangle, size=1.85cm, magnification=4, connect spies, ultra thick, every spy on node/.append style={thick}},
	style1/.style={cyan!90!black,thick},
	style2/.style={orange!90!black},
	style3/.style={blue!90!black},
	style4/.style={red!90!black},
	]
	
	\def\padding{0.005\linewidth}
	\newcommand{\subfig}[2][0px 0px 0px 0px]{\includegraphics[height=2.0cm, trim=#1, clip]{figures/#2}}

	\node [image]                 (col1-row1) at (0,0)             {\subfig{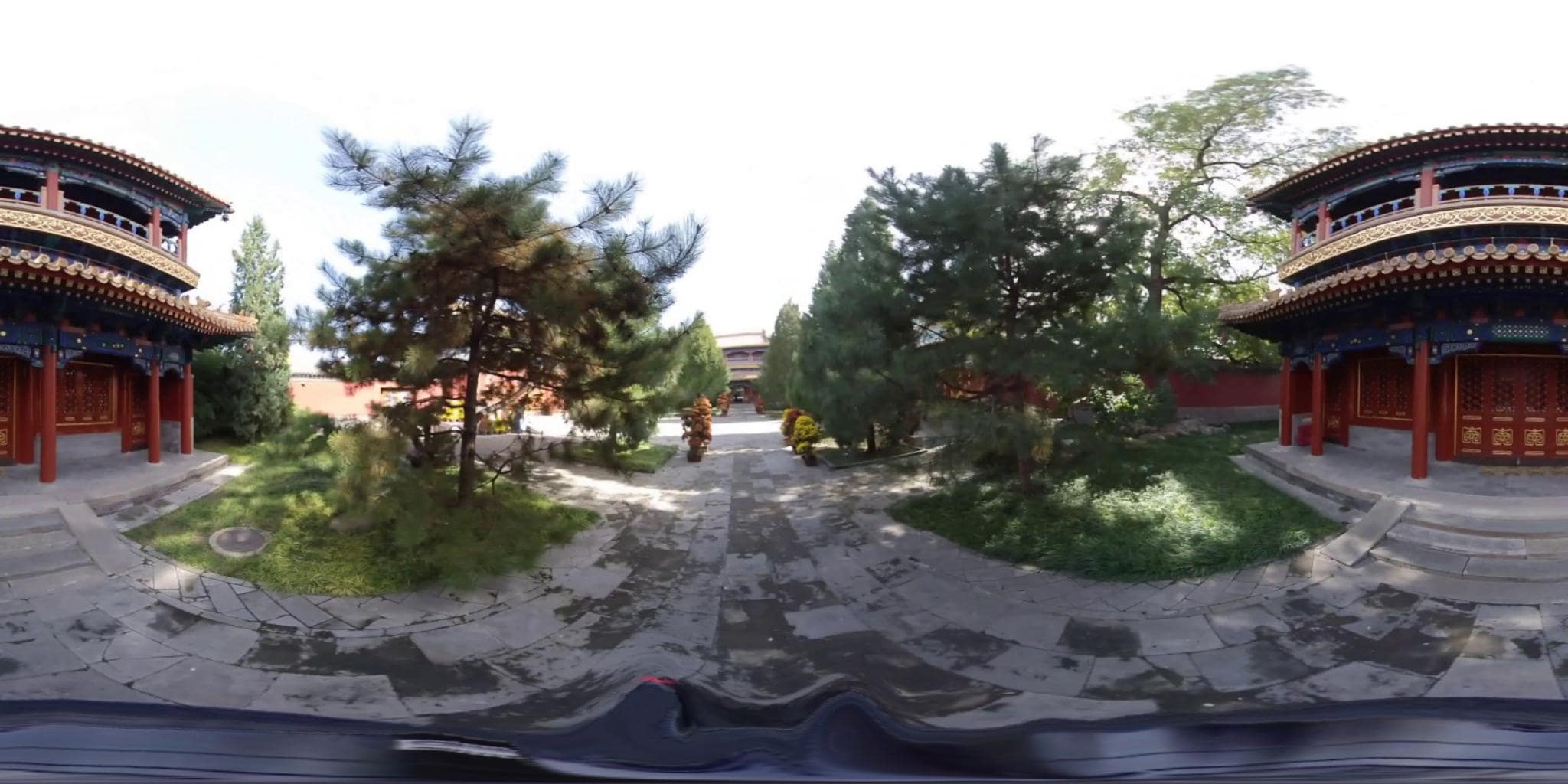}};
	\node [image,right=\padding]  (col2-row1) at (col1-row1.east)  {\subfig{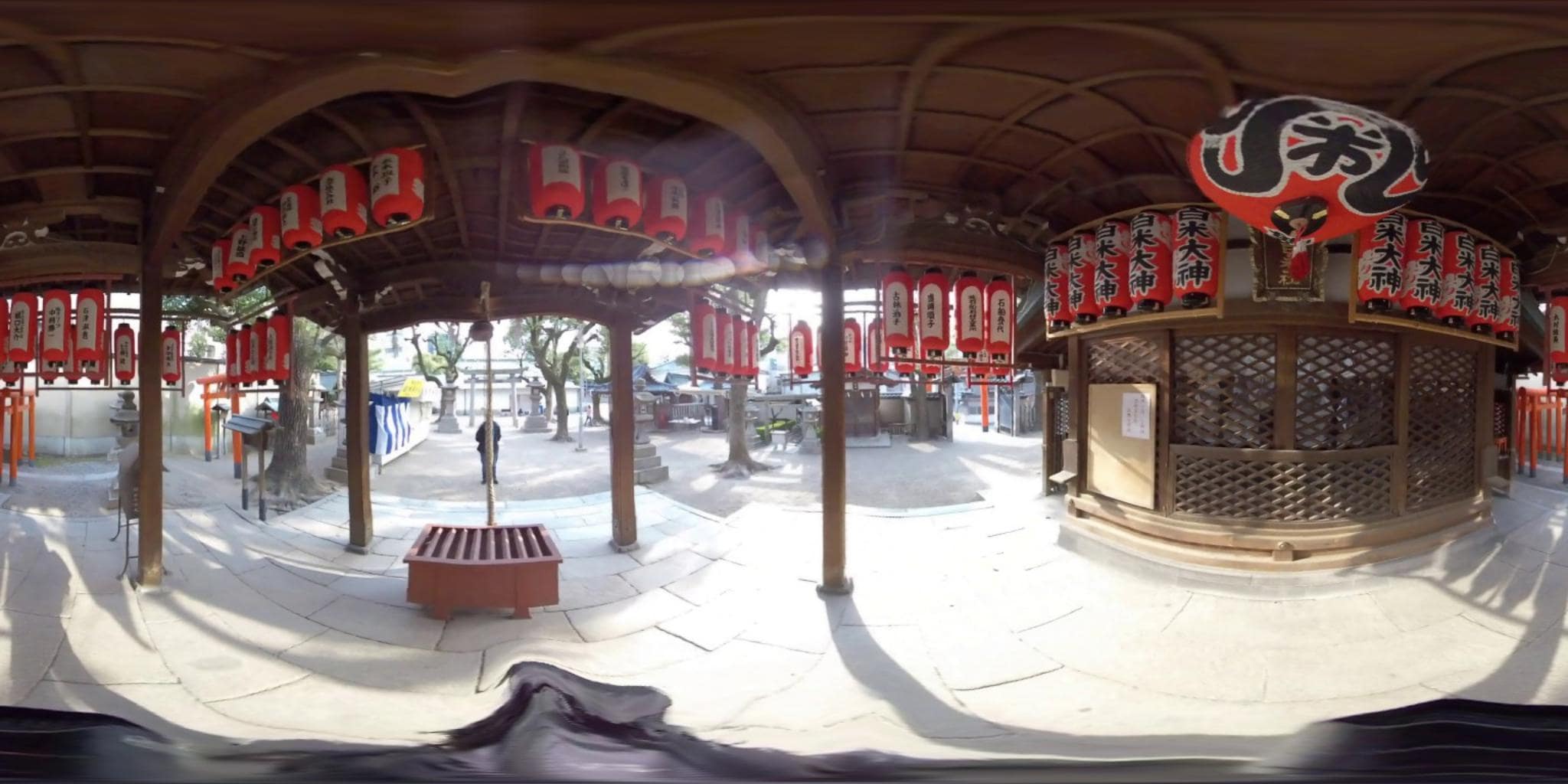}};

	\node [image,below=\padding]  (col1-row2) at (col1-row1.south) {\subfig{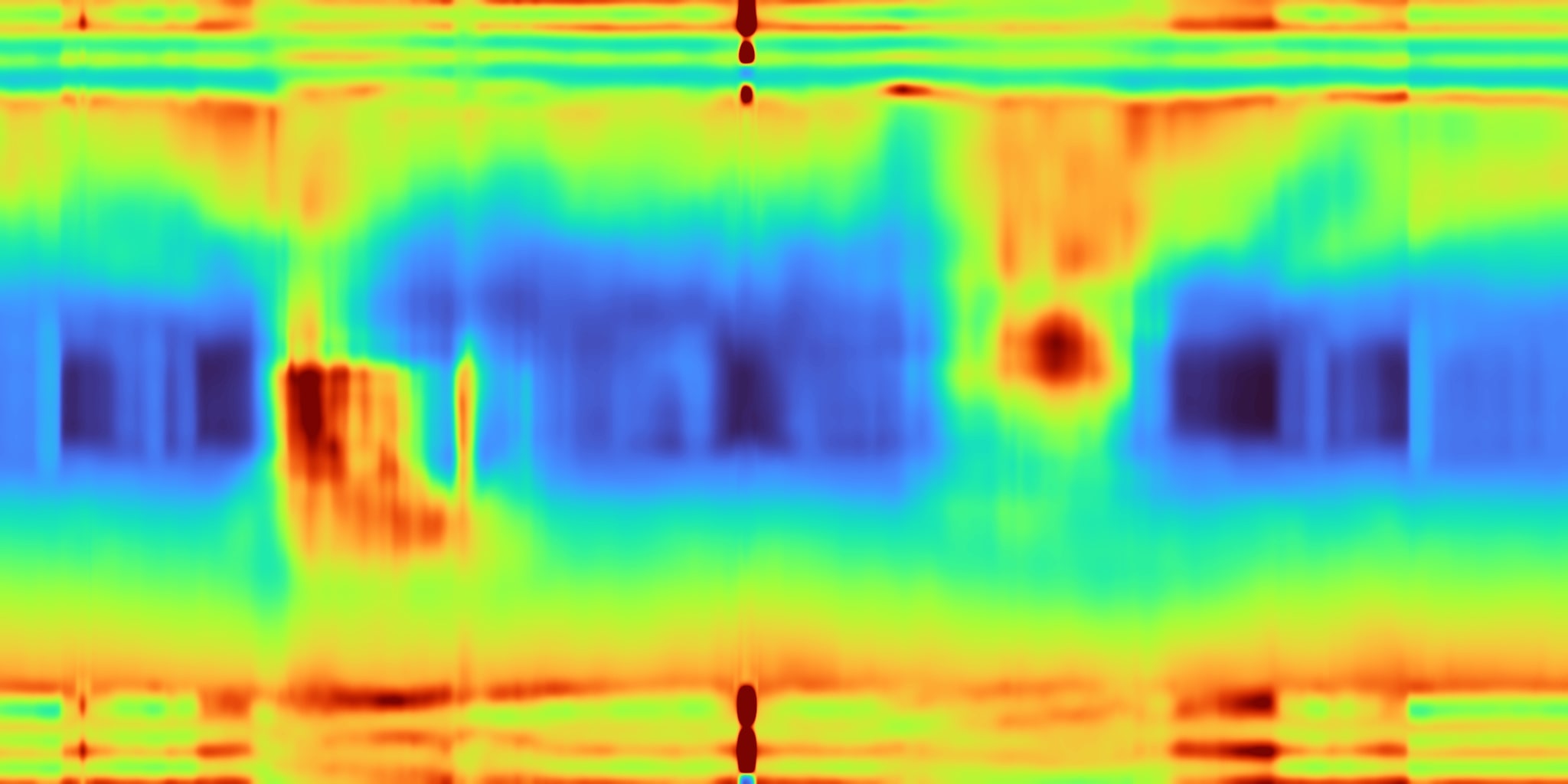}};
	\node [image,below=\padding]  (col2-row2) at (col2-row1.south) {\subfig{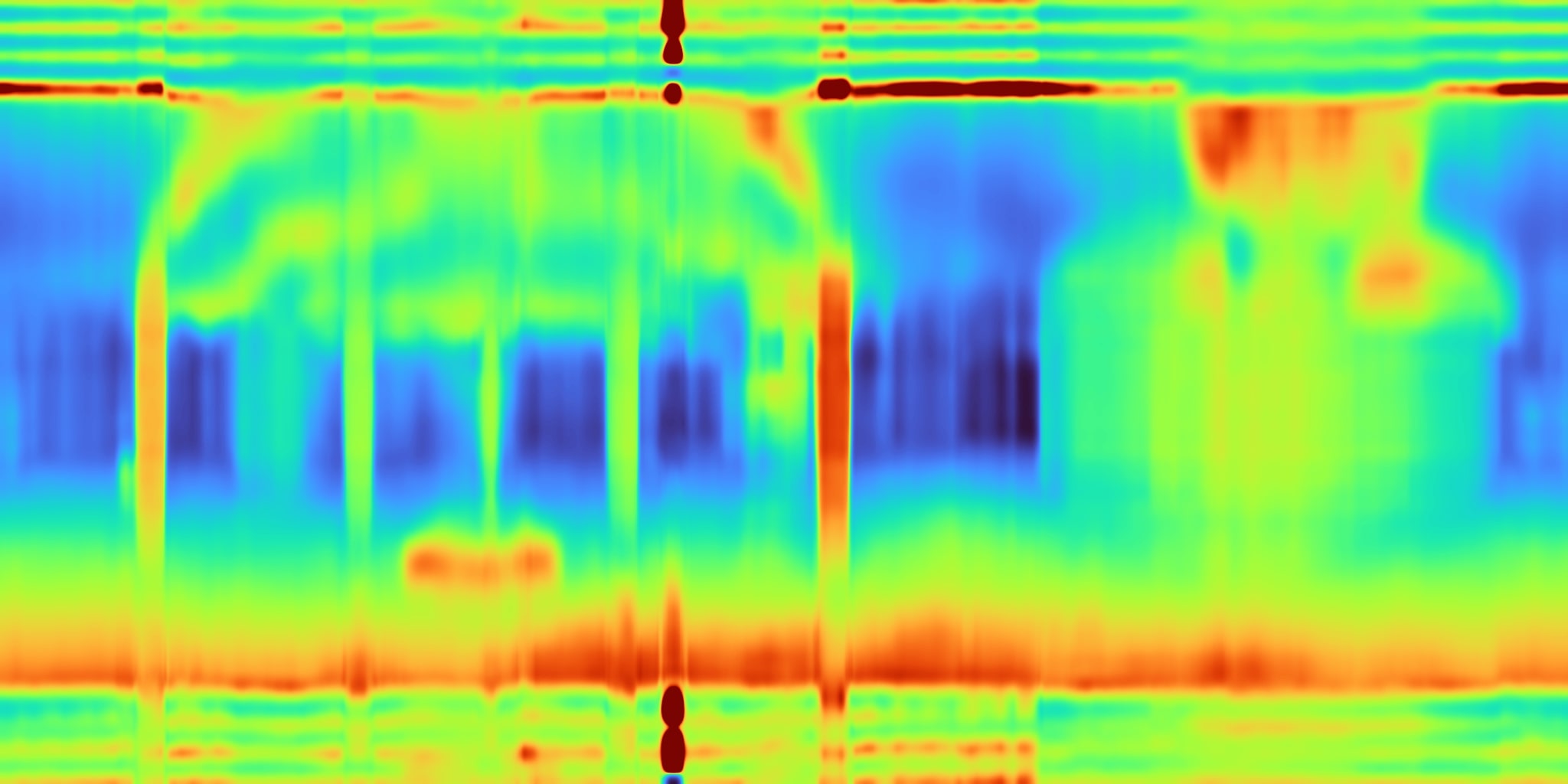}};
	
	\node [image,below=\padding]  (col1-row3) at (col1-row2.south) {\subfig{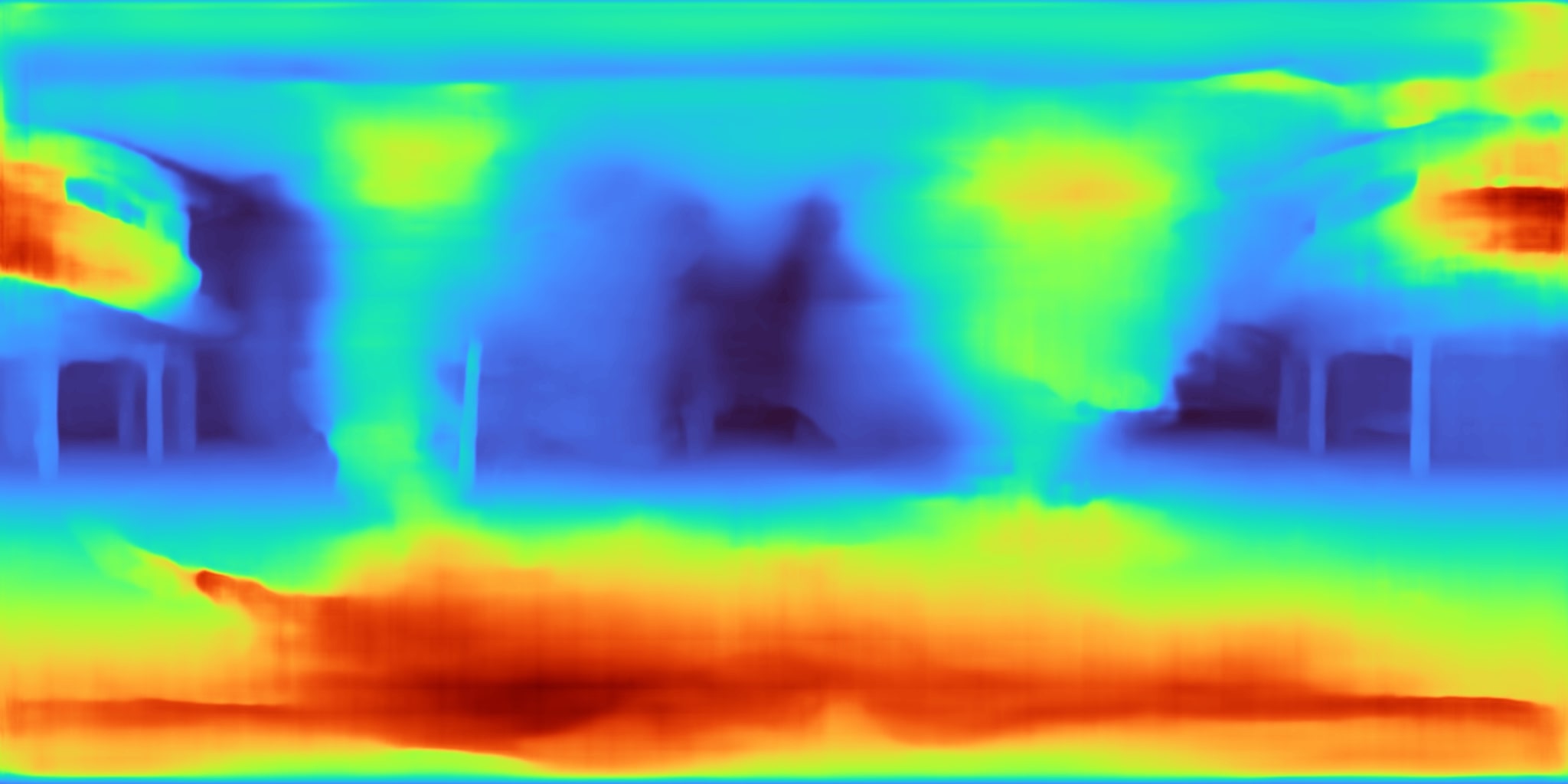}};
	\node [image,below=\padding]  (col2-row3) at (col2-row2.south) {\subfig{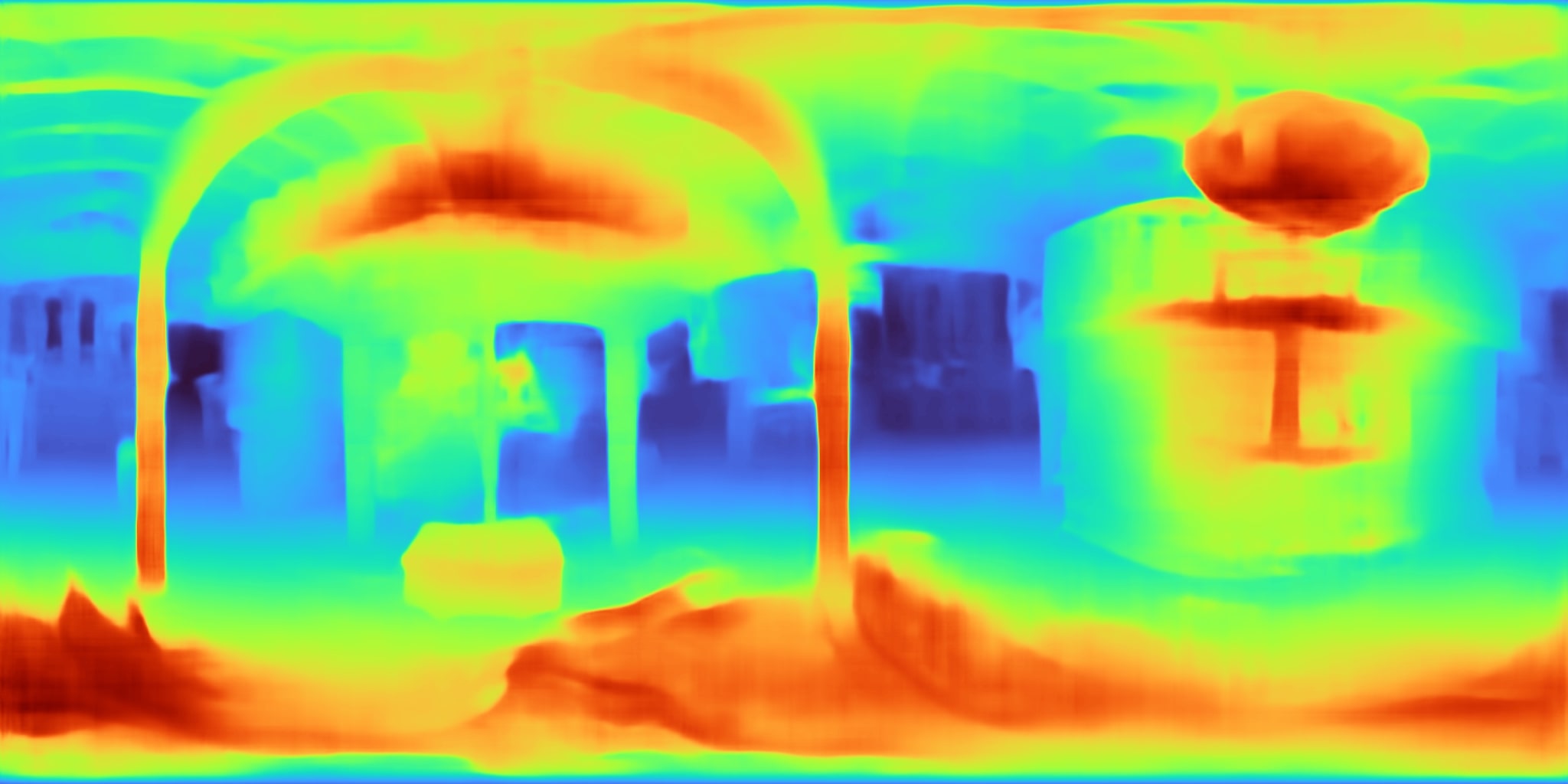}};
	
	\node [image,below=\padding]  (col1-row4) at (col1-row3.south) {\subfig{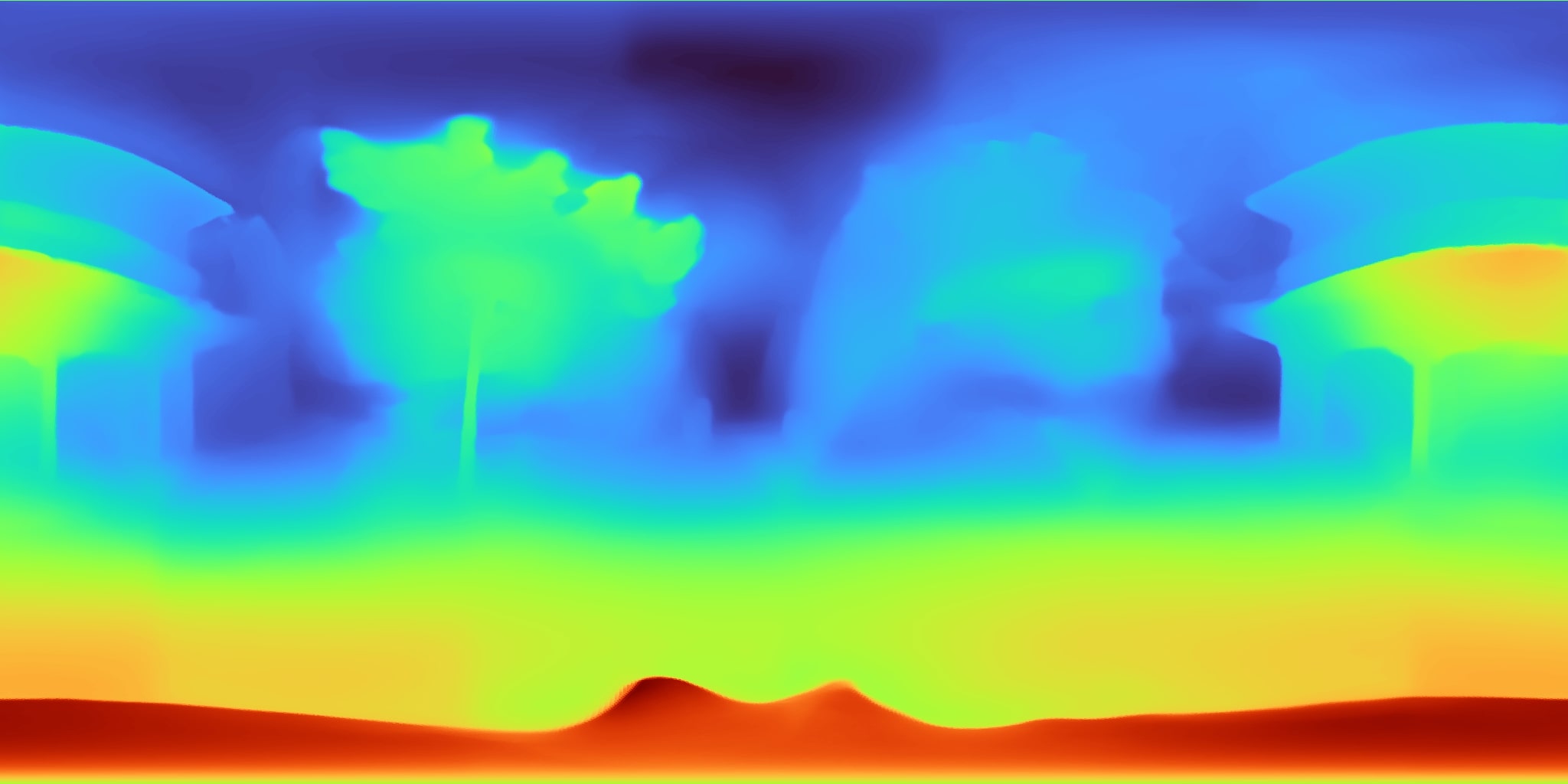}};
	\node [image,below=\padding]  (col2-row4) at (col2-row3.south) {\subfig{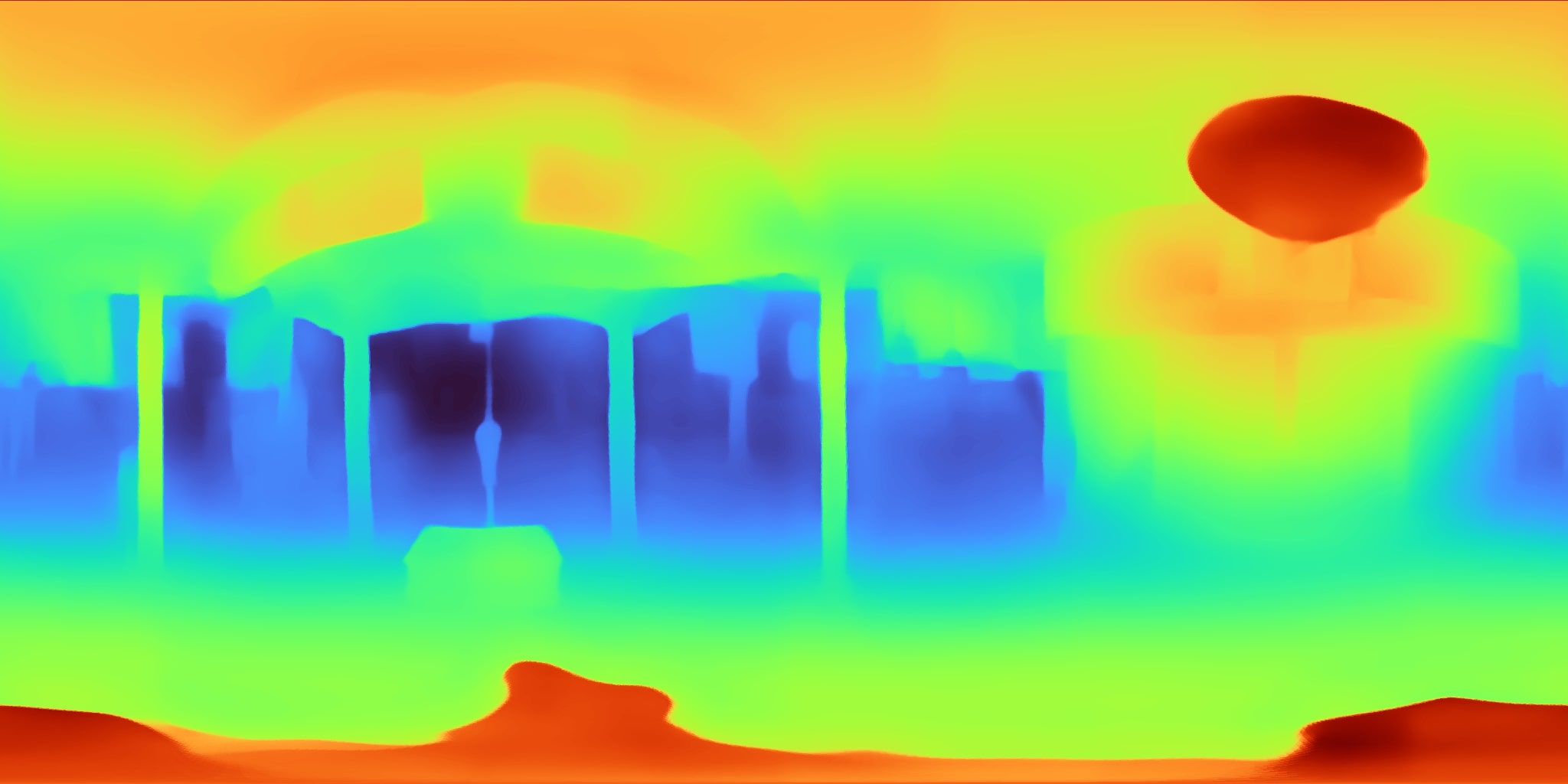}};

	\node [rowlabel] at (col1-row1.west) {Input image};
	\node [rowlabel] at (col1-row2.west) {HoHoNet\MData \cite{SunSC2021}};
	\node [rowlabel] at (col1-row3.west) {UniFuse\MData \cite{JiangSZDH2021}};
	\node [rowlabel] at (col1-row4.west) {Ours\Mviii};

\end{tikzpicture}%
	\caption{\label{fig:omniphotos-comparison}%
		Estimated 360° disparity maps at 2048$\times$1024 for outdoor environments \cite{BerteYLR2020}.
		Our results are more consistent geometrically.
	}
\end{figure}

%% file: figures/quantitative-table-4k.tex
\begin{table}
\newcommand{\ul}{\color{blue!70!black}\bf}
\caption{\label{tab:results-4k}%
	Quantitative results for Replica360-4K at 4096$\times$2048 with frustum blending (best trade-off between runtime and performance).
	For superscripts, see \cref{tab:results}.
	Highlighting: {\ul best}, \textbf{second-best}.
}
\renewcommand*{\arraystretch}{1.15}%
\setlength{\tabcolsep}{2pt}%
\centering%
\resizebox{\linewidth}{!}{%
	\begin{tabular}{lccccccc}
		\toprule
		Method &
		\scriptsize AbsRel\tridown &
		\scriptsize MAE\tridown &
		\scriptsize RMSE\tridown &
		\scriptsize RMSE-log\tridown &
		\scriptsize $\delta \!<$1.25\triup &
		\scriptsize $\delta \!<$1.25\textsuperscript{2}\triup &
		\scriptsize $\delta \!<$1.25\textsuperscript{3}\triup
		\\ \midrule
		OmniDepth                          &     0.337 &     0.582 &     0.778 &     0.161 &     0.484 &     0.785 &     0.920 \\
		BiFuse                             &     0.292 &     0.445 &     0.637 &     0.143 &     0.606 &     0.857 &     0.941 \\
		HoHoNet\MData                      &     0.251 &     0.379 &     0.509 &     0.127 &     0.670 &     0.884 &     0.948 \\
		HoHoNet\SData                      &     0.208 & \bf 0.335 & \ul 0.455 &     0.106 &     0.728 &     0.909 &     0.961 \\
		UniFuse\MData                      &     0.223 & \ul 0.324 & \bf 0.464 &     0.116 &     0.744 &     0.910 &     0.959 \\ \midrule
		Ours\Mvii  (multi-scale)           & \ul 0.150 & \bf 0.335 &     0.558 & \ul 0.081 & \ul 0.813 & \ul 0.953 & \bf 0.983 \\
		Ours\Mviii (multi-scale)           & \bf 0.161 & 	 0.363 &     0.607 & \bf 0.085 & \bf 0.781 & \bf 0.951 & \ul 0.984 \\ \bottomrule
	\end{tabular}%
}%
\end{table}

%% file: figures/ablations-table.tex
\begin{table}
	\newcommand{\ul}{\color{blue!70!black}\bf}
	\caption{\label{tab:ablations-table}%
		Ablation studies for disparity map alignment (top) and blending (bottom), evaluated on the Matterport3D test set.
		Multi-scale deformable alignment outperforms all single-scale alignments across all metrics when using MiDaS v3.
		Gradient-based Poisson blending outperforms simpler blending modes in all but one metric when using MiDaS v2.
		Highlighting: {\ul best}, \textbf{second-best}.
	}
	\renewcommand*{\arraystretch}{1.15}%
	\setlength{\tabcolsep}{3pt}%
	\centering%
	\resizebox{\linewidth}{!}{%
		\begin{tabular}{lccccccc}
			\toprule
			Method &
			\scriptsize AbsRel\tridown &
			\scriptsize MAE\tridown &
			\scriptsize RMSE\tridown &
			\scriptsize RMSE-log\tridown &
			\scriptsize $\delta \!<$1.25\triup &
			\scriptsize $\delta \!<$1.25\textsuperscript{2}\triup &
			\scriptsize $\delta \!<$1.25\textsuperscript{3}\triup
			\\ \midrule
			No alignment\Mviii                 &     0.259 &     0.600 &     0.969 &     0.150 &     0.532 &     0.821 &     0.933 \\
			2$\times$2 single-scale\Mviii      & \bf 0.210 &     0.476 & \bf 0.838 &     0.122 &     0.654 &     0.888 & \bf 0.959 \\
			4$\times$3 single-scale\Mviii      & \bf 0.210 & \bf 0.475 & \bf 0.838 & \bf 0.121 & \bf 0.655 & \bf 0.889 & \bf 0.959 \\
			8$\times$7 single-scale\Mviii      & \bf 0.210 &     0.476 &     0.840 & \bf 0.121 & \ul 0.656 & \bf 0.889 &     0.958 \\
			16$\times$14 single-scale\Mviii    &     0.231 &     0.528 &     0.905 &     0.134 &     0.609 &     0.859 &     0.944 \\
			multi-scale\Mviii                  & \ul 0.208 & \ul 0.446 & \ul 0.791 & \ul 0.119 & \ul 0.656 & \ul 0.890 & \ul 0.961 \\ \midrule \midrule
			NN blending\Mvii                   & \bf 0.226 &     0.501 &     0.841 & \bf 0.131 & \bf 0.611 & \bf 0.864 & \bf 0.952 \\
			Mean blending\Mvii                 &     0.230 &     0.501 & \bf 0.828 &     0.132 &     0.601 &     0.859 & \bf 0.952 \\
			Frustum blending\Mvii              &     0.229 & \bf 0.499 & \ul 0.826 & \bf 0.131 &     0.604 &     0.861 & \ul 0.953 \\
			Poisson blending\Mvii              & \ul 0.224 & \ul 0.494 &     0.831 & \ul 0.130 & \ul 0.616 & \ul 0.866 & \ul 0.953 \\ \bottomrule
		\end{tabular}%
	}\\[1mm]
	\scriptsize%
	\Mvii~Using MiDaS v2 \cite{RanftLHSK2021}
	\qquad\Mviii~Using MiDaS v3 \cite{RanftBK2021}
\end{table}

%% file: 5-discussion.tex
\section{Discussion and Conclusion}

Our method can fail if the tangent disparity estimates are incorrect, e.g. for large plain walls, saturated skies, or photorealistic wallpapers.
As these estimates improve over time, our method can take advantage of them.
In some cases, the least-squares rescaling to fit the ground-truth disparity results in negative disparities, which produces incorrect, negative depth values.
We also saw inconsistencies in the ground-truth depth maps, such as mirrors or missing lamps or chandeliers that are visible in the image.
We show examples of these failure cases in the supplemental document.

\noindent
We found in our experiments that blending disparity maps with the `frustum' weights (see \cref{fig:blending_weights}) usually produces results that are nearly as good as (see \cref{tab:ablations-table}) but considerably faster than the Poisson blending of our complete method.
This is a good compromise if speed is of essence.
Concurrent to our work, \citet{LiGYHDR2022} use transformers for aligning and blending tangent depth maps based on predicted confidence.

Our proposed framework is the first to deal with high-resolution 360° images, and not limited to indoor scenes.
Projecting the spherical input image onto a set of tangent images lets us overcome both the distortions of spherical projections and the resolution limits of deep monocular depth estimation methods.
We proposed specially tailored optimisation techniques for global deformable multi-scale alignment and gradient-domain blending of the individual tangent disparity maps to overcome the discontinuous nature of tangent images.
A major advantage of our approach is that we can leverage the high performance of MiDaS (or any future method) to generalise to new 360° datasets with higher accuracy and resolution than previous approaches.
The resulting disparity maps at 2K resolution show a high level of geometric detail for both indoor and outdoor scenes.

%% file: s-supplement.tex
\section{Metrics and evaluation procedure}
Like MiDaS, our disparity estimates are ambiguous up to scale and offset.
We therefore determine the optimal scale and offset to match the ground-truth disparity map (inverse depth) using least squares \cite[][Equation~14]{RanftLHSK2021}.
As all baselines predict depth and not disparity, we rescale them similarly but in depth space.
In the following metrics, $z$ and $z^*$ represent the predicted and ground-truth depth, respectively:

\begin{itemize}
	\itemsep0em %
	
	\item Absolute relative error (AbsRel):
	$\frac{1}{N} \sum_{i=1}^{N} \frac{\abs{z_i - z_i^*}}{z_i^*}$

	\item Mean absolute error (MAE):
	$\frac{1}{N} \sum_{i=1}^{N} \abs{z_i - z_i^*}$
	
	\item RMSE:
	$\sqrt{\frac{1}{N} \sum_{i=1}^{N} \norm{z_i - z_i^*}^2}$
	
	\item RMSE (log):
	$\sqrt{\frac{1}{N} \sum_{i=1}^{N} \norm{\log_{10} z+i - \log_{10} z_i^*}^2}$
	
	\item Accuracy $\delta \!<\! \tau$:
	\% of $z$ s.t. $\delta = \max\!\big(\frac{z_i}{z_i^*}, \frac{z_i^*}{z_i}\big) < \tau$
\end{itemize}

\section{Runtime measurements}

We measured the runtime of our method on a 2.1–3.2\,GHz 16-core Xeon Silver 4216 processor with an NVIDIA RTX 3090 GPU.
\Cref{tab:runtime} list the runtime for preprocessing, including factorisation of the Poisson blending problem matrix, and the time required for each of the four stages of our method.

\begin{table}[b]
	\caption{\label{tab:runtime}%
		Runtime measurements of our framework for different stages nd input resolutions (`Res.'), in seconds.
			For Poisson blending, we factorise the linear system in a preprocessing step once.
	}
	\centering%
	\resizebox{\linewidth}{!}{%
		\renewcommand*{\arraystretch}{1.15}%
		\setlength\tabcolsep{3pt}%
		\begin{tabular}{lcccccc}
			\toprule
			&& once & \multicolumn{4}{c}{per image} \\ \cmidrule(lr){3-3} \cmidrule(lr){4-7}
			Blending & Res. & Preproc. & Projection & MiDaS & Alignment & Blending\\
			\midrule
			Frustum \Mvii  & 2K &  --- & 1.0 & 11.2 & 37.0 & \phantom{0}3.7 \\
			Frustum \Mviii & 2K &  --- & 1.0 & 24.3 & 39.6 & \phantom{0}3.0 \\
			Poisson \Mvii  & 2K & 43.5 & 1.0 & 10.3 & 37.8 &  17.4 \\
			Poisson \Mviii & 2K & 46.7 & 1.0 & 25.4 & 41.5 &  17.9 \\
			\midrule \midrule
			Frustum \Mvii  & 4K & --- & 1.1 & 11.3 & 37.8 & 13.1 \\
			Frustum \Mviii & 4K & --- & 1.1 & 24.5 & 37.1 & 18.8 \\
			\bottomrule
	\end{tabular}}\\[1mm]
	\scriptsize%
	\Mvii~Using MiDaS v2 \cite{RanftLHSK2021}
	\qquad\Mviii~Using MiDaS v3 \cite{RanftBK2021}
\end{table}

\section{Extended discussion}

Our method can fail if the tangent disparity estimates are incorrect, e.g. for large plain walls, saturated skies, or large photorealistic wallpapers, as shown in \cref{fig:failure-cases} (left).
As monocular depth estimates improve over time, our method can take advantage of them immediately.
In some cases, the least-squares rescaling to fit the ground-truth disparity map pushes disparity values out of bounds, towards negative disparities.
These negative disparities correspond to negative depth values that are incorrect (see \cref{fig:failure-cases}, right).

\input{figures/failure-cases.tex}
\input{figures/incorrect-GT.tex}

We also found
inconsistencies
in the reconstructed meshes of Matterport3D \cite{ChangDFHNSSZZ2017}, such as windows and mirrors with depths labelled at their surface instead of corresponding to the visible scene outside or being reflected, or missing lamps or chandeliers that are clearly visible in the image.
We show examples in \cref{fig:incorrect-GT}, in which our method reconstructs arguably more plausible depth than the ground truth.

%% file: figures/failure-cases.tex
\begin{figure}[b]%
\centering
\begin{tikzpicture}[
	image/.style={inner sep=0pt, outer sep=0pt},
	collabel/.style={above=9pt, anchor=north, inner ysep=0pt, scale=0.8, align=center},
	rowlabel/.style={left=9pt, rotate=90, anchor=north, inner ysep=0pt, scale=0.8, align=center},
	subcaption/.style={inner xsep=0.75mm, inner ysep=0.75mm, below right},
	arrow/.style={-{Latex[length=2.5mm,width=4mm]}, line width=2mm},
	spy using outlines={rectangle, size=1.85cm, magnification=4, connect spies, ultra thick, every spy on node/.append style={thick}},
	style1/.style={cyan!90!black,thick},
	style2/.style={orange!90!black},
	style3/.style={blue!90!black},
	style4/.style={red!90!black},
	]
	
	\def\padding{0.005\linewidth}
	\newcommand{\subfig}[2][0px 0px 0px 0px]{\includegraphics[width=4.0cm, trim=#1, clip]{figures/#2}}

	\node [image]                 (col1-row1) at (0,0)             {\subfig[500 350 800 300]{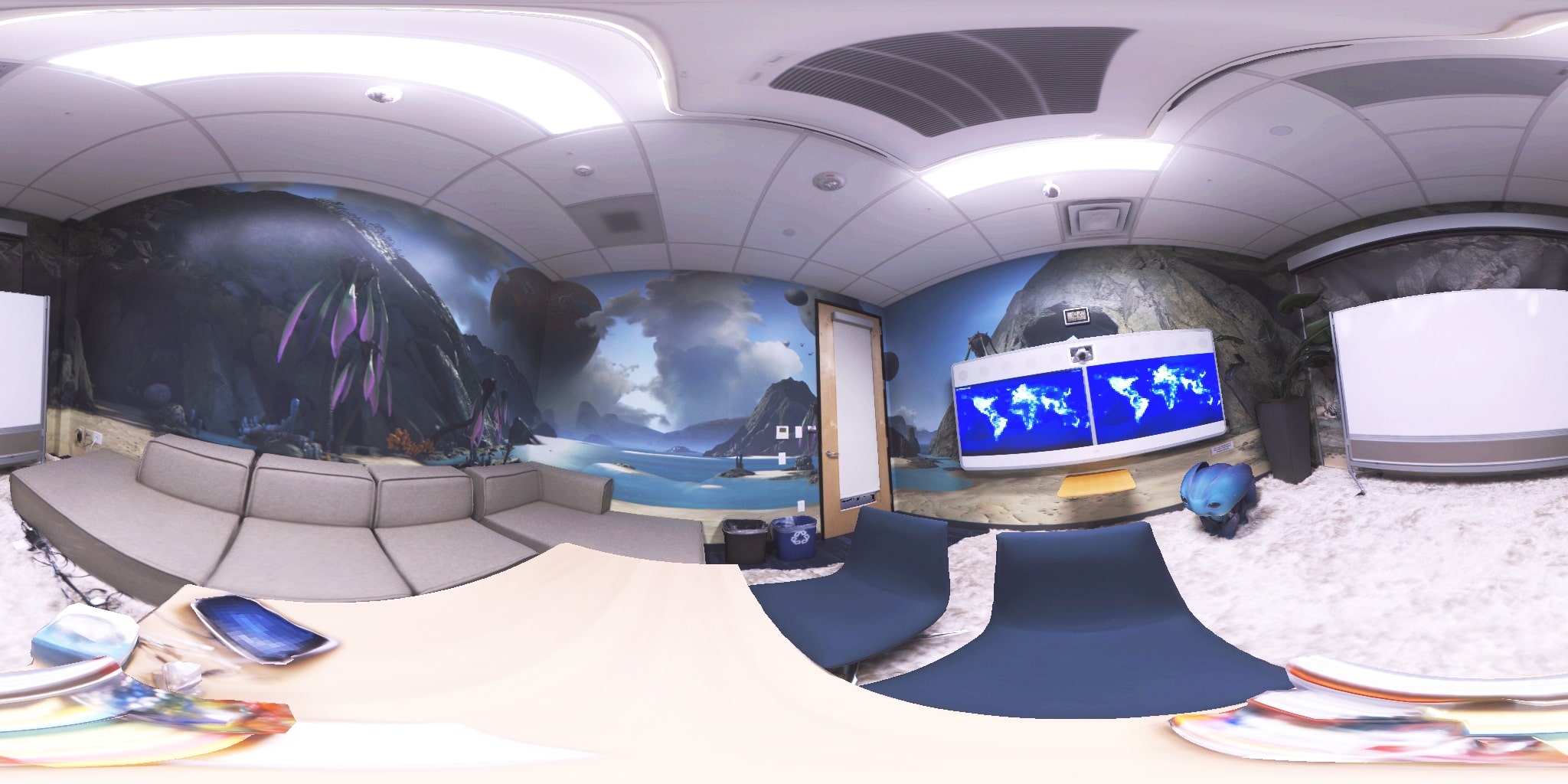}};
	\node [image,right=\padding]  (col2-row1) at (col1-row1.east)  {\subfig{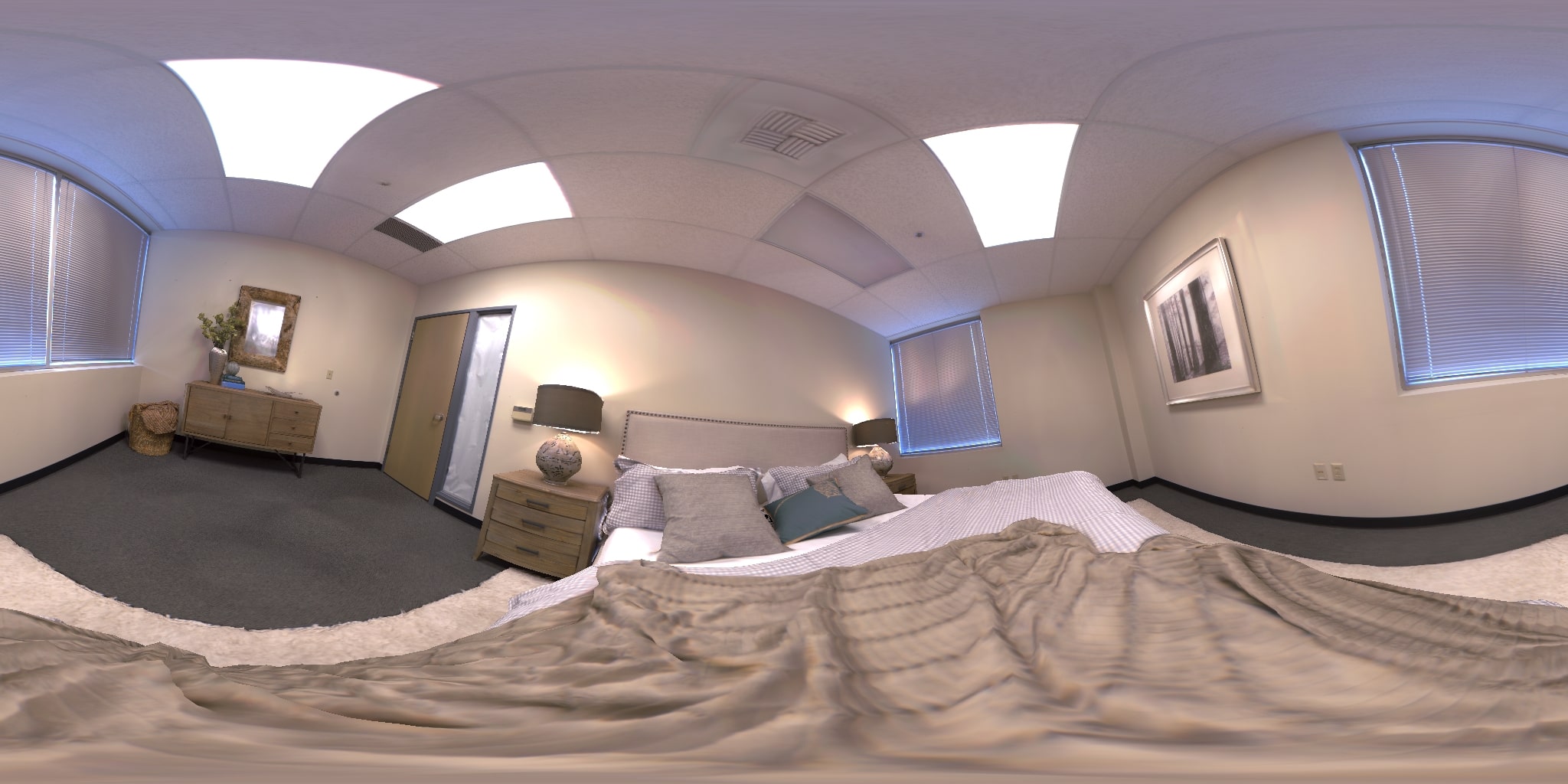}};

	\node [image,below=\padding]  (col1-row2) at (col1-row1.south) {\subfig[500 350 800 300]{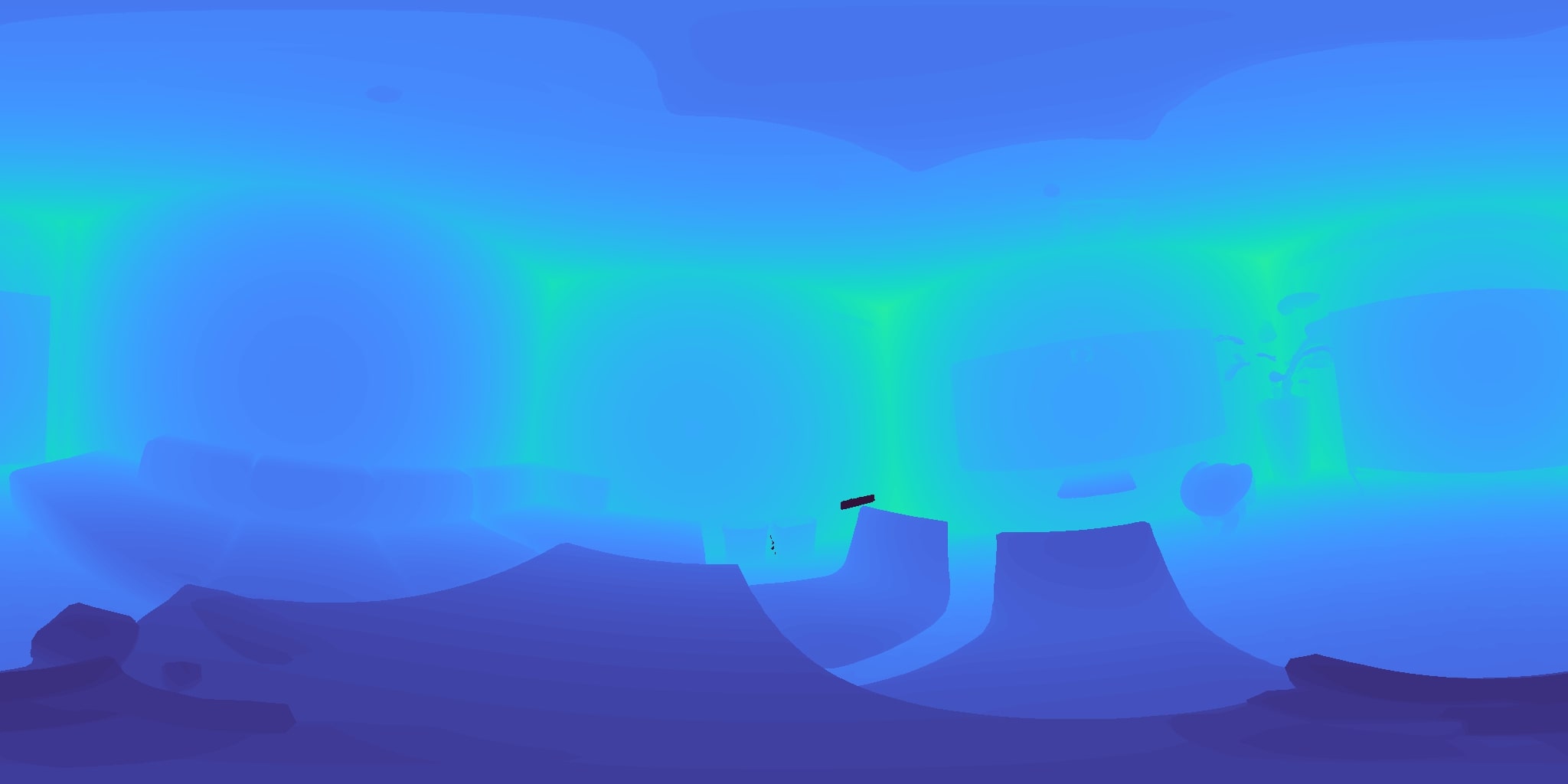}};
	\node [image,below=\padding]  (col2-row2) at (col2-row1.south) {\subfig{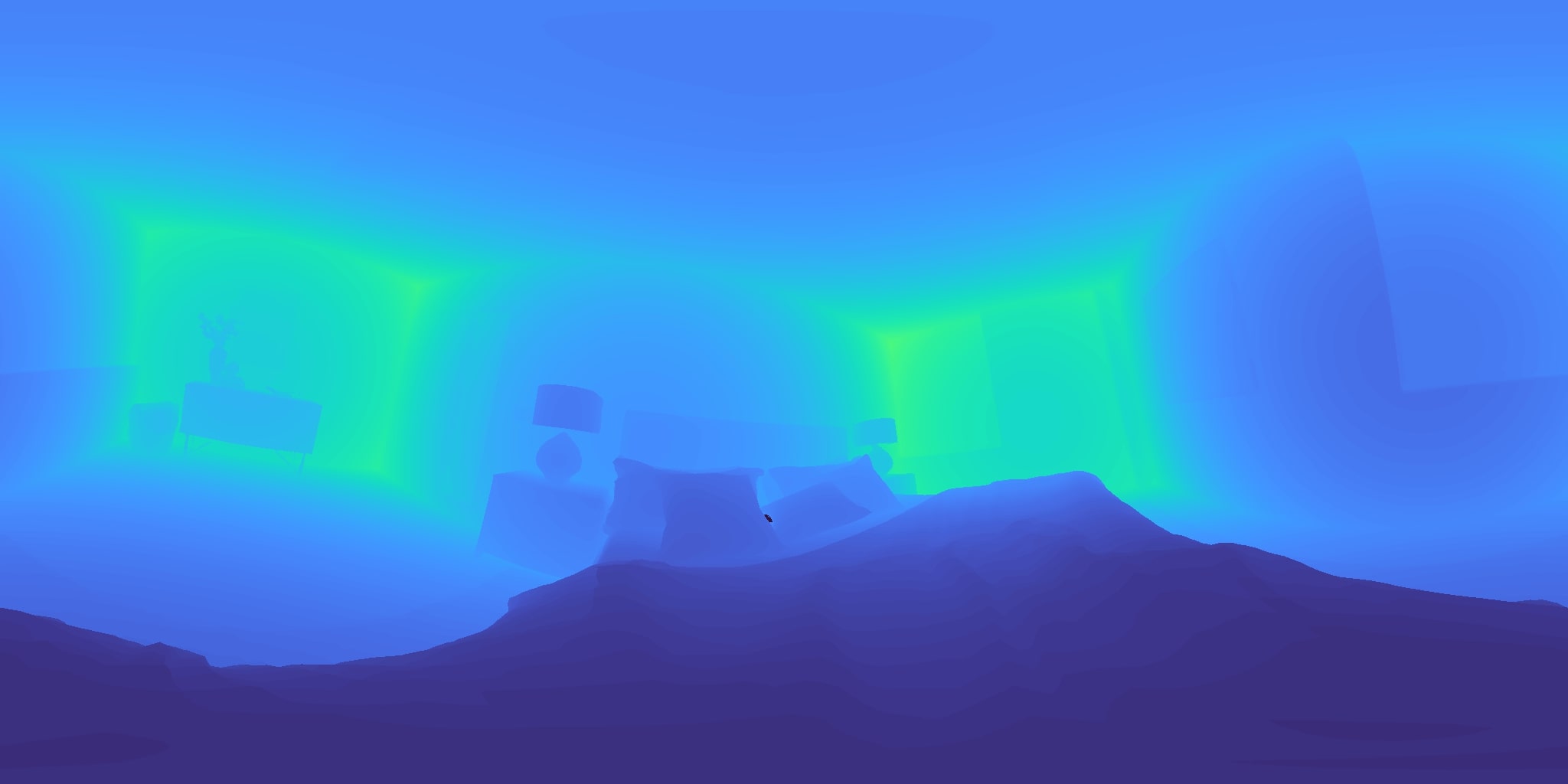}};
	
	\node [image,below=\padding]  (col1-row3) at (col1-row2.south) {\subfig[0 50 0 50]{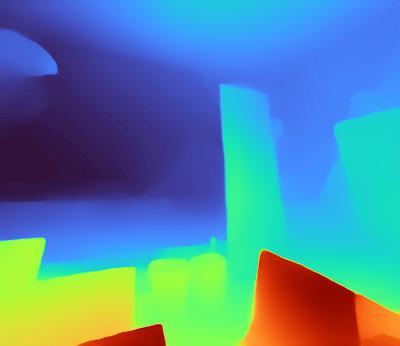}};
	\node [image,below=\padding]  (col2-row3) at (col2-row2.south) {\subfig[0 50 0 50]{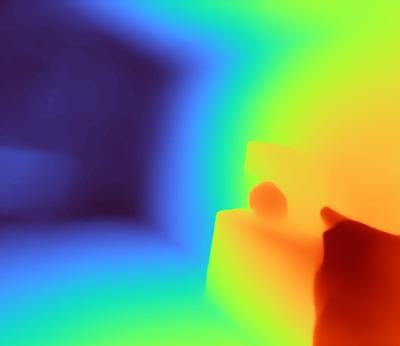}};
	
	\node [image,below=\padding]  (col1-row4) at (col1-row3.south) {\subfig[500 350 800 300]{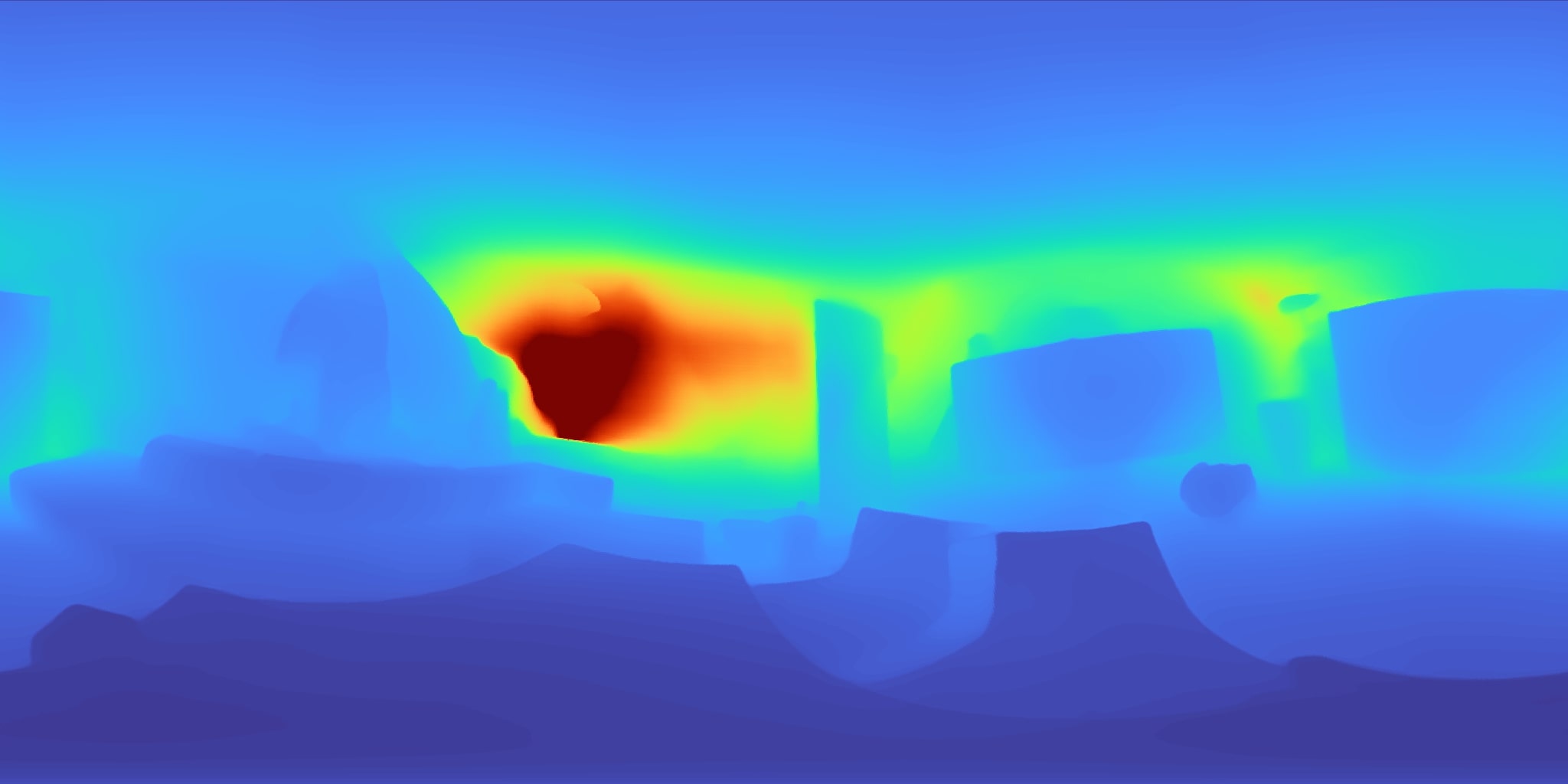}};
	\node [image,below=\padding]  (col2-row4) at (col2-row3.south) {\subfig{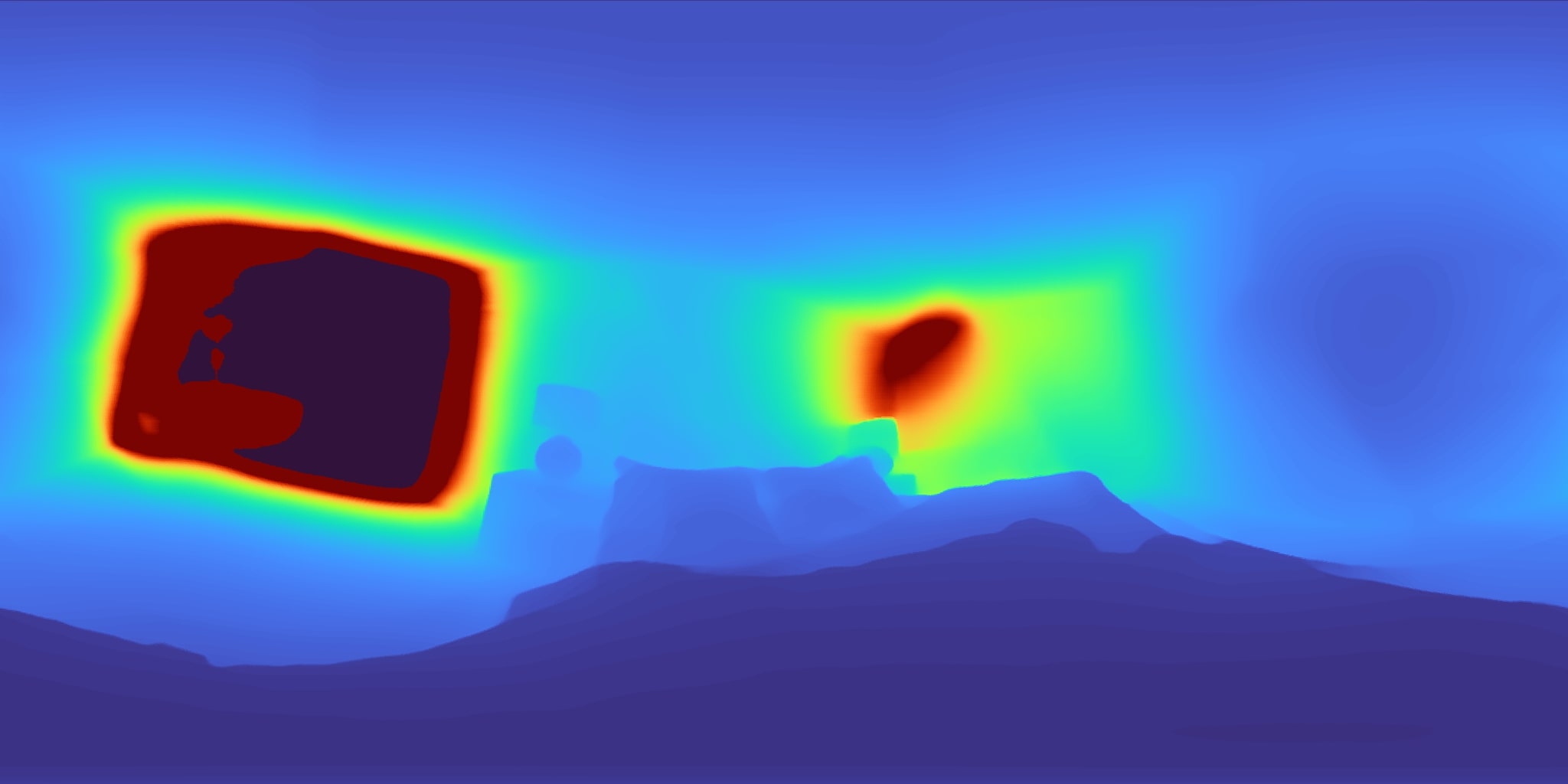}};

	\node [collabel] at ($(col1-row1.north west)!0.5!(col1-row1.north east)$) {Photorealistic textures (crop)};
	\node [collabel] at ($(col2-row1.north west)!0.5!(col2-row1.north east)$) {Error in least-squares rescaling};
	\node [rowlabel] at (col1-row1.west) {Input image};
	\node [rowlabel] at (col1-row2.west) {Ground truth};
	\node [rowlabel] at (col1-row3.west) {Tangent disparity};
	\node [rowlabel] at (col1-row4.west) {Ours\Mviii};

\end{tikzpicture}%
	\caption{\label{fig:failure-cases}%
		Failure cases for our method.
		\textbf{Left:}
		Our method cannot overcome incorrect tangent disparity estimates such as this photorealistic textured wall, which is treated as if it was an island view and not a wall.
		\textbf{Right:}
		In some cases, the least-squares rescaling to fit the ground-truth disparity range results in negative disparities, which produces incorrect, negative depth values (dark purple).
	}
\end{figure}

%% file: figures/incorrect-GT.tex
\begin{figure*}%
\centering
\begin{tikzpicture}[
	image/.style={inner sep=0pt, outer sep=0pt},
	collabel/.style={above=9pt, anchor=north, inner ysep=0pt, scale=0.8, align=center},
	rowlabel/.style={left=9pt, rotate=90, anchor=north, inner ysep=0pt, scale=0.8, align=center},
	subcaption/.style={inner xsep=0.75mm, inner ysep=0.75mm, below right},
	arrow/.style={-{Latex[length=2.5mm,width=4mm]}, line width=2mm},
	spy using outlines={rectangle, size=1.85cm, magnification=4, connect spies, ultra thick, every spy on node/.append style={thick}},
	style1/.style={cyan!90!black,thick},
	style2/.style={orange!90!black},
	style3/.style={blue!90!black},
	style4/.style={red!90!black},
	]
	
	\def\padding{0.005\linewidth}
	\newcommand{\subfig}[2][0px 0px 0px 0px]{\includegraphics[height=3.4cm, trim=#1, clip]{figures/#2}}

	\node [image]                 (col1-row1) at (0,0)             {\subfig{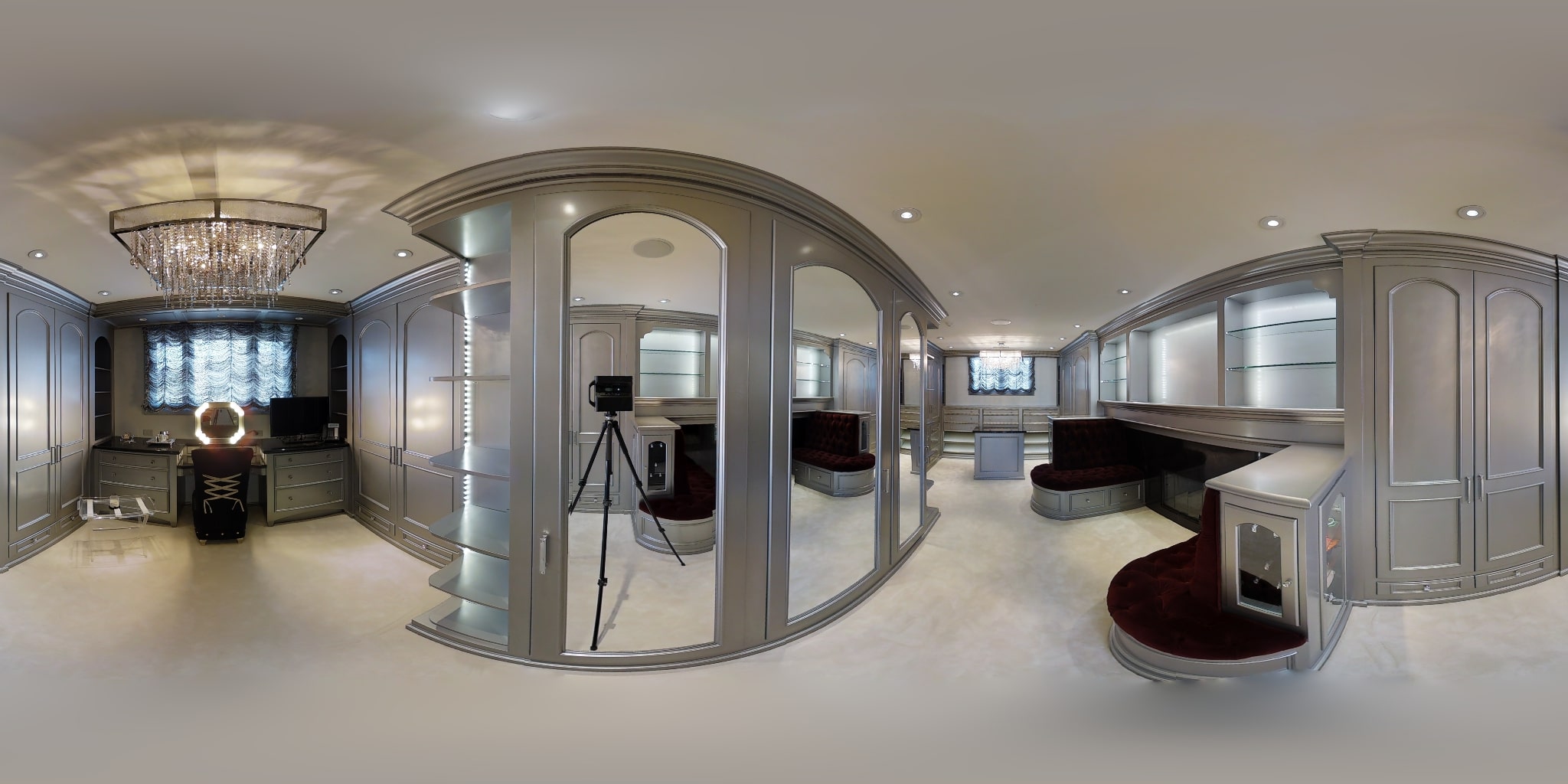}};
	\node [image,right=\padding]  (col2-row1) at (col1-row1.east)  {\subfig{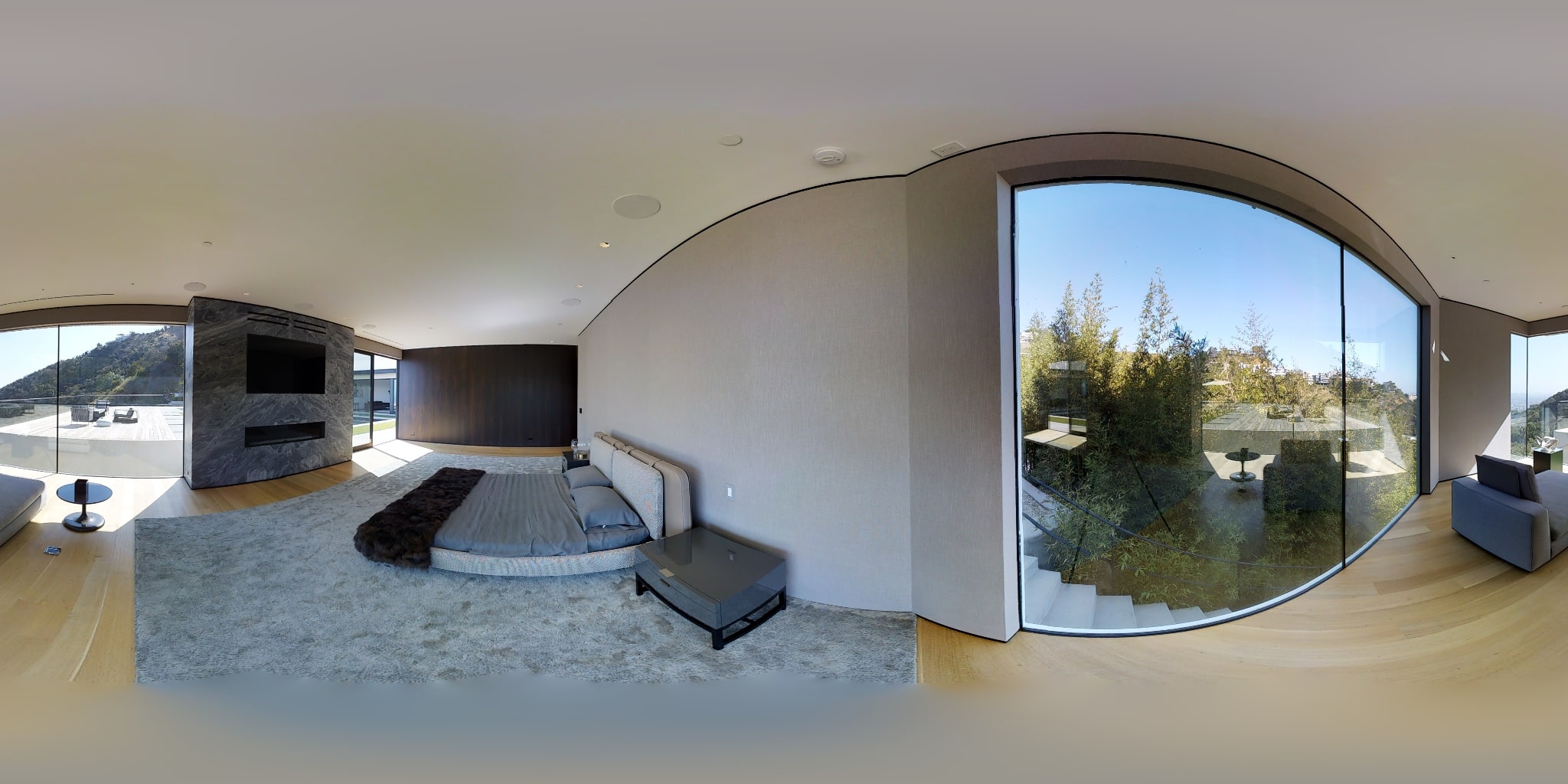}};
	\node [image,right=\padding]  (col3-row1) at (col2-row1.east)  {\subfig[704 480 864 64]{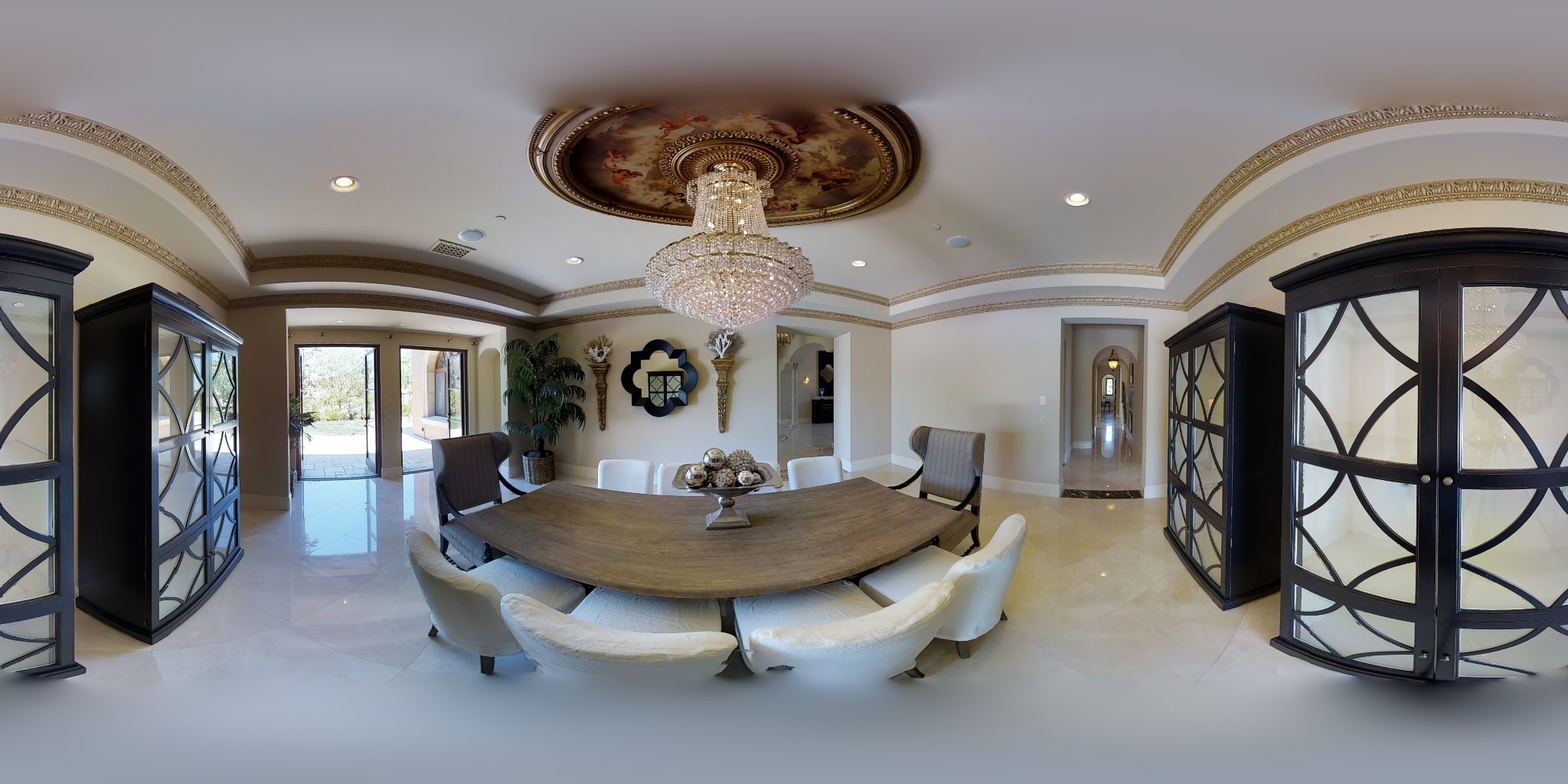}};

	\node [image,below=\padding]  (col1-row2) at (col1-row1.south) {\subfig{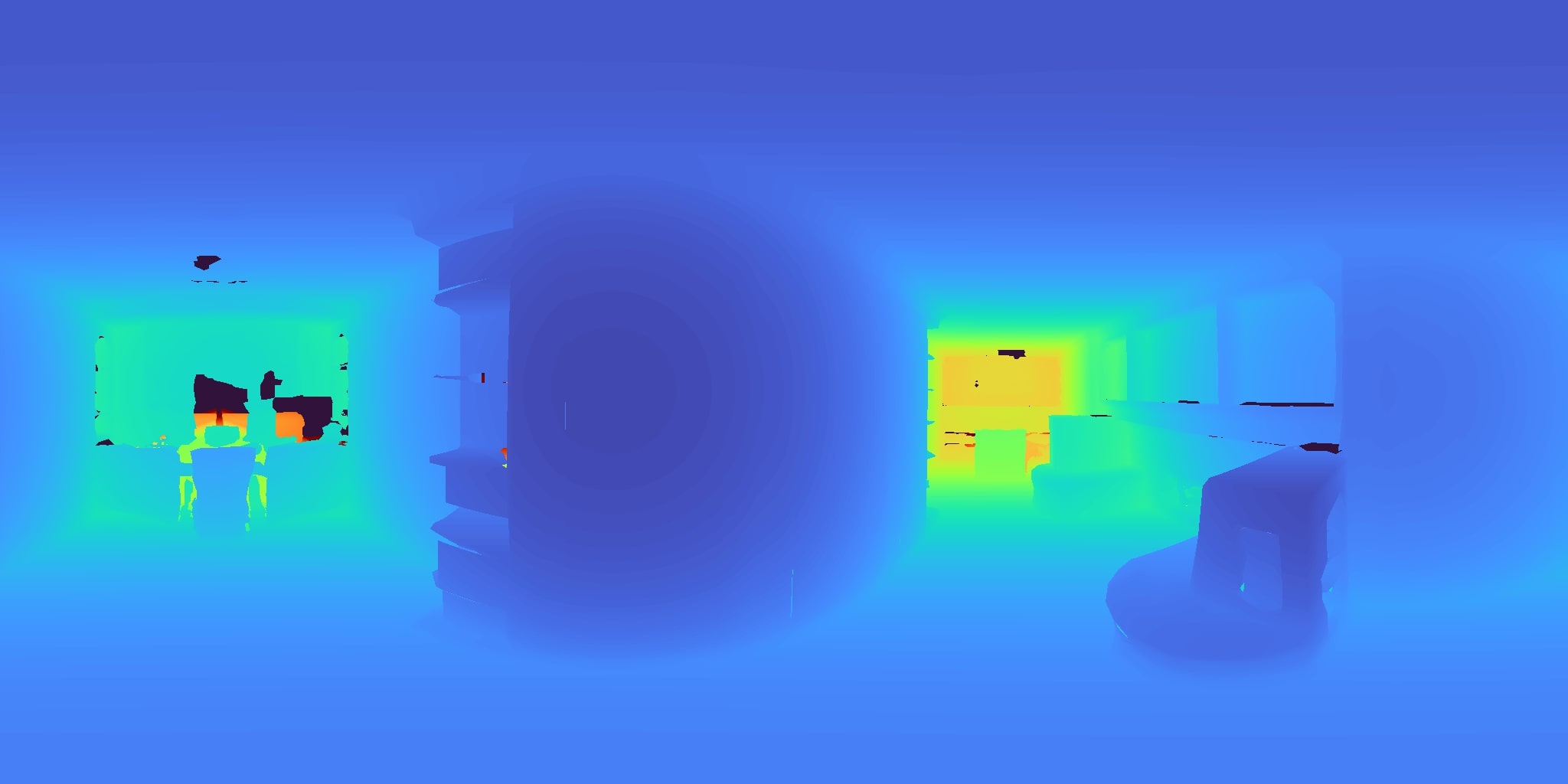}};
	\node [image,below=\padding]  (col2-row2) at (col2-row1.south) {\subfig{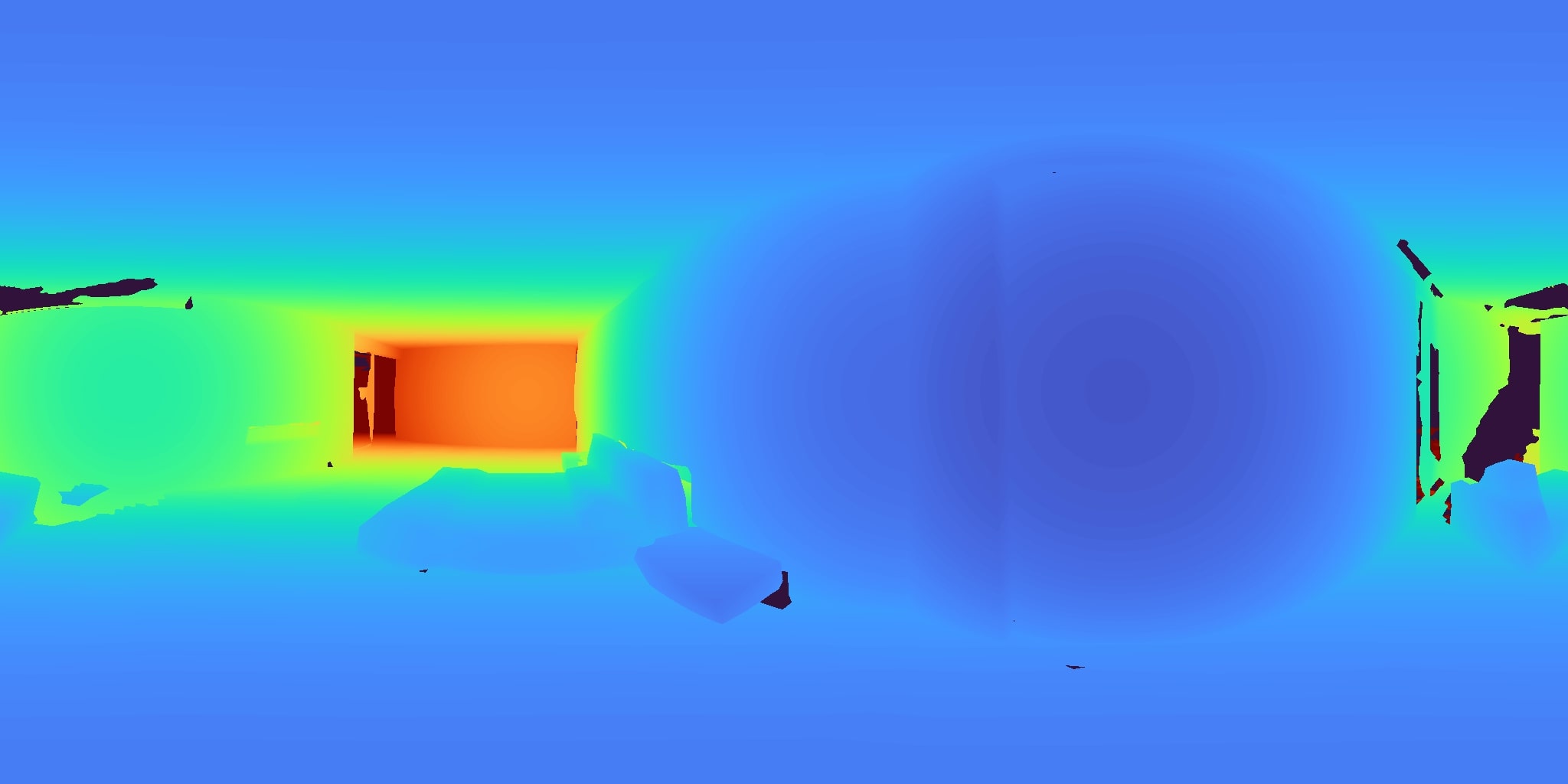}};
	\node [image,below=\padding]  (col3-row2) at (col3-row1.south) {\subfig[704 480 864 64]{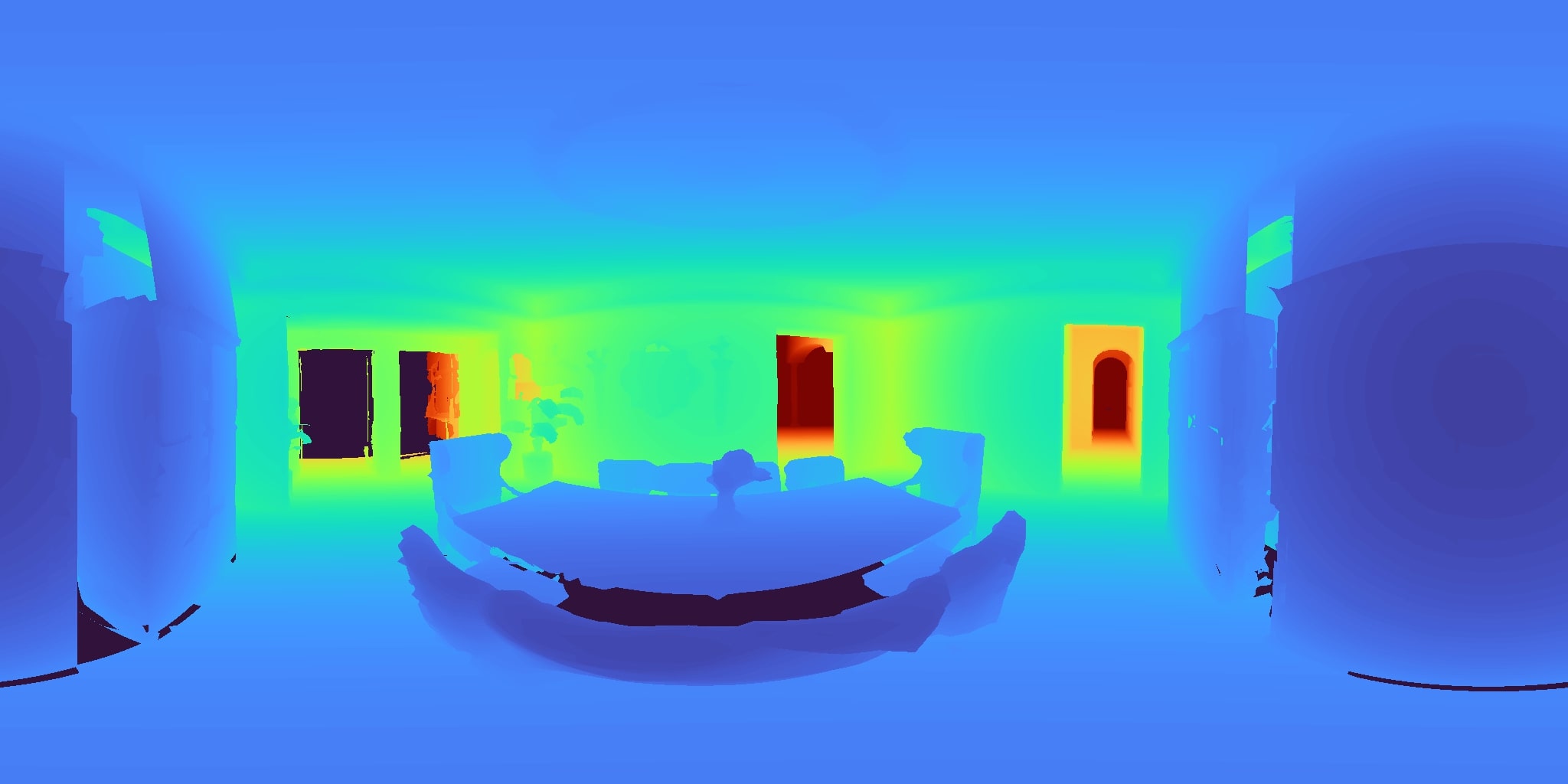}};
	
	\node [image,below=\padding]  (col1-row3) at (col1-row2.south) {\subfig{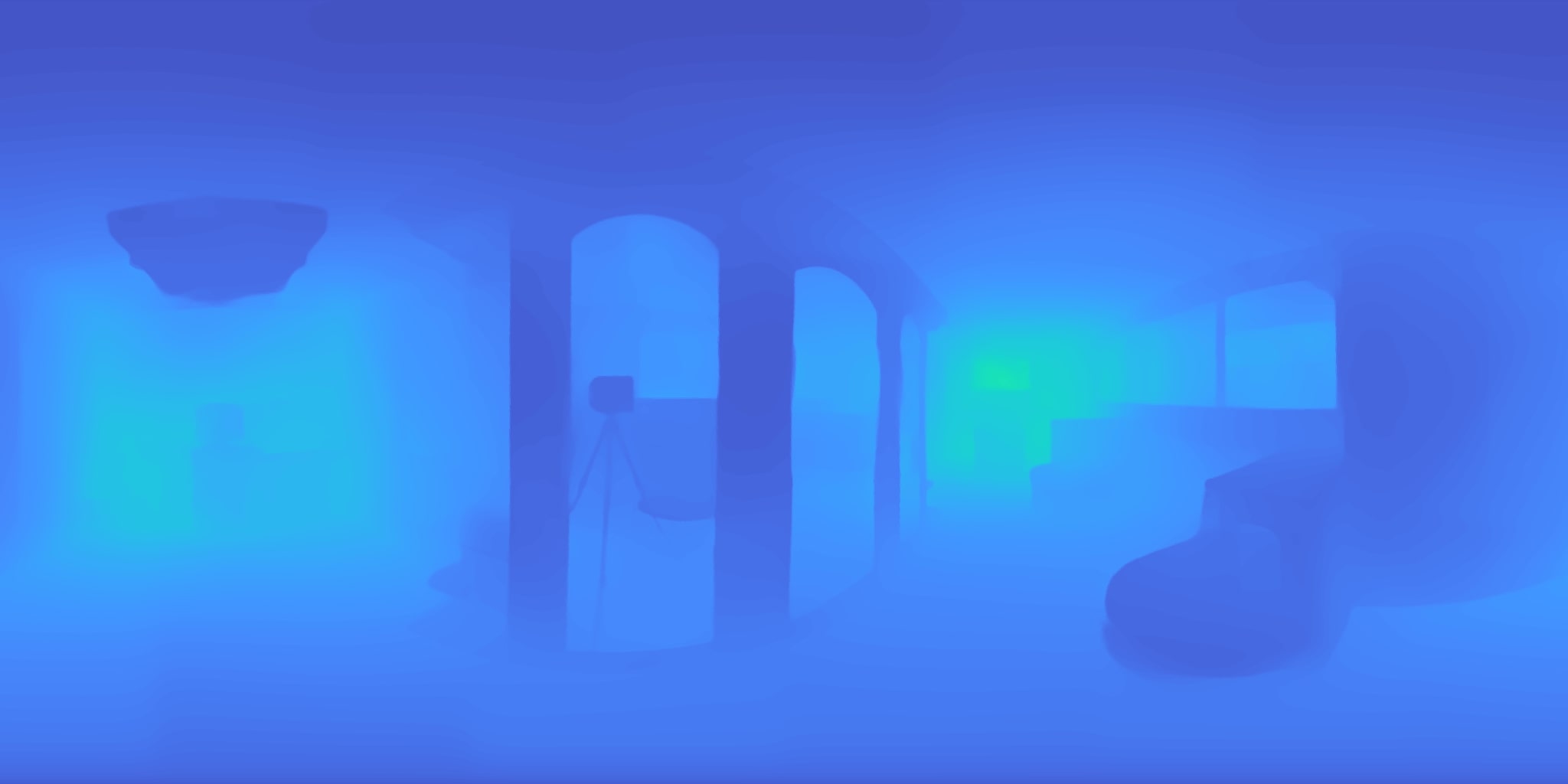}};
	\node [image,below=\padding]  (col2-row3) at (col2-row2.south) {\subfig{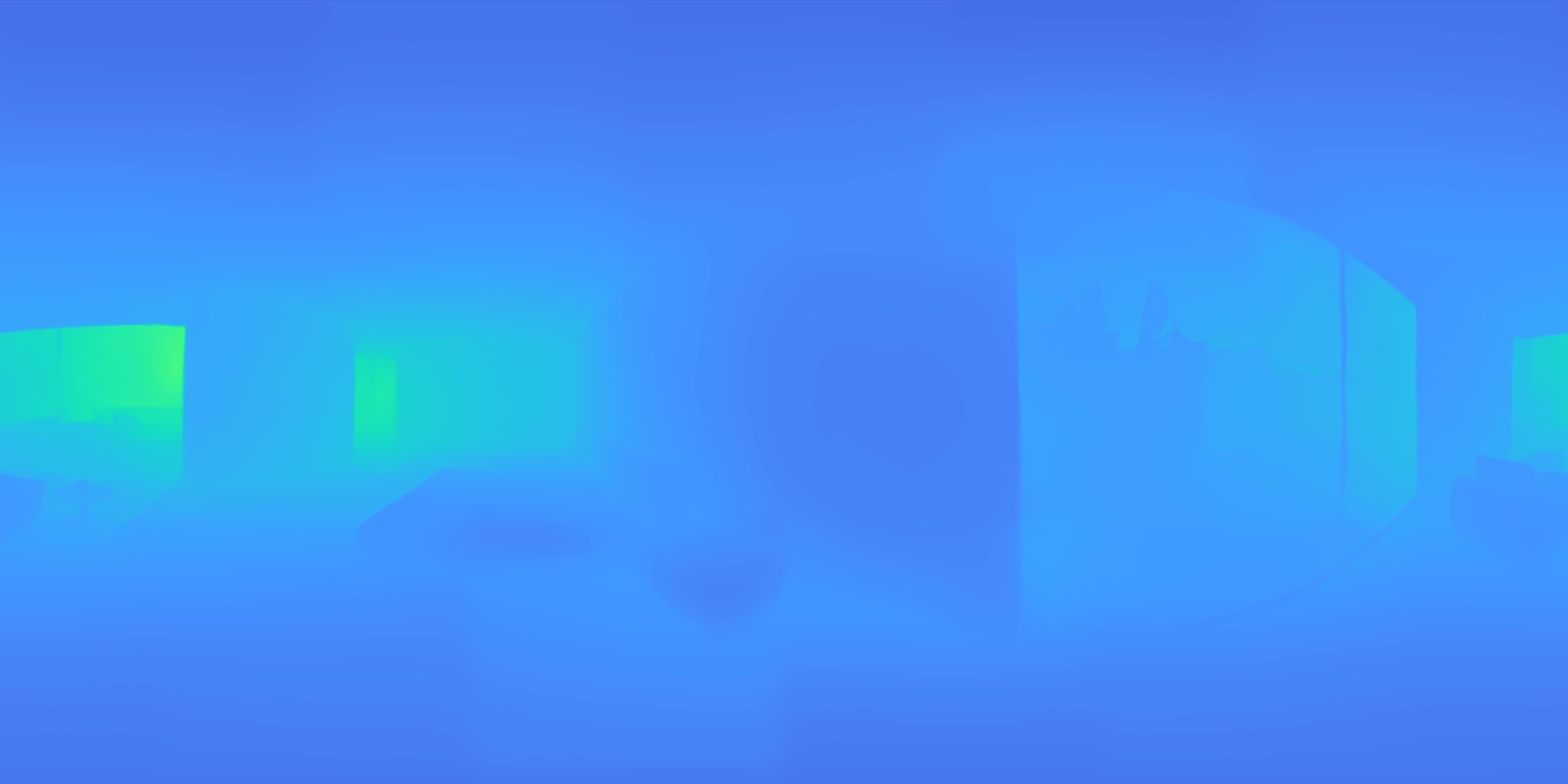}};
	\node [image,below=\padding]  (col3-row3) at (col3-row2.south) {\subfig[704 480 864 64]{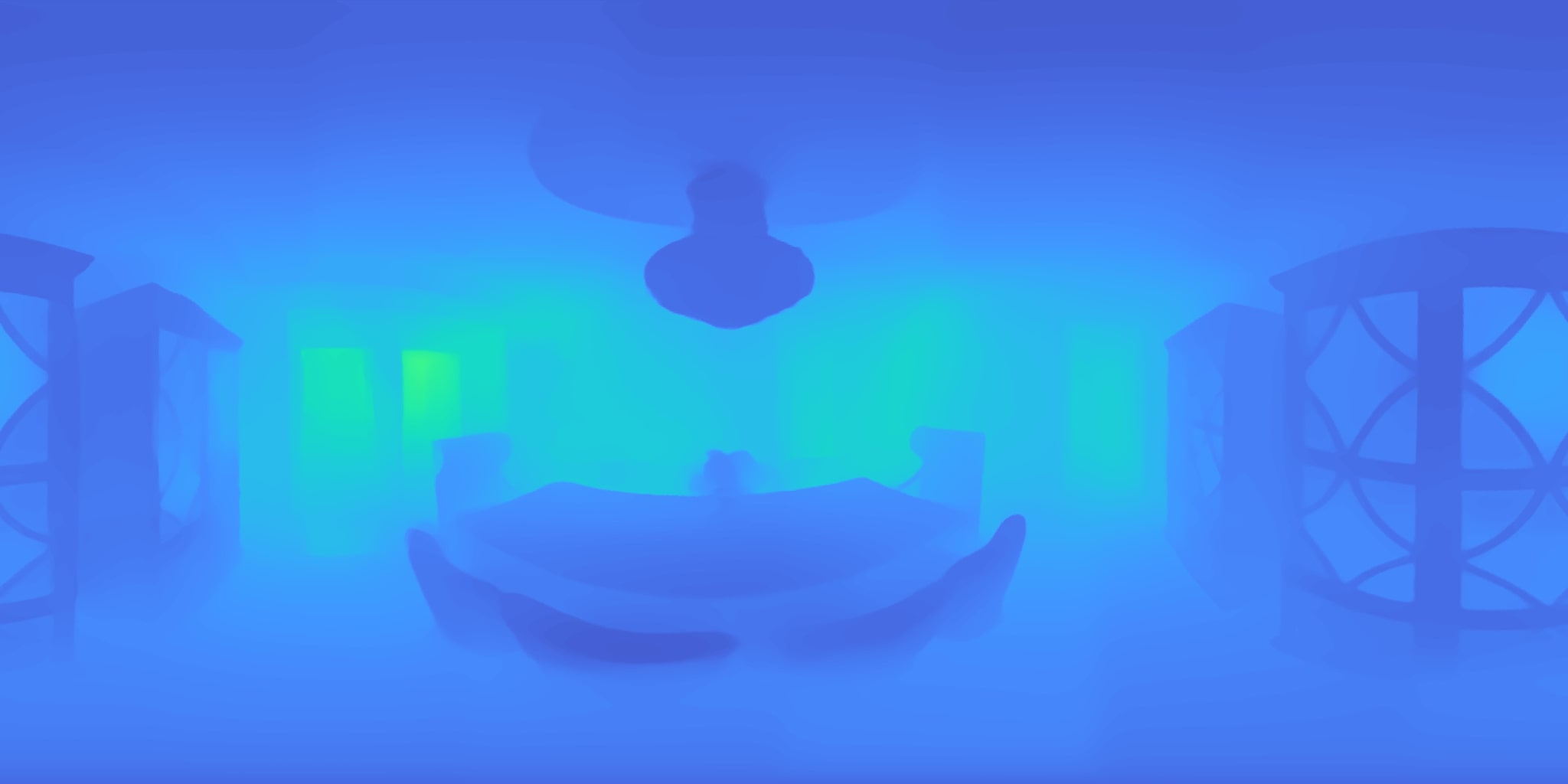}};

	\node [collabel] at ($(col1-row1.north west)!0.5!(col1-row1.north east)$) {Mirror surface (GT) versus visible reflection (ours)};
	\node [collabel] at ($(col2-row1.north west)!0.5!(col2-row1.north east)$) {Window surface (GT) versus outside scene (ours)};
	\node [collabel] at ($(col3-row1.north west)!0.5!(col3-row1.north east)$) {Crop: Missing chandelier};
	\node [rowlabel] at (col1-row1.west) {Input image};
	\node [rowlabel] at (col1-row2.west) {Ground-truth depth map};
	\node [rowlabel] at (col1-row3.west) {Ours\Mviii};
	
	\draw[arrow,blue] ($(col1-row1.center)+(-0.2,-1.4)$) -- ++(-0.4,0.4);
	\draw[arrow,blue] ($(col1-row1.center)+( 0.6,-1.2)$) -- ++(-0.4,0.4);
	\draw[arrow,blue] ($(col2-row1.center)+(-2.5,-0.6)$) -- ++(-0.4,0.4);
	\draw[arrow,blue] ($(col2-row1.center)+( 1.0, 0.2)$) -- ++( 0.4,0.4);
	\draw[arrow,blue] ($(col3-row1.center)+(-0.5,-0.8)$) -- ++( 0.4,0.4);
	
	\draw[arrow,red!80!black] ($(col1-row2.center)+(-0.2,-1.4)$) -- ++(-0.4,0.4);
	\draw[arrow,red!80!black] ($(col1-row2.center)+( 0.6,-1.2)$) -- ++(-0.4,0.4);
	\draw[arrow,red!80!black] ($(col2-row2.center)+(-2.5,-0.6)$) -- ++(-0.4,0.4);
	\draw[arrow,red!80!black] ($(col2-row2.center)+( 1.0,-0.2)$) -- ++( 0.4,0.4);
	\draw[arrow,red!80!black] ($(col3-row2.center)+(-0.5,-0.8)$) -- ++( 0.4,0.4);
	
	\draw[arrow,green!85!black] ($(col1-row3.center)+(-0.2,-1.4)$) -- ++(-0.4,0.4);
	\draw[arrow,green!85!black] ($(col1-row3.center)+( 0.6,-1.2)$) -- ++(-0.4,0.4);
	\draw[arrow,green!90!black] ($(col2-row3.center)+(-2.5,-0.6)$) -- ++(-0.4,0.4);
	\draw[arrow,green!85!black] ($(col2-row3.center)+( 1.0,-0.2)$) -- ++( 0.4,0.4);
	\draw[arrow,green!85!black] ($(col3-row3.center)+(-0.5,-0.8)$) -- ++( 0.4,0.4);

\end{tikzpicture}%
	\caption{\label{fig:incorrect-GT}%
		Inconsistent ground-truth depth maps in Matterport3D \cite{ChangDFHNSSZZ2017}.
		\textbf{Left:}
		The mesh geometry covers the surface of the mirrors instead of representing the reflection of the visible scene.
		\textbf{Centre:}
		The large windows in the room are treated as if they were opaque, instead of showing the depth of the environment outside or being masked out.
		\textbf{Right:}
		The chandelier is missing in the mesh but reconstructed by our method.
	}
\end{figure*}